%% file: main.tex
\documentclass[10pt,twocolumn,letterpaper]{article}

\usepackage[pagenumbers]{cvpr} 

\input{preamble}

\definecolor{cvprblue}{rgb}{0.21,0.49,0.74}
\usepackage[pagebackref,breaklinks,colorlinks,allcolors=cvprblue]{hyperref}

\def\confName{CVPR}
\def\confYear{2027}
\newcolumntype{x}[1]{>{\centering\let\newline\\\arraybackslash\hspace{0pt}}p{#1}}

\newcommand{\methodname}{ReCoCo}

\title{ReCoCo: From Patches to Compositions for Visual Representation Learning
}

\author{
Juan Gutiérrez$^{1,2}$ \quad
Saman Zadtootaghaj$^{3}$ \quad
José Luis Blanco-Murillo$^{1}$ \quad
Sara Atito$^{2,4}$\\[0.5em]
$^{1}$Information Processing and Telecommunications Center (IPTC), Universidad Politécnica de Madrid, ES\\
$^{2}$Centre for Vision, Speech and Signal Processing (CVSSP), University of Surrey, UK\\
$^{3}$Sony Interactive Entertainment, Berlin, DE\\
$^{4}$Surrey Institute for People-Centred Artificial Intelligence (PAI), University of Surrey, UK 
}

\begin{document}
\twocolumn[{%
\renewcommand\twocolumn[1][]{#1}%
\maketitle
\begin{center}
    \captionsetup{type=figure}
    \begin{minipage}[c]{0.47\textwidth}\centering\includegraphics[width=\linewidth]{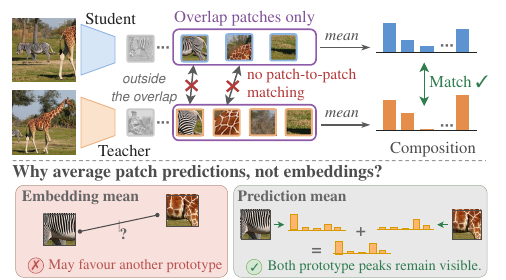}\end{minipage}%
    \begin{minipage}[c]{0.265\textwidth}\centering\includegraphics{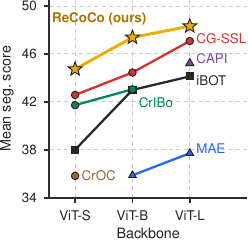}\end{minipage}%
    \begin{minipage}[c]{0.265\textwidth}\centering\includegraphics{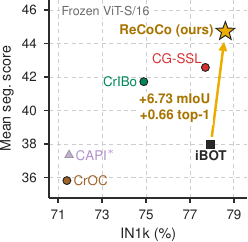}\end{minipage}\\[1pt]
    {\footnotesize%
    \begin{minipage}[t]{0.47\textwidth}\centering (a) ReCoCo matches compositions of the overlap.\end{minipage}%
    \begin{minipage}[t]{0.265\textwidth}\centering\hspace*{0.8cm}(b) Dense vs.\ backbone.\end{minipage}%
    \begin{minipage}[t]{0.265\textwidth}\centering\hspace*{0.8cm}(c) Dense vs.\ global transfer.\end{minipage}}\\[2pt]
    \captionof{figure}{
    \textbf{\methodname{} matches regional compositions across views.} (a) The student and teacher process two overlapping views of an image. Rather than matching individual patches, \methodname{} averages the patch predictions within the shared region to obtain one composition per view, then aligns the two compositions. Unlike averaging embeddings, averaging predictions preserves the contributions of multiple visual concepts (bottom).
    (b) Mean mIoU of six frozen segmentation settings ($k$-NN and linear probes on ADE20K, PASCAL VOC and Cityscapes) across backbones.
    (c) The same mean against ImageNet-1K linear accuracy (ViT-S/16). $^{*}$Pretrained on LVD-142M~\cite{oquab2024dinov2}.
\vspace*{1pt}
    }
    \label{fig:dense_overview}
\end{center}%
}]
\input{sec/0_abstract}    
\input{sec/1_intro}
\input{sec/2_related}
\input{sec/3_method}
\input{sec/4_experiments}

\input{sec/5_conclusions}


{
    \small
    \bibliographystyle{IEEEtran}
    \bibliography{main}
}

\input{sec/suppl}

\end{document}

%% file: preamble.tex
\usepackage{bbm}
\usepackage{amsmath}
\usepackage{amsthm}

\theoremstyle{remark}

\theoremstyle{plain}

\makeatletter
\newbox\recoco@labels
\newif\ifrecoco@inlabel
\newif\ifrecoco@newlist
\@ifundefined{@EveryShipout@Hook}{}{%
  \preto\@EveryShipout@Hook{%
    \global\setbox\recoco@labels\box\@labels
    \global\setbox\@labels\box\voidb@x
    \if@inlabel\global\recoco@inlabeltrue\else\global\recoco@inlabelfalse\fi
    \if@newlist\global\recoco@newlisttrue\else\global\recoco@newlistfalse\fi
    \global\@inlabelfalse\global\@newlistfalse}%
  \appto\@EveryShipout@Hook{%
    \global\setbox\@labels\box\recoco@labels
    \ifrecoco@inlabel\global\@inlabeltrue\else\global\@inlabelfalse\fi
    \ifrecoco@newlist\global\@newlisttrue\else\global\@newlistfalse\fi}%
}
\makeatother
\usepackage{soul}
\usepackage[table]{xcolor}
\usepackage{multirow}
\usepackage{booktabs}
\usepackage{tabularx}
\usepackage{pifont}
\usepackage{adjustbox}
\usepackage{makecell}

\definecolor{blue}{RGB}{211,243,253}

\usepackage{pgfplots}
\usepgfplotslibrary{groupplots}
\usepackage{tikz}
\pgfplotsset{compat=1.18}
\usepackage{float}
\usepackage{placeins}

%% file: sec/0_abstract.tex
\begin{abstract}
Self-supervised learning has produced powerful visual representations, yet learning strong dense features remains challenging. While consistency between augmented views provides strong image-level supervision, extending this principle to local visual content is less straightforward. Existing approaches often match individual patches across views or identify coherent regions, introducing uncertainties in correspondence estimation and assumptions about how visual content should be grouped. We introduce ReCoCo (Regional Composition Consistency), a principled approach that extends cross-view consistency to the regional level without requiring such correspondences or semantic grouping. ReCoCo represents the content shared between augmented views through compositions of their patch-level predictions over learned visual prototypes. By aligning these compositions, the model learns from regions containing multiple visual concepts without requiring them to be separated beforehand. ReCoCo integrates naturally with teacher-student self-distillation and introduces no additional learned parameters. Extensive experiments demonstrate substantial improvements in frozen-feature segmentation, clustering, and regional retrieval while maintaining competitive image-level recognition, alleviating the trade-off between dense and global representation quality.
\enlargethispage{1\baselineskip}
\end{abstract}


%% file: sec/1_intro.tex
\section{Introduction}
\label{section:intro}




Self-supervised learning (SSL) has established cross-view consistency as a powerful principle for learning transferable visual representations \cite{chen2020simclr,chen2021empirical,atito2021sit,caron2021dino,zhou2022ibot}.
By encouraging different augmented views of an image to produce compatible representations, models learn to capture visual information that persists across changes in appearance, scale, and spatial coverage. However, when augmented views share only part of an image, it becomes less clear what their representations should agree on. At the image level, agreement can be enforced between global representations. At the patch level, it typically requires identifying which visual elements correspond. Neither formulation directly expresses agreement on the heterogeneous content shared between two views.

This limitation becomes more apparent in dense representation learning. Methods such as DINO \cite{caron2021dino} enforce agreement between global representations of augmented views, but do not explicitly capture how objects, parts, and background contribute to them. iBOT \cite{zhou2022ibot} introduces finer supervision through masked-patch prediction, although its patch-level objective operates within the same crop rather than across different views. Neither objective directly enforces consistency over the local content shared between views, leaving a gap between global agreement and individual-patch supervision.

Extending cross-view consistency to local content is challenging because augmentation changes not only appearance but also the spatial support of visual tokens. Existing approaches commonly address this by establishing patch correspondences through feature similarity
%
%
\cite{li2022esvit,wang2021densecl} or augmentation geometry \cite{xie2021pixpro,lebailly2023globallocal,bardes2022vicregl}. The former relies on potentially unreliable learned features, while the latter matches spatial locations that may capture different visual content due to differences in crop scale and patch resolution. Both approaches therefore depend on potentially imperfect one-to-one correspondences. 

To overcome the limitations of individual patch matching, existing methods explore supervision over larger groups of visual content. However, a region may contain multiple objects, parts, and background, whose distinct information can be obscured when their features are averaged into a single representation. To address this, geometry-based methods align smaller corresponding regions within overlapping views to preserve local information \cite{roh2021scrl,xiao2021resim}, while others identify object-like groups through segmentation, clustering, or learned feature similarity \cite{henaff2021detcon,ziegler2022leopart,stegmuller2023croc,lebailly2024cribo,atito2025cgssl}. Although these approaches provide more focused supervision, they require either selecting smaller regions or identifying meaningful groups, introducing additional assumptions about which visual content should be aligned.

We take a different perspective in which cross-view consistency is enforced through the composition of shared visual content, rather than individual patch correspondences or pooled feature representations. By representing a region through the distribution of its patch-level prototype predictions, we can account for multiple visual concepts without explicitly separating them. The strength of this formulation lies in its simplicity. It provides a principled way to enforce consistency over shared visual content across different crop scales and patch resolutions, without requiring explicit correspondences or semantic grouping. Augmentation geometry determines which content is shared, while prototype distributions define how agreement is enforced.

Building on this idea, we introduce ReCoCo (\textbf{Re}gional \textbf{Co}mposition \textbf{Co}nsistency), which aligns the prototype compositions of overlapping augmented views (Figure~\ref{fig:method_overview}). For each pair of views, we identify their shared region and independently average the patch-level prototype predictions within it. The student is then trained to match the teacher's composition, without requiring aligned patch grids or equal numbers of patches. This provides regional supervision alongside image-level and masked-patch objectives within the same prototype space, without extra parameters.


Empirically, ReCoCo delivers substantial improvements over existing SSL approaches in frozen-feature dense prediction while maintaining competitive image-level recognition (Figure~\ref{fig:dense_overview}). For example, on PASCAL VOC, it improves over the baseline by 14.6 and 10.8 mIoU points under $k$-NN and linear probing, respectively, with gains extending across downstream tasks and larger backbones.

\noindent
Our contributions are summarised as follows:
\begin{itemize}
    \item We introduce ReCoCo,  a regional self-supervised objective that enforces consistency through the composition of shared visual content, without requiring patch correspondences, semantic grouping, or extra parameters.
    \item We investigate the design choices underlying effective regional supervision through systematic ablations, demonstrating the importance of prediction-space aggregation, patch selection, and view pairing. Our feature analyses further reveal stronger within-object coherence and between-class separation.
    \item We demonstrate substantial improvements across semantic segmentation, clustering, and retrieval tasks while preserving image-level recognition, alleviating the trade-off between dense and global representation quality.
\end{itemize}

%% file: sec/2_related.tex
\begin{figure*}[t]
    \centering
    \includegraphics[width=0.97\linewidth]{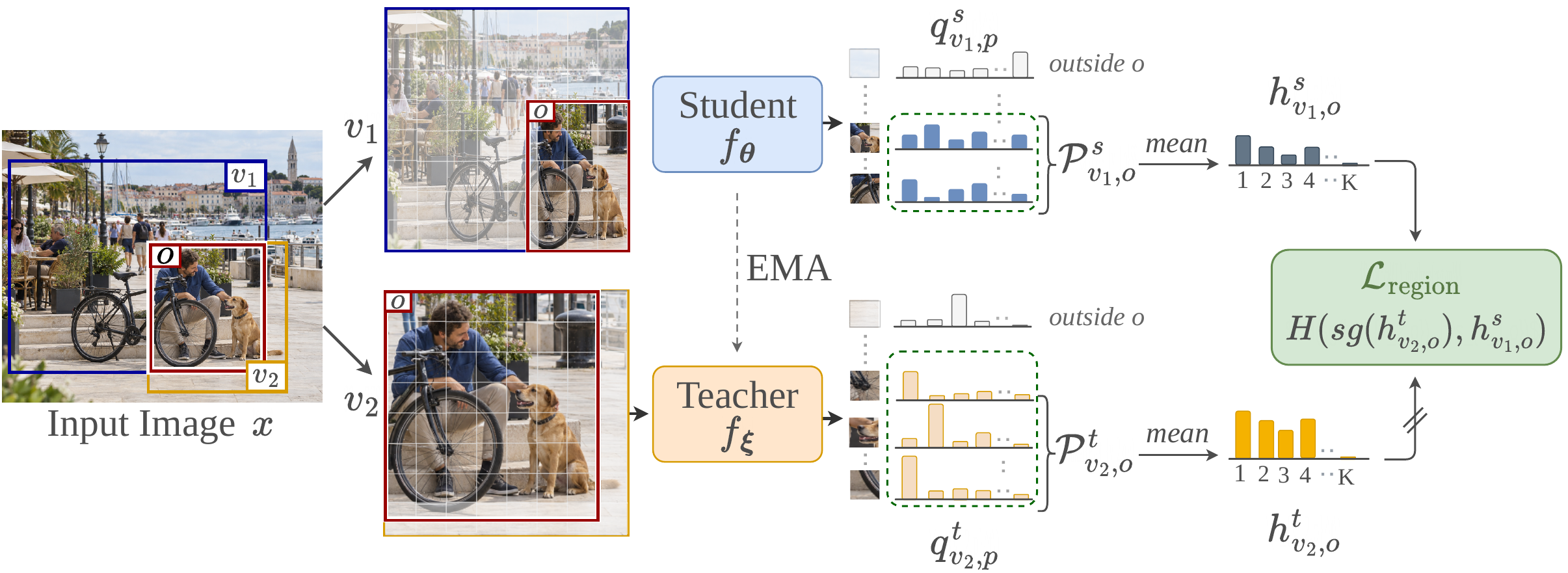}
    \caption{
    \textbf{Overview of ReCoCo's regional composition consistency.} Two augmented views, $v_1$ and $v_2$, of an image $x$ share an overlapping region $o$. The student and its EMA teacher process the full views and predict distributions over learned prototypes for individual patches. Only patches fully contained within the overlap are selected, and their predictions are independently averaged to obtain the student and teacher regional compositions, $h_{v_1,o}^{s}$ and $h_{v_2,o}^{t}$. The regional consistency loss aligns the student composition with the stop-gradient teacher composition, without requiring patch-wise correspondences. Both networks comprise an encoder and a projection head; the projection heads and student masking are omitted for visual simplicity. 
    }
    \label{fig:method_overview}
\end{figure*}

\section{Related Work}
\label{section:related_works}

\noindent
\textbf{Image- and patch-level self-supervision.}
Self-supervised pretraining objectives typically operate at two spatial granularities. At the image level, contrastive~\cite{chen2020simclr,he2020moco}, non-contrastive~\cite{grill2020byol}, and clustering-based~\cite{caron2020swav} methods enforce agreement between the global representations of augmented views. DINO~\cite{caron2021dino} casts this agreement as teacher-student self-distillation over learned prototypes and shows that it also yields spatially meaningful features with Vision Transformers. At the patch level, masked image modelling predicts the pixels~\cite{he2022mae}, latent representations~\cite{assran2023ijepa}, or cluster assignments~\cite{darcet2025capi} of masked patches. iBOT~\cite{zhou2022ibot} combines both levels, complementing DINO with masked-patch self-distillation within the same view, and DINOv2~\cite{oquab2024dinov2} scales this combination to curated data and large models. However, strong global features do not ensure equally strong dense features, and the two levels of supervision can interfere~\cite{simeoni2025dinov3,marouani2026cls}. To improve dense features, recent pretraining recipes strengthen patch-level supervision further, extending the patch-level loss to visible tokens~\cite{cao2026tipsv2,murlabadia2026vjepa21} and intermediate layers~\cite{murlabadia2026vjepa21}, or giving masked boundary tokens a geometric target~\cite{fu2026lingbot}. Yet in these frameworks, the image-level loss compares two views only as wholes and the patch-level loss acts within a single view, so neither enforces agreement on the local content the views share. ReCoCo adds this agreement as a regional pretraining objective alongside image-level and masked-patch losses.

\noindent
\textbf{Cross-view patch correspondence.}
To extend cross-view agreement to local content, most dense methods first establish which patches correspond. Similarity-based methods pair each local feature with its most similar feature in the other view~\cite{wang2021densecl,li2022esvit} or contrast attention-selected pixels across views~\cite{wang2022setsim}. Such matches are only as reliable as the features being learned and can be harmful in low-data regimes~\cite{lebailly2023globallocal}. Other methods instead rely on augmentation geometry, matching locations that lie close together in the original image~\cite{xie2021pixpro,lebailly2023globallocal}, and VICRegL~\cite{bardes2022vicregl} combines both kinds of matching. However, when crop scales differ, matched patches differ in size and may contain different content. ADCLR~\cite{zhang2023adclr} avoids patch correspondences by appending identical small crops as query tokens to both views and distilling their outputs. However, this introduces additional tokens and relies on shared query crops rather than directly supervising the content within the overlapping views.

\noindent
\textbf{Region-level self-supervision.}
Region-level methods compare pooled representations of regions rather than individual patches. Pooling features over the overlap of two views, however, merges the objects, parts, and background it may contain into one representation. Geometry-based methods therefore pool smaller regions inside the overlap, random boxes in SCRL~\cite{roh2021scrl} and fixed-size sliding windows in ReSim~\cite{xiao2021resim}. Semantic grouping methods pool features within object-like regions, identified by unsupervised segmentation~\cite{henaff2021detcon}, soft prototype assignments~\cite{wen2022slotcon}, clustering~\cite{stegmuller2023croc,lebailly2024cribo}, or the attention of concept tokens~\cite{atito2025cgssl}. Both families focus supervision on smaller or explicitly grouped regions, and neither aggregates the overlap of two views as a whole. ReCoCo averages patch predictions instead of pooling features. The resulting composition preserves the probability mass assigned to each prototype even when the overlap contains several concepts, so ReCoCo can supervise all patches fully inside the overlap of every view pair. The supervised region is thus determined by the augmentations alone, rather than by sampled boxes, sliding windows, or discovered groups. The regional target also lies in the same prototype space as iBOT's image-level and masked-patch targets, so the three losses differ only in granularity.

%% file: sec/3_method.tex
\section{Method}
\label{sec:method}

\subsection{Preliminaries}

ReCoCo builds on teacher-student self-distillation with image- and patch-level objectives, as in DINO \cite{caron2021dino} and iBOT \cite{zhou2022ibot}. Given an image $x$, we generate $g$ global views and $\ell$ local views, denoted by $\mathcal G=\{v_1,\ldots,v_g\}$ and $\mathcal V_{\mathrm{loc}}$, respectively, forming the set of augmented views $\mathcal V=\mathcal G\cup\mathcal V_{\mathrm{loc}}$. The student $f_\theta$ processes all views, with masking applied to the global views. We denote by $\mathcal M_i$ the set of masked patch positions in global view $v_i$.
The teacher $f_\xi$, whose parameters are an exponential moving average (EMA) of the student's, processes only the unmasked global views.

In each network, the $[\mathrm{CLS}]$ and patch tokens are mapped through a shared projection head to distributions over $K$ learned prototypes, with the teacher predictions centred and sharpened. We denote the $[\mathrm{CLS}]$ and patch distributions of view $v$ by $\mathbf q_v^a$ and $\mathbf q_{v,p}^a$, respectively, where $p$ indexes a patch and $a\in\{s,t\}$ indicates student or teacher. Student predictions are indexed by the underlying view, with global views processed in their masked form and local views processed without masking.
The image-level loss $\mathcal L_{\mathrm{cls}}$ and masked-patch loss $\mathcal L_{\mathrm{patch}}$ are defined as
\begin{equation}
\begin{aligned}
    \mathcal L_{\mathrm{cls}}
    &=
    \frac{1}{g}
    \sum_{i=1}^{g}
    \frac{1}{|\mathcal V|-1}
    \sum_{\substack{v\in\mathcal V\\v\neq v_i}}
    H\!\left(
        \operatorname{sg}(\mathbf q_{v_i}^t),
        \mathbf q_v^s
    \right),
    \end{aligned}
\end{equation}
    \begin{equation}
\begin{aligned}
    \mathcal L_{\mathrm{patch}}
    &=
    \frac{1}{g}
    \sum_{i=1}^{g}
    \frac{1}{|\mathcal M_i|}
    \sum_{p\in\mathcal M_i}
    H\!\left(
        \operatorname{sg}(\mathbf q_{v_i,p}^t),
        \mathbf q_{v_i,p}^s
    \right).
\end{aligned}
\label{eq:distillation_losses}
\end{equation}
Here, $H(\mathbf a,\mathbf b)=-\sum_{k=1}^{K}a_k\log b_k$ denotes cross-entropy, and $\operatorname{sg}(\cdot)$ denotes the stop-gradient operation.

\subsection{Regional Composition}
\label{sec:region_representation}

ReCoCo encourages consistent representations of the visual content shared between augmented views. Rather than matching individual patches, whose spatial supports may differ across crops, we represent the shared region through the aggregate distribution of its patch predictions. This provides a correspondence-free representation of the common content without equal patch counts or aligned positions.

For each teacher global view $v_i\in\mathcal G$ and distinct student view $v\in\mathcal V$, let $o=o(v_i,v)$ denote their overlapping region in the original image, determined from the crop and flip parameters. As illustrated in Figure~\ref{fig:method_overview}, both networks process their respective full views, while only patches whose spatial footprints lie fully within the overlap are selected for regional composition. We denote these patch sets by $\mathcal P_{v_i,o}^t$ and $\mathcal P_{v,o}^s$ for the teacher and student, respectively. Patches crossing the overlap boundary are excluded to avoid incorporating content outside the shared region. The selected patch sets may differ in cardinality due to differences in crop scale and patch grids.

We obtain the regional compositions by averaging the independently normalised predictions of the selected patches:%
\begin{equation}
    \mathbf h_{v_i,o}^t
    =
    \frac{1}{|\mathcal P_{v_i,o}^t|}
    \sum_{p\in\mathcal P_{v_i,o}^t}
    \mathbf q_{v_i,p}^t,\quad
    \mathbf h_{v,o}^s
    =
    \frac{1}{|\mathcal P_{v,o}^s|}
    \sum_{p\in\mathcal P_{v,o}^s}
    \mathbf q_{v,p}^s.
\label{eq:regional_composition}
\end{equation}
\noindent
\textbf{Composition of sharpened patch predictions.}
Each composition is a distribution over the $K$ learned prototypes, with each component representing the average probability mass assigned to a prototype within the shared region. We obtain these compositions by averaging independently normalised patch predictions, with teacher predictions centred and sharpened before aggregation. While sharpening encourages confident teacher predictions at individual patches, the resulting composition can still distribute probability mass across multiple prototypes when the region contains diverse content. Averaging patch features before applying the projection head would not generally preserve this mixture of predictions.

\noindent
\textbf{Complementary patch-level constraints.}
By discarding patch identity and spatial ordering, regional composition allows two views to agree on the relative prototype mass within their shared content without establishing patch-wise correspondences. However, this agreement imposes a collective rather than an individual constraint, since different sets of patch predictions can produce the same regional composition. For instance, a region containing patches confidently assigned to two different prototypes may have the same composition as one in which all patches assign equal probability to both. This ambiguity is mitigated by the complementary masked-patch loss $\mathcal L_{\mathrm{patch}}$, which encourages confident individual predictions by aligning student patches with the teacher's sharpened targets at masked positions. The two losses therefore encourage consistency in both the composition of shared content and the predictions of its constituent patches, without requiring the region to be semantically homogeneous. In Appendix~\ref{sec:appx_information}, we further characterise these complementary constraints from an information-theoretic perspective.

\subsection{Training Objective}
\label{sec:objective}
We train the student to match the teacher's regional composition across different augmented views. Each teacher global view $v_i\in\mathcal G$ is paired with every other global view and every student local view. View pairs for which either view has no patches fully contained within the shared region are excluded from the regional loss. We average the regional cross-entropy separately over global-to-global and global-to-local comparisons, assigning equal weight to both.
For each teacher-student view pair, we define the regional cross-entropy as
\begin{equation}
    c_{v_i,v}
    =
    H\!\left(
        \operatorname{sg}(\mathbf h_{v_i,o}^t),
        \mathbf h_{v,o}^s
    \right),
    \qquad o=o(v_i,v).
\end{equation}

The regional consistency loss is then
\begin{equation}
\begin{aligned}
    \mathcal L_{\mathrm{region}}
    =
    \frac{1}{2g}
    \sum_{i=1}^{g}
    \Bigg[
    \frac{1}{g-1}
    \sum_{\substack{v\in\mathcal G\\v\neq v_i}}
    c_{v_i,v}
    +\frac{1}{\ell}
    \sum_{v\in\mathcal V_{\mathrm{loc}}}
    c_{v_i,v}
    \Bigg].
\end{aligned}
\label{eq:region_loss}
\end{equation}
For simplicity, we assume all view pairs have nonempty patch selections within their shared region. In practice, pairs with empty selections are excluded, and each term is normalised by the number of valid pairs.

The overall training objective combines the image-level, masked-patch, and regional consistency losses:

\begin{equation}
    \mathcal L
    =
    \mathcal L_{\mathrm{cls}}
    +
    \mathcal L_{\mathrm{patch}}
    +
    \lambda\mathcal L_{\mathrm{region}}
    \label{eq:full_objective}
\end{equation}

where $\lambda$ controls the contribution of the regional consistency loss.

%% file: sec/4_experiments.tex
\section{Experiments}
\label{section:experiments}

In this section, we evaluate the effectiveness of \methodname{} in learning transferable dense and global visual representations.
We first describe the training setup in Sec.~\ref{sec:experimental_setup} and compare \methodname{} against prior SSL methods in Sec.~\ref{sec:main_results}. We then examine the regional objective through ablations (Sec.~\ref{sec:ablations}) and feature visualisations (Sec.~\ref{sec:visualizations}). Appendix~\ref{sec:additional_experiments} adds transferability and representation analyses.

\subsection{Experimental Setup}
\label{sec:experimental_setup}
We initialise ViT-S/16, ViT-B/16, and ViT-L/16 from the released iBOT~\cite{zhou2022ibot} checkpoints and pretrain them on ImageNet-1K~\cite{deng2009imagenet} for 200, 100, and 40 epochs, respectively. Each image supplies two $224\times224$ global views and ten $96\times96$ local views, and the regional consistency loss $\mathcal L_{\mathrm{region}}$ (Eq.~\ref{eq:region_loss}) is applied to both with weight $\lambda=0.4$ (Sec.~\ref{sec:objective}). Optimisation uses AdamW with a cosine learning-rate schedule. Unless stated otherwise, we report ViT-S/16; Appendix~\ref{sec:pretraining_details} gives the full details.

Starting from iBOT keeps each run within our compute budget and lets us study many variants of the regional loss; Appendix~\ref{sec:from_scratch} reports \methodname{} trained from scratch. However, the gains of \methodname{} could then partly reflect the additional epochs rather than the regional loss. To separate the two, we train a control, iBOT (+200 ep.), that continues iBOT for the same 200 epochs and hyperparameters with only its original objectives (full comparison in Appendix~\ref{sec:fair_comparison}).

\subsection{Main Results}
\label{sec:main_results}

We evaluate \methodname{} across a diverse range of dense and image-level tasks, including semantic segmentation, clustering, image classification, multi-label recognition, and object and region retrieval, covering both dense and global representation quality. Methods marked with an asterisk ($^*$) and shown in grey are pretrained on LVD-142M and included for reference rather than direct comparison.

 \noindent \textbf{Semantic segmentation.}
Regional consistency encourages patches with the same content to share features. We measure this with the CAPI~\cite{darcet2025capi} $k$-NN and linear protocols on ADE20K~\cite{zhou2017ade20k}, PASCAL VOC~\cite{everingham2010voc}, and Cityscapes~\cite{cordts2016cityscapes} (mIoU, Table~\ref{tab:semantic_segmentation}). \methodname{} outperforms all methods pretrained on ImageNet-1K in nearly every reported setting.
The gains are largest under $k$-NN, which has no learned parameters (e.g.\ $+14.6$ VOC mIoU over iBOT~\cite{zhou2022ibot} with ViT-S/16), showing that the patch features are already semantically organised before any task-specific training. Notably, despite using over 100 times less pretraining data, \methodname{} also outperforms CAPI \cite{darcet2025capi} in most settings and DINOv2 \cite{oquab2024dinov2} on VOC, both pretrained on LVD-142M.

\begin{table}[t]
\centering
\caption{
\textbf{Semantic segmentation.}
mIoU (\%) with frozen backbones under the CAPI~\cite{darcet2025capi} $k$-NN and linear protocols. We highlight the \textbf{best} and \underline{second-best}. $^{*}$Pretrained on LVD-142M~\cite{oquab2024dinov2}.
}
\label{tab:semantic_segmentation}

\footnotesize
\setlength{\tabcolsep}{2.2pt}
\renewcommand{\arraystretch}{1.0}

\begin{adjustbox}{max width=\columnwidth}
\begin{tabular}{lcccccc}
\toprule
& \multicolumn{2}{c}{ADE20K~\cite{zhou2017ade20k}}
& \multicolumn{2}{c}{PASCAL VOC~\cite{everingham2010voc}}
& \multicolumn{2}{c}{Cityscapes~\cite{cordts2016cityscapes}} \\
\cmidrule(lr){2-3}
\cmidrule(lr){4-5}
\cmidrule(lr){6-7}
Model
& $k$-NN & Linear
& $k$-NN & Linear
& $k$-NN & Linear \\
\midrule

& \multicolumn{6}{c}{\textit{ViT-Small}} \\
\cmidrule(l){2-7}
SiT~\cite{atito2021sit} & 17.6 & 22.3 & 37.6 & 47.1 & 31.4 & 35.6 \\
DINO~\cite{caron2021dino} & 17.5 & 20.9 & 37.2 & 42.3 & 30.9 & 35.4 \\
iBOT~\cite{zhou2022ibot} & 22.0 & 27.0 & 49.7 & 58.0 & 33.5 & 38.0 \\
iBOT (+200 ep.) & 22.6 & 27.7 & 51.0 & 59.3 & 32.9 & 37.9 \\
CrOC~\cite{stegmuller2023croc} & 19.6 & 24.2 & 49.0 & 58.5 & 29.1 & 34.6 \\
CrIBo~\cite{lebailly2024cribo} & 24.9 & 27.4 & \underline{60.9} & 66.0 & 33.3 & 37.9 \\
CG-SSL~\cite{atito2025cgssl} & \underline{25.8} & \underline{28.9} & 60.8 & \underline{66.7} & \underline{34.2} & \underline{39.1} \\
\rowcolor{blue}
\textbf{\methodname{}} & \textbf{29.2} & \textbf{31.2} & \textbf{64.3} & \textbf{68.8} & \textbf{35.0} & \textbf{39.8} \\
\textcolor{gray}{CAPI$^{*}$}~\cite{darcet2025capi} & \textcolor{gray}{20.6} & \textcolor{gray}{25.8} & \textcolor{gray}{50.5} & \textcolor{gray}{59.4} & \textcolor{gray}{31.3} & \textcolor{gray}{36.5} \\

\midrule
& \multicolumn{6}{c}{\textit{ViT-Base}} \\
\cmidrule(l){2-7}
MAE~\cite{he2022mae} & 18.7 & 25.6 & 44.2 & 56.5 & 32.2 & 38.1 \\
Data2Vec 2.0~\cite{baevski2022data2vec2} & 18.4 & 22.2 & 46.9 & 48.4 & 28.3 & 35.0 \\
iBOT~\cite{zhou2022ibot} & 27.0 & \underline{31.8} & 56.6 & 67.1 & \underline{35.9} & 39.6 \\
CrIBo~\cite{lebailly2024cribo} & 26.9 & 30.0 & 61.2 & 68.1 & 33.7 & 38.4 \\
CG-SSL~\cite{atito2025cgssl} & \underline{28.3} & 31.5 & \underline{62.7} & \underline{69.0} & 35.2 & \underline{40.0} \\
\rowcolor{blue}
\textbf{\methodname{}} & \textbf{31.0} & \textbf{33.9} & \textbf{67.9} & \textbf{72.1} & \textbf{37.8} & \textbf{41.6} \\
\textcolor{gray}{CAPI$^{*}$}~\cite{darcet2025capi} & \textcolor{gray}{28.3} & \textcolor{gray}{33.6} & \textcolor{gray}{61.8} & \textcolor{gray}{70.3} & \textcolor{gray}{36.9} & \textcolor{gray}{41.9} \\
\midrule
& \multicolumn{6}{c}{\textit{ViT-Large}} \\
\cmidrule(l){2-7}
MAE~\cite{he2022mae} & 21.7 & 28.0 & 46.2 & 55.9 & 34.9 & 39.6 \\
Data2Vec 2.0~\cite{baevski2022data2vec2} & 23.6 & 27.0 & 49.8 & 50.7 & 33.1 & 38.6 \\
iBOT~\cite{zhou2022ibot} & 27.2 & 32.6 & 58.6 & 67.8 & \underline{36.7} & 41.9 \\
CAPI~\cite{darcet2025capi} & 29.2 & 34.4 & 60.7 & 69.7 & 35.6 & 41.7 \\
CG-SSL~\cite{atito2025cgssl} & \underline{30.5} & \textbf{35.6} & \underline{64.0} & \underline{70.6} & 36.3 & \textbf{45.5} \\
\rowcolor{blue}
\textbf{\methodname{}} & \textbf{32.7} & \underline{35.2} & \textbf{68.2} & \textbf{73.9} & \textbf{37.9} & \underline{42.1} \\
\textcolor{gray}{CAPI$^{*}$}~\cite{darcet2025capi} & \textcolor{gray}{32.1} & \textcolor{gray}{37.2} & \textcolor{gray}{63.8} & \textcolor{gray}{72.7} & \textcolor{gray}{38.9} & \textcolor{gray}{44.3} \\
\textcolor{gray}{DINOv2$^{*}$}~\cite{oquab2024dinov2} & \textcolor{gray}{34.0} & \textcolor{gray}{39.0} & \textcolor{gray}{63.0} & \textcolor{gray}{72.8} & \textcolor{gray}{42.0} & \textcolor{gray}{46.8} \\
\bottomrule
\end{tabular}
\end{adjustbox}

\end{table}

\begin{table}[t]
\centering
\caption{
\textbf{Frozen clustering evaluation.}
mIoU (\%) after clustering frozen ViT-S patch features into one cluster per class (GT) or into 300 and 500 clusters. We highlight the \textbf{best} and \underline{second-best}.
}
\label{tab:neco_clustering}

\footnotesize
\renewcommand{\arraystretch}{1.0}
\setlength{\tabcolsep}{4pt}

\begin{adjustbox}{max width=\columnwidth}
\begin{tabular}{p{1.6cm}x{0.8cm}x{0.8cm}x{0.8cm}x{0.8cm}x{0.8cm}x{0.8cm}}
\toprule
&
\multicolumn{3}{c}{PASCAL VOC~\cite{everingham2010voc}} &
\multicolumn{3}{c}{COCO-Things~\cite{lin2014coco}} \\
\cmidrule(lr){2-4}
\cmidrule(lr){5-7}
Method
& GT & 300 & 500
& GT & 300 & 500 \\
\midrule

DINO~\cite{caron2021dino} & 4.3 & 13.9 & 17.3 & 5.4 & 18.8 & 19.2 \\

iBOT~\cite{zhou2022ibot} & 5.8 & 27.8 & 33.0 & 9.4 & 29.7 & 34.0 \\

CrOC~\cite{stegmuller2023croc} & 3.4 & 16.4 & 20.0 & 4.9 & 14.7 & 18.1 \\

TimeT~\cite{salehi2023timet} & 12.2 & 43.6 & 46.2 & \underline{17.5} & 42.7 & 44.6 \\

CrIBo~\cite{lebailly2024cribo} & \underline{18.3} & \underline{51.3} & \underline{54.5} & 14.5 & \underline{46.0} & \underline{48.3} \\

\rowcolor{blue}
\textbf{\methodname{}} & \textbf{18.7} & \textbf{61.7} & \textbf{64.8} & \textbf{24.6} & \textbf{59.3} & \textbf{61.6} \\

\textcolor{gray}{DINOv2$^{*}$~\cite{oquab2024dinov2}} & \textcolor{gray}{17.1} & \textcolor{gray}{{53.5}} & \textcolor{gray}{{58.2}} & \textcolor{gray}{{19.7}} & \textcolor{gray}{{51.6}} & \textcolor{gray}{{53.8}} \\

\bottomrule
\end{tabular}
\end{adjustbox}
\end{table}

 \noindent \textbf{Frozen clustering evaluation.}
Clustering measures whether frozen patch features group by semantic class without any supervision~\cite{ziegler2022leopart,salehi2023timet}. We apply $k$-means to the patch features of all VOC and COCO-Things~\cite{lin2014coco} images and report mIoU after matching the clusters to the ground-truth classes (Table~\ref{tab:neco_clustering}). Setting $K$ to the number of classes requires each class to form a single cluster, whereas overclustering with $K=300$ and $500$ allows a class to split into finer groups, such as object parts, provided each group stays within that class. \methodname{} achieves the best mIoU in all six settings, and its margin over the strongest competitor grows from $0.4$--$7.1$ points with one cluster per class to $10.3$--$13.3$ points under overclustering. Hence, not only do whole classes separate, but small neighbourhoods of its features also remain within a single class. For visualisations, please refer to Appendix~\ref{sec:visual_examples}.

\begin{table}[t]
\centering
\caption{
\textbf{Multi-class classification.}
Top-1 accuracy (\%) of $k$-NN and linear probes on frozen ViT-S/16 features, with the fraction of the training set used by each probe. 
}
\label{tab:imagenet_classification}

\footnotesize
\setlength{\tabcolsep}{4.0pt}
\renewcommand{\arraystretch}{1.0}

\begin{adjustbox}{max width=\columnwidth}
\begin{tabular}{p{2.2cm}x{0.9cm}x{0.9cm}x{0.9cm}x{1.5cm}}
\toprule
& \multicolumn{3}{c}{$k$-NN}
& Linear \\
\cmidrule(lr){2-4}
Method
& 1\% & 10\% & 100\% & 100\% \\
\midrule

iBOT~\cite{zhou2022ibot} & \underline{62.3} & \underline{70.1} & \underline{74.9} & \underline{78.0} \\

iBOT (+200 ep.) & 61.8 & 69.9 & 74.8 & -- \\

CrOC~\cite{stegmuller2023croc} & 49.0 & 58.0 & 66.5 & 71.4 \\

CrIBo~\cite{lebailly2024cribo} & 58.1 & 64.4 & 70.7 & 74.9 \\

CG-SSL~\cite{atito2025cgssl} & -- & -- & 74.7 & 77.7 \\

\rowcolor{blue}
\textbf{\methodname{}} & \textbf{63.3} & \textbf{70.3} & \textbf{75.1} & \textbf{78.6} \\

\textcolor{gray}{CAPI$^{*}$~\cite{darcet2025capi}} & \textcolor{gray}{--} & \textcolor{gray}{--} & \textcolor{gray}{--} & \textcolor{gray}{71.5} \\

\bottomrule
\end{tabular}
\end{adjustbox}

\end{table}

 \noindent \textbf{Multi-class classification.}
Dense self-supervised objectives often improve patch features at the expense of the global representation. We test this trade-off on ImageNet-1K~\cite{deng2009imagenet} with a weighted $k$-NN classifier on the $[\mathrm{CLS}]$ token, using $1\%$, $10\%$ or $100\%$ of the training set as reference bank, and with a linear probe trained on the full training set (Table~\ref{tab:imagenet_classification}). The dense methods CrOC~\cite{stegmuller2023croc}, CrIBo~\cite{lebailly2024cribo} and CG-SSL~\cite{atito2025cgssl} all fall below iBOT~\cite{zhou2022ibot}, whereas \methodname{} maintains its accuracy in every setting. The regional loss thus strengthens the patch features without compromising image-level recognition.

 \noindent \textbf{Multi-label classification.}
Multi-label recognition requires a single image representation to capture every object in a scene, not only the dominant one. We train linear probes on frozen features at $224\times224$ on VOC~\cite{everingham2010voc}, COCO~\cite{lin2014coco} and Visual Genome~\cite{krishna2017visualgenome}, and in low-shot regimes on VOC with one, two or five training images per class (Table~\ref{tab:multilabel_classification}). \methodname{} achieves the highest mAP in all six settings. Its advantage over iBOT is largest when training images are scarce ($+4.3$ and $+4.0$ mAP with one and two images per class) and, with full data, on COCO ($+1.5$), whose images typically contain several objects. This suggests that supervising the composition of regions also benefits the image-level features of scenes with multiple objects.

\begin{table}[t]
\centering
\caption{
\textbf{Multi-label classification.}
mAP (\%) of linear probes on frozen ViT-S/16 features, trained with one, two or five images per class or the full training set.
}
\label{tab:multilabel_classification}

\footnotesize
\setlength{\tabcolsep}{2.0pt}
\renewcommand{\arraystretch}{1.0}

\begin{adjustbox}{max width=\columnwidth}
\begin{tabular}{p{1.7cm}cccccx{1.2cm}}
\toprule
& \multicolumn{4}{c}{PASCAL VOC~\cite{everingham2010voc}}
& COCO~\cite{lin2014coco}
& VG~\cite{krishna2017visualgenome} \\
\cmidrule(lr){2-5}
Method
& 1 Img & 2 Imgs & 5 Imgs & Full
& Full & Full \\
\midrule

iBOT~\cite{zhou2022ibot} & \underline{47.0} & \underline{56.6} & 73.0 & \underline{89.2} & \underline{69.6} & \underline{25.8} \\

CrOC~\cite{stegmuller2023croc} & 44.2 & 54.3 & 71.8 & 86.6 & 62.5 & 22.2 \\

CrIBo~\cite{lebailly2024cribo} & 46.6 & 55.5 & \underline{74.3} & \underline{89.2} & 67.7 & 25.4 \\

\rowcolor{blue}
\textbf{\methodname{}} & \textbf{51.3} & \textbf{60.6} & \textbf{75.1} & \textbf{89.8} & \textbf{71.1} & \textbf{26.8} \\

\bottomrule
\end{tabular}
\end{adjustbox}

\end{table}

\begin{table}[t]
\centering
\caption{
\textbf{Object and region retrieval.}
mAP, Precision@50 (P@50) and Recall@1 (R@1) (\%) of COCO objects and VOC regions retrieved with mean-pooled frozen ViT-S/16 features.
}
\label{tab:object_region_retrieval}\label{tab:semantic_region_retrieval}

\footnotesize
\setlength{\tabcolsep}{4pt}
\renewcommand{\arraystretch}{1.0}

\begin{tabular}{p{1.6cm}x{1.1cm}x{1.1cm}x{1.1cm}x{1.1cm}}
\toprule
& \multicolumn{2}{c}{Objects (COCO)}
& \multicolumn{2}{c}{Regions (VOC)} \\
\cmidrule(lr){2-3}\cmidrule(l){4-5}
Method & mAP & P@50 & mAP & R@1 \\
\midrule
iBOT~\cite{zhou2022ibot} & 20.9 & 32.1 & 49.3 & 82.3 \\
\rowcolor{blue}
\textbf{\methodname{}} & \textbf{38.4} & \textbf{48.4} & \textbf{63.0} & \textbf{86.0} \\
\bottomrule
\end{tabular}
\end{table}

 \noindent \textbf{Object and region retrieval.}
We evaluate whether pooled patch features capture objects and semantic regions. Following~\cite{shlapen2024region}, we mean-pool frozen backbone features within ground-truth COCO query masks and SAM~\cite{kirillov2023sam} gallery regions, ranking images by their best-matching region (Table~\ref{tab:object_region_retrieval}). \methodname{} nearly doubles iBOT's mAP, from $20.9$ to $38.4$, and raises Precision@50 from $32.1$ to $48.4$. Although the regional loss acts on patch predictions, the gains carry over to pooled backbone descriptors.

\begin{figure}[t]
\centering
\footnotesize
\setlength{\tabcolsep}{2.3pt}
\begin{minipage}[b]{0.44\columnwidth}
\centering
\input{img/variance_src/variance_bars}\\[1pt]
(a) Feature variance
\end{minipage}\hfill 
\begin{minipage}[b]{0.55\columnwidth}
\centering
\begin{tabular}{lccc}
\toprule
& Corr. $\uparrow$ & \multicolumn{2}{c}{Depth $\downarrow$} \\
\cmidrule(lr){2-2}\cmidrule(l){3-4}
Model & SPair & NYU & KITTI \\
\midrule
iBOT~\cite{zhou2022ibot} & \textbf{29.6} & 0.513 & 3.605 \\
iBOT (+200 ep.) & 29.2 & \textbf{0.500} & \textbf{3.574} \\
\rowcolor{blue}
\textbf{\methodname{}} & 26.7 & 0.502 & 3.622 \\
\bottomrule
\end{tabular}\\[14pt]
(b) Benchmarks
\end{minipage}
\caption{\textbf{Feature analysis} with frozen ViT-S/16 features. (a) Variance of the patch features of VOC validation objects (mean squared cosine distance to the mean) with standard deviation. (b) SPair-71k PCK@0.1 and depth RMSE.}
\label{fig:feature_analysis}\label{tab:beyond}
\end{figure}

 \noindent \textbf{Feature analysis.}
To understand what the regional loss changes in the features, we measure the variance of the patch features inside the annotated objects of the VOC validation set (Figure~\ref{fig:feature_analysis}a). \methodname{} redistributes this variance rather than removing it. This follows from the objective: two views agree on the composition of a shared region only if patches of the same content select the same prototypes, so the patches of an object concentrate on a few prototypes and their variance falls. As the prototypes are shared across images, objects of a class select the same ones and different classes select different ones, which increases the variance between classes. Segmentation, clustering and retrieval gain from compact objects and well-separated classes. Tasks that instead read the fine variation inside an object, such as patch-to-patch correspondence and depth estimation (Figure~\ref{fig:feature_analysis}b; protocols of Probe3D~\cite{elbanani2024probe3d}), do not draw on this organisation: depth is unchanged and correspondence slightly degrades.

\begin{figure*}[t!]
\begin{minipage}[t]{\columnwidth}\vspace{0pt}
\captionof{table}{\textbf{Ablation study.} VOC $k$-NN and linear (lin.) mIoU and IN1k $k$-NN accuracy (\%, 10\% reference bank); the reference is shown in blue. In (a), $\mathbf q_p$ and $\mathbf f_p$ are the prediction and backbone embedding of patch $p$.}
\label{tab:ablations}
\centering
\footnotesize
\setlength{\tabcolsep}{1.75pt}
\makeatletter
\newcommand{\ctedgeL}{\gdef\CT@cell@color{\@tempdimb\z@\global\let\CT@cell@color\relax}}
\newcommand{\ctedgeR}{\gdef\CT@cell@color{\@tempdimc\z@\global\let\CT@cell@color\relax}}
\makeatother
\newlength{\ablLa}\settowidth{\ablLa}{Global \& local}
\newlength{\ablLb}\settowidth{\ablLb}{+ prototype std.}
\begin{tabular}[t]{@{}>{\raggedright\arraybackslash}p{\ablLa}ccc@{}}
\multicolumn{4}{c}{(a) Embeddings vs.\ prototypes.} \\[2pt]
\toprule
& \multicolumn{2}{c}{VOC} & IN1k \\ \cmidrule(lr){2-3}\cmidrule(l){4-4}
Region & $k$-NN & lin. & $k$-NN \\
\midrule
\rowcolor{blue}
\ctedgeL $\mathrm{mean}(\mathbf q_p)$ & \textbf{61.8} & \textbf{66.7} & \ctedgeR 69.1 \\
$\mathrm{mean}(\mathbf f_p)$ & 49.8 & 55.6 & 69.4 \\
$\mathbf q(\mathrm{mean}(\mathbf f_p))$ & 59.9 & 64.6 & \textbf{69.5} \\
\bottomrule
\end{tabular}\hfill
\begin{tabular}[t]{@{}>{\raggedright\arraybackslash}p{\ablLb}ccc@{}}
\multicolumn{4}{c}{(b) Region statistic.} \\[2pt]
\toprule
& \multicolumn{2}{c}{VOC} & IN1k \\ \cmidrule(lr){2-3}\cmidrule(l){4-4}
Configuration & $k$-NN & lin. & $k$-NN \\
\midrule
\rowcolor{blue}
\ctedgeL \textit{Mean} & \textbf{61.8} & \textbf{66.7} &  \textbf{69.1} \\
+ prototype std. & 61.6 & 66.6 & \textbf{69.1} \\
+ covariance & 61.6 & 66.5 & \textbf{69.1} \\
\bottomrule
\end{tabular}
\par\vspace{6pt}
\raisebox{3pt}[\height][\depth]{%
\begin{tabular}[t]{@{}>{\raggedright\arraybackslash}p{\ablLa}ccc@{}}
\multicolumn{4}{c}{(c) View and patch selection.} \\[2pt]
\toprule
& \multicolumn{2}{c}{VOC} & IN1k \\ \cmidrule(lr){2-3}\cmidrule(l){4-4}
Configuration & $k$-NN & lin. & $k$-NN \\
\midrule
\rowcolor{blue}
\ctedgeL \textit{Global} & 61.8 & 66.7 & \ctedgeR 69.1 \\
Local & 58.0 & 65.0 & 69.8 \\
Global \& local & \textbf{62.4} & \textbf{67.1} & \textbf{70.0} \\
\midrule
Aligned grid & 60.8 & 65.0 & 68.9 \\
Visible only & 61.6 & 66.9 & 69.2 \\
\bottomrule
\end{tabular}}\hfill
\begin{tabular}[t]{@{}>{\raggedright\arraybackslash}p{\ablLb}ccc@{}}
\multicolumn{4}{c}{(d) Other components.} \\[2pt]
\toprule
& \multicolumn{2}{c}{VOC} & IN1k \\ \cmidrule(lr){2-3}\cmidrule(l){4-4}
Configuration & $k$-NN & lin. & $k$-NN \\
\midrule
iBOT++~\cite{cao2026tipsv2} & 55.3 & 63.2 & 67.6 \\
Registers~\cite{darcet2024vision} & 61.7 & 66.7 & 69.0 \\
KoLeo~\cite{sablayrolles2019spreading} & 61.6 & 66.6 & \textbf{70.5} \\
Separate heads & 61.0 & 66.7 & 68.7 \\
Multi-depth & 59.4 & 65.6 & 68.8 \\
\bottomrule
\end{tabular}
\end{minipage}\hfill
\begin{minipage}[t]{\columnwidth}\vspace{0pt}
\centering
\input{img/ablation_src/ablation_charts}
\newsavebox{\ablboxC}\sbox{\ablboxC}{\ablchartC}
\ablchartA\hfill\ablchartB\par\vspace{4pt}\nointerlineskip
\noindent\begin{minipage}[t]{\wd\ablboxC}\vspace{0pt}\usebox{\ablboxC}\end{minipage}\hfill
\begin{minipage}[t]{\dimexpr\columnwidth-\wd\ablboxC-8pt\relax}\vspace{0pt}
\captionsetup{justification=raggedright, skip=0pt}
\caption{\textbf{Shared area, patch coverage and loss weight.} (a) Minimum area shared by a view pair, as a fraction of the image. (b) Minimum fraction of a patch inside the shared region; \textit{weighted} keeps every patch, weighted by this fraction. (c) Weight $\lambda$ of the regional loss. Filled markers denote the reference.}
\label{fig:region_selection}
\end{minipage}
\end{minipage}
\end{figure*}

\subsection{Ablations}
\label{sec:ablations}

We ablate the regional loss with ViT-S/16, trained on ImageNet-1K~\cite{deng2009imagenet} for 50 epochs on four GPUs with a total batch size of 256. Blue rows in Table~\ref{tab:ablations} and filled markers in Figure~\ref{fig:region_selection} denote the reference setting shared by all ablations, and each variant changes one of its choices. We report $k$-NN and linear mIoU on PASCAL VOC~\cite{everingham2010voc} and $k$-NN accuracy on ImageNet-1K with a $10\%$ reference bank. Further ablations are reported in Appendix~\ref{sec:appx_ablations}.

 \noindent \textbf{Embeddings vs.\ prototypes.}
 Table \ref{tab:ablations}a compares three ways of enforcing regional consistency. Our formulation, $\mathrm{mean}(\mathbf q_p)$, averages patch predictions, while the alternatives average backbone features, either aligning them directly with cosine distance, $\mathrm{mean}(\mathbf f_p)$, or applying the projection head before matching, $\mathbf q(\mathrm{mean}(\mathbf f_p))$. Our formulation outperforms both alternatives on dense tasks while maintaining comparable ImageNet accuracy, highlighting the importance of averaging predictions rather than embeddings to preserve multiple visual concepts within a region.
 

 \noindent \textbf{Region statistic.}
Table~\ref{tab:ablations}b examines whether the mean patch prediction is sufficient to represent a region. We compare the composition (Sec.~\ref{sec:region_representation}) with richer statistics, including the standard deviation of each prototype's predictions across patches and their full covariance, which captures relationships between predictions without patch correspondences. Neither provides meaningful gains over the mean, suggesting that regional composition alone provides sufficient supervision, allowing individual patches to retain distinct predictions while the complementary masked-patch loss encourages confident assignments.

 \noindent \textbf{View and patch selection.}
Table~\ref{tab:ablations}c examines which views and patches should form the compositions. Being an average, a composition needs no one-to-one correspondence and can compare regions of different sizes and scales, an advantage over patch-wise matching. Adding global-to-local pairs (global \& local) exploits this: it gives the best result on all three metrics and is the setting of our main experiments, whereas local pairs alone (local) are not sufficient. Aligning the grids of the two views by restricting crop displacements and flips (aligned grid) lowers $k$-NN mIoU by $1.0$ point, so the composition does not benefit from patch-wise correspondence, and excluding the student's masked patches (visible only) brings no consistent benefit.

 \noindent \textbf{Other components.}
Table~\ref{tab:ablations}d tests whether components of other self-supervised methods complement the regional loss. We add iBOT++~\cite{cao2026tipsv2}, which extends the masked-patch loss to the visible patches; four register tokens~\cite{darcet2024vision}; the KoLeo regulariser~\cite{sablayrolles2019spreading} on the $[\mathrm{CLS}]$ embeddings; separate projection heads for the $[\mathrm{CLS}]$ and patch tokens; or multi-depth supervision, which applies the regional loss at blocks 3, 6, 9 and 12, each with a copy of the head. None improves dense transfer. KoLeo is the only addition that improves image-level accuracy, raising it by $1.4$ points. We nevertheless exclude KoLeo from our main experiments to ensure a fair comparison with the baseline.

\begin{figure*}[t!]
\centering
\input{img/patch_query_marker}

\newlength{\pcalabw}
\newlength{\pcasep}
\newlength{\pcalabelsep}
\newlength{\pcaw}
\newlength{\pcagroupsep}
\newlength{\pcagroupw}

\setlength{\pcalabw}{14pt}

\setlength{\pcasep}{0.7pt}
\setlength{\pcalabelsep}{2pt}
\setlength{\pcagroupsep}{8pt}

\setlength{\pcaw}{
    \dimexpr(\textwidth-\pcalabw-\pcalabelsep-9\pcasep-2\pcagroupsep)/12\relax
}
\setlength{\pcagroupw}{\dimexpr4\pcaw+3\pcasep\relax}
\newcommand{\pcagroupgap}{\hspace{\pcagroupsep}}

\scriptsize
\setlength{\tabcolsep}{0pt}
\renewcommand{\arraystretch}{0}

\mbox{%
\begin{tabular}{@{}>{\centering\arraybackslash}m{\pcalabw}@{\hspace{\pcalabelsep}}
>{\centering\arraybackslash}m{\pcaw}@{\hspace{\pcasep}}
>{\centering\arraybackslash}m{\pcaw}@{\hspace{\pcasep}}
>{\centering\arraybackslash}m{\pcaw}@{\hspace{\pcasep}}
>{\centering\arraybackslash}m{\pcaw}@{}}
\rotatebox{90}{\footnotesize original} &
\includegraphics[width=\pcaw]{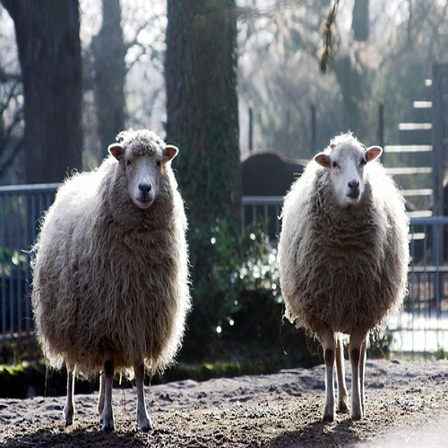} &
\includegraphics[width=\pcaw]{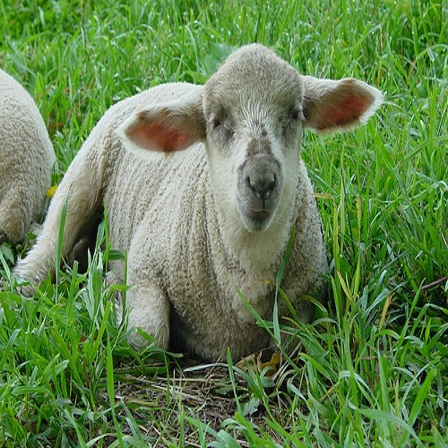} &
\includegraphics[width=\pcaw]{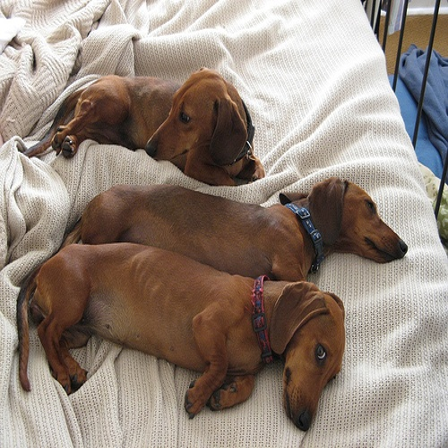} &
\includegraphics[width=\pcaw]{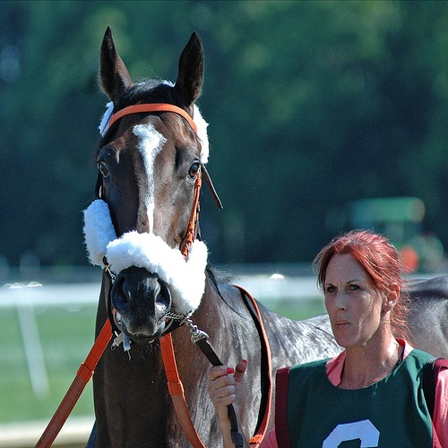}
\\
\noalign{\vspace{0.7pt}}
\rotatebox{90}{\footnotesize iBOT} &
\includegraphics[width=\pcaw]{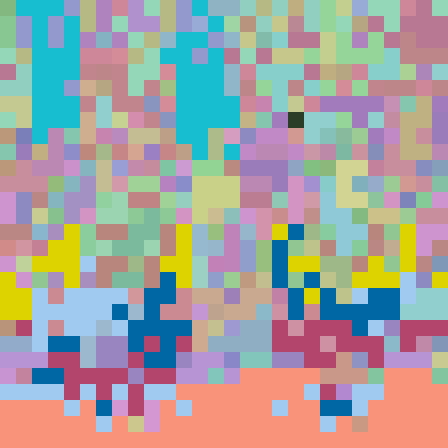} &
\includegraphics[width=\pcaw]{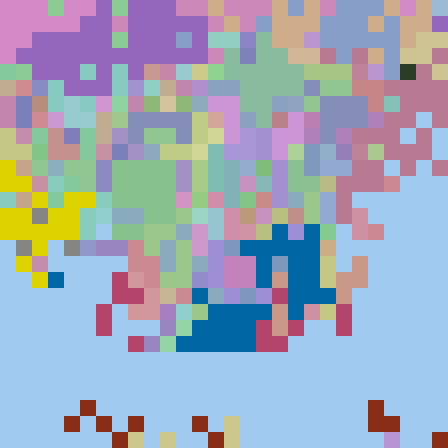} &
\includegraphics[width=\pcaw]{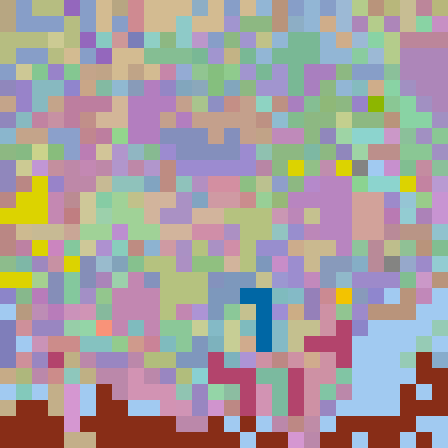} &
\includegraphics[width=\pcaw]{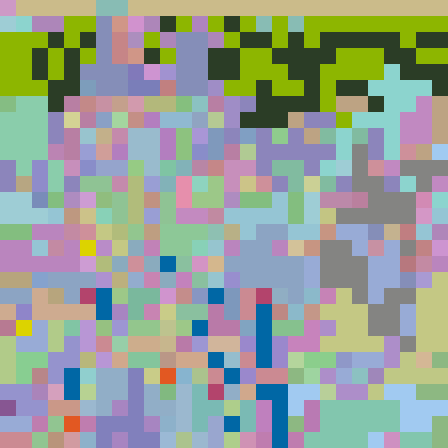}
\\
\noalign{\vspace{0.7pt}}
\rotatebox{90}{\footnotesize \methodname{}} &
\includegraphics[width=\pcaw]{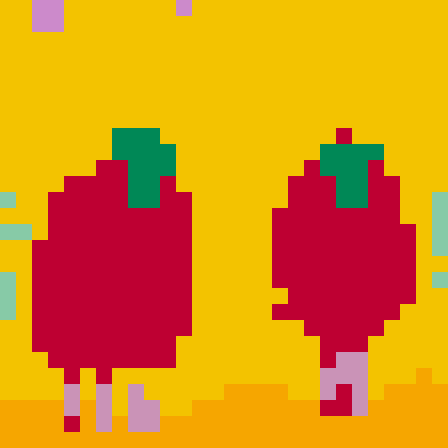} &
\includegraphics[width=\pcaw]{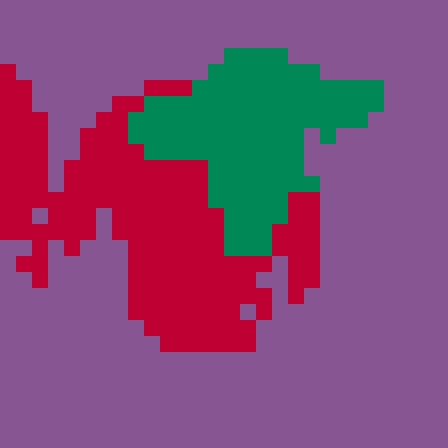} &
\includegraphics[width=\pcaw]{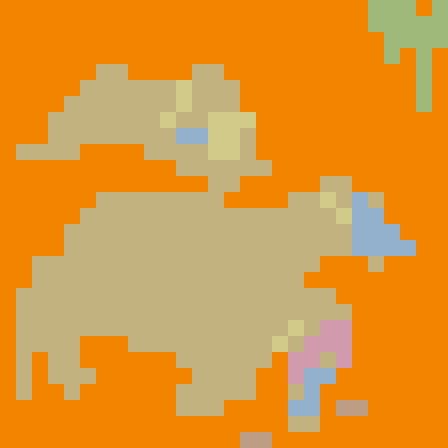} &
\includegraphics[width=\pcaw]{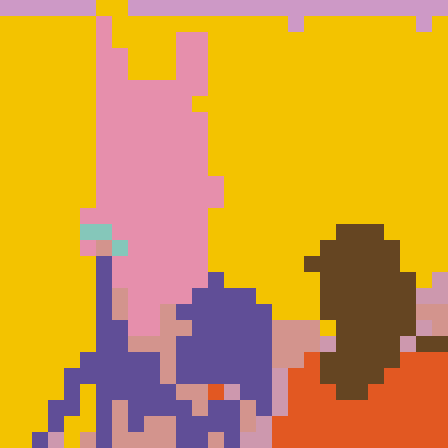}
\\
\end{tabular}%
\pcagroupgap
\begin{tabular}{@{}
>{\centering\arraybackslash}m{\pcaw}@{\hspace{\pcasep}}
>{\centering\arraybackslash}m{\pcaw}@{\hspace{\pcasep}}
>{\centering\arraybackslash}m{\pcaw}@{\hspace{\pcasep}}
>{\centering\arraybackslash}m{\pcaw}@{}}
\includegraphics[width=\pcaw]{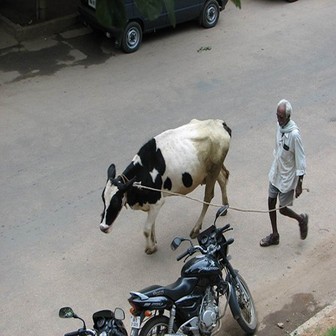} &
\includegraphics[width=\pcaw]{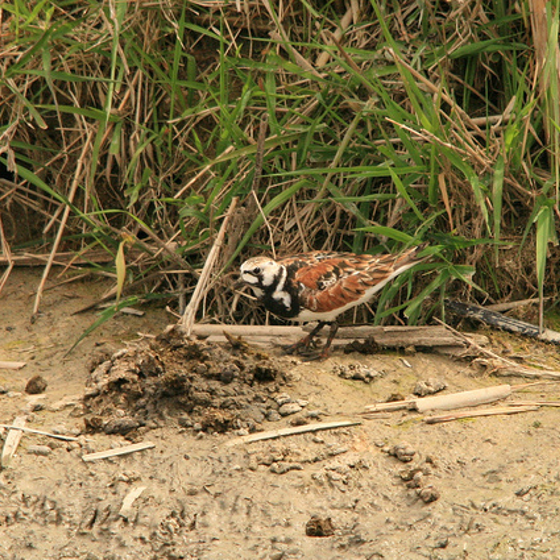} &
\includegraphics[width=\pcaw]{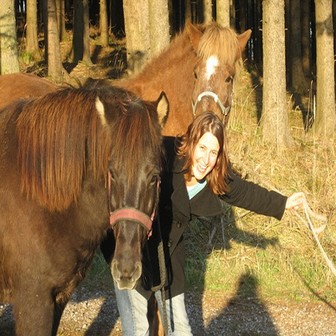} &
\includegraphics[width=\pcaw]{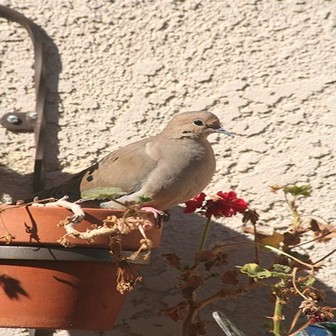}
\\
\noalign{\vspace{0.7pt}}
\includegraphics[width=\pcaw]{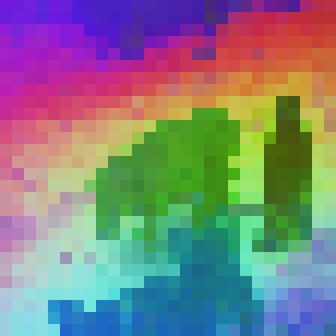} &
\includegraphics[width=\pcaw]{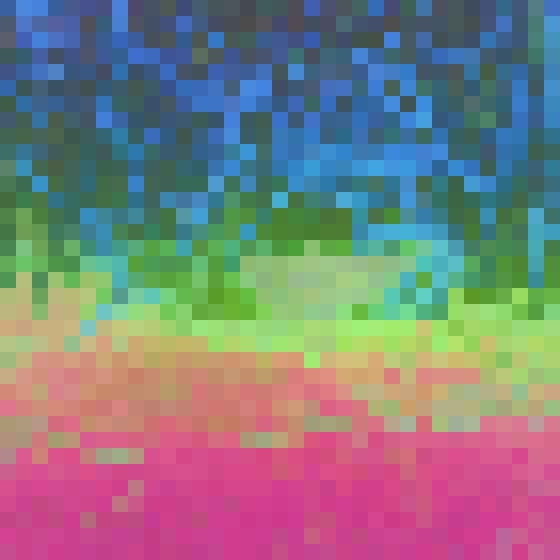} &
\includegraphics[width=\pcaw]{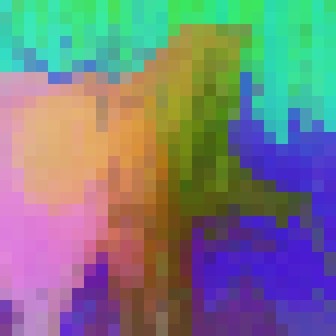} &
\includegraphics[width=\pcaw]{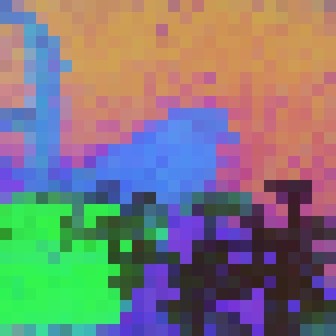}
\\
\noalign{\vspace{0.7pt}}
\includegraphics[width=\pcaw]{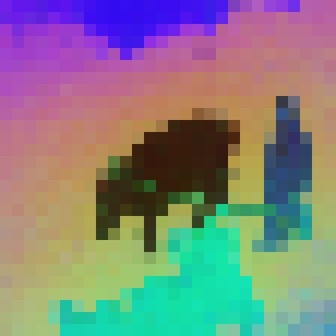} &
\includegraphics[width=\pcaw]{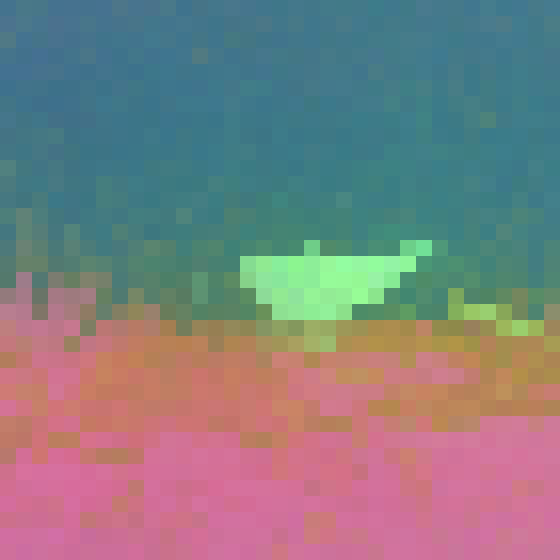} &
\includegraphics[width=\pcaw]{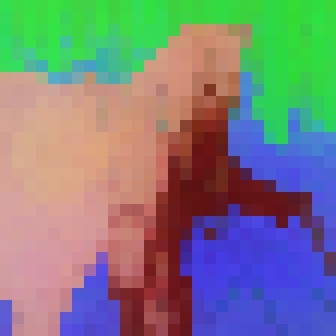} &
\includegraphics[width=\pcaw]{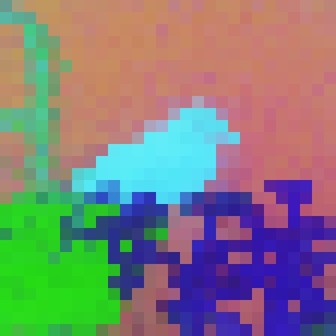}
\\
\end{tabular}%
\pcagroupgap
\begin{tabular}{@{}
>{\centering\arraybackslash}m{\pcaw}@{\hspace{\pcasep}}
>{\centering\arraybackslash}m{\pcaw}@{\hspace{\pcasep}}
>{\centering\arraybackslash}m{\pcaw}@{\hspace{\pcasep}}
>{\centering\arraybackslash}m{\pcaw}@{}}
\includegraphics[width=\pcaw]{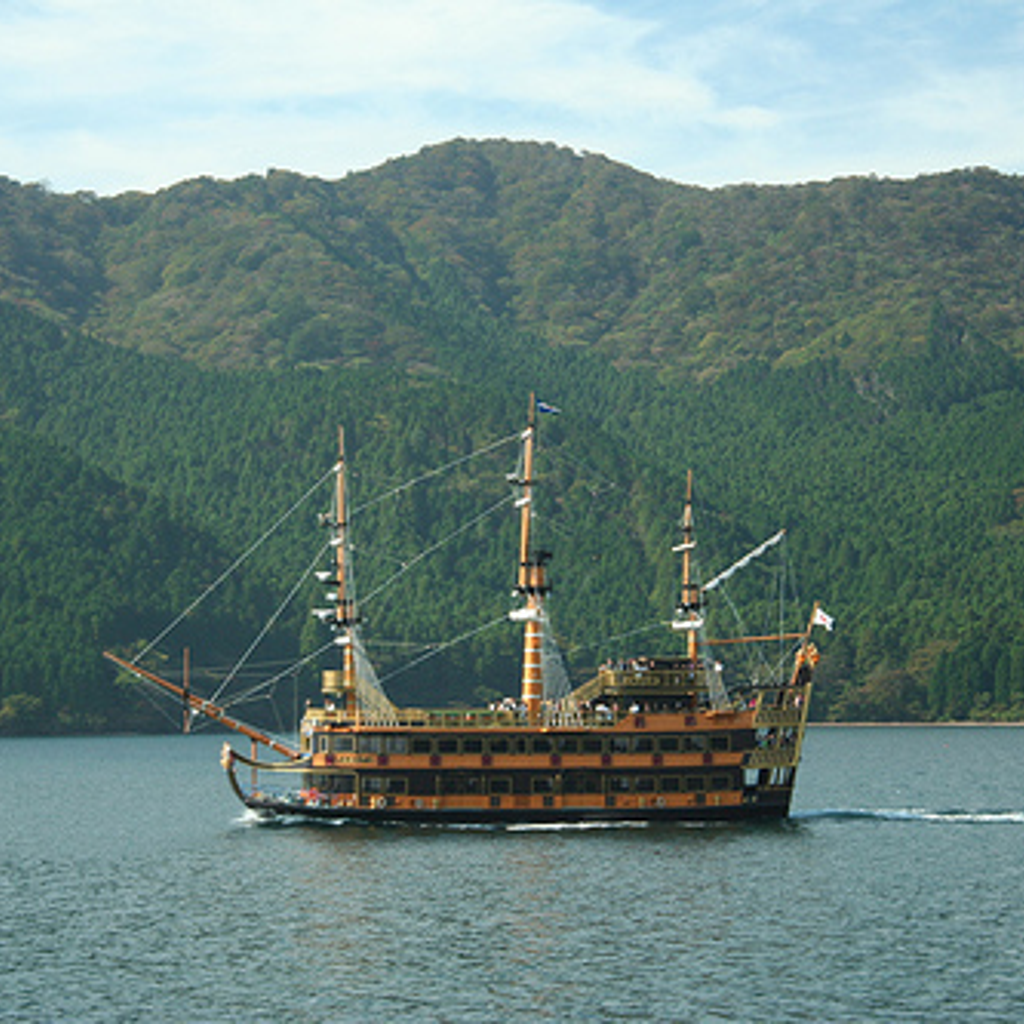} &
\includegraphics[width=\pcaw]{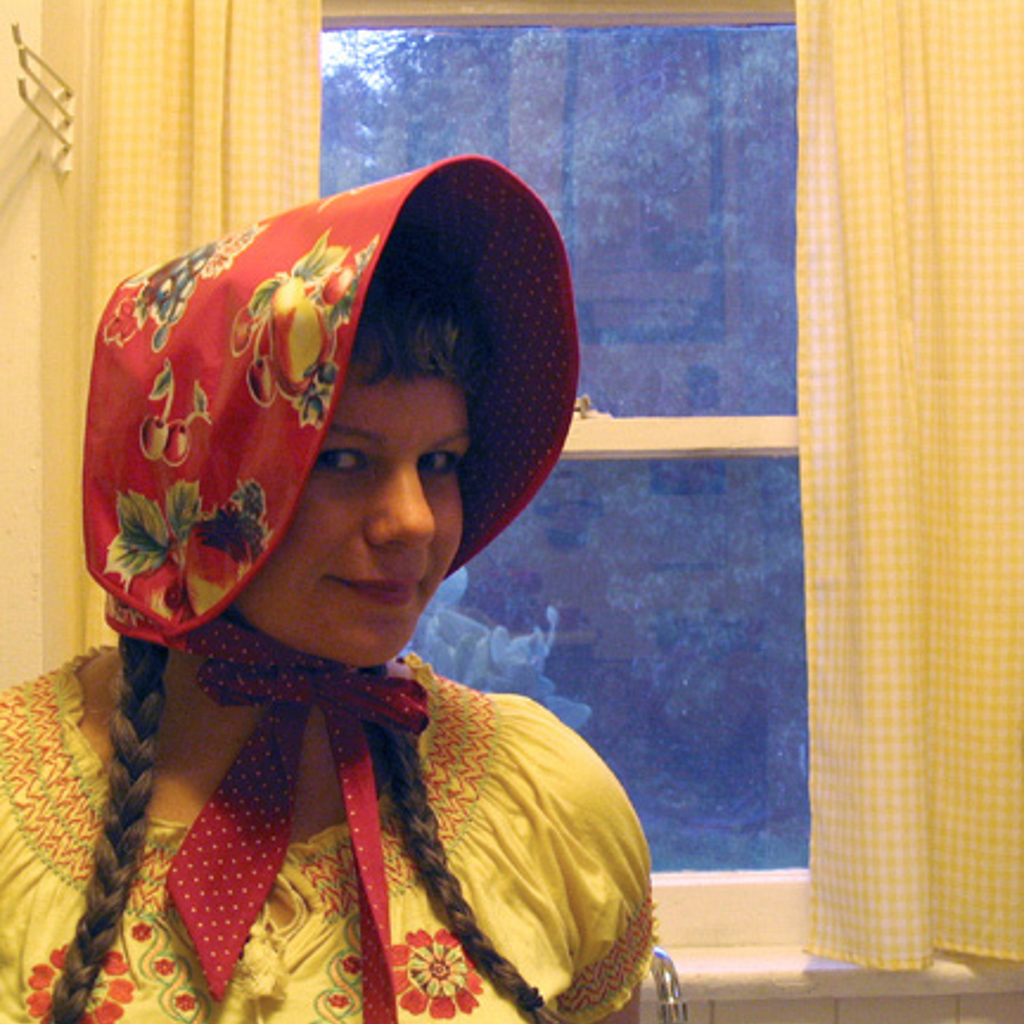} &
\includegraphics[width=\pcaw]{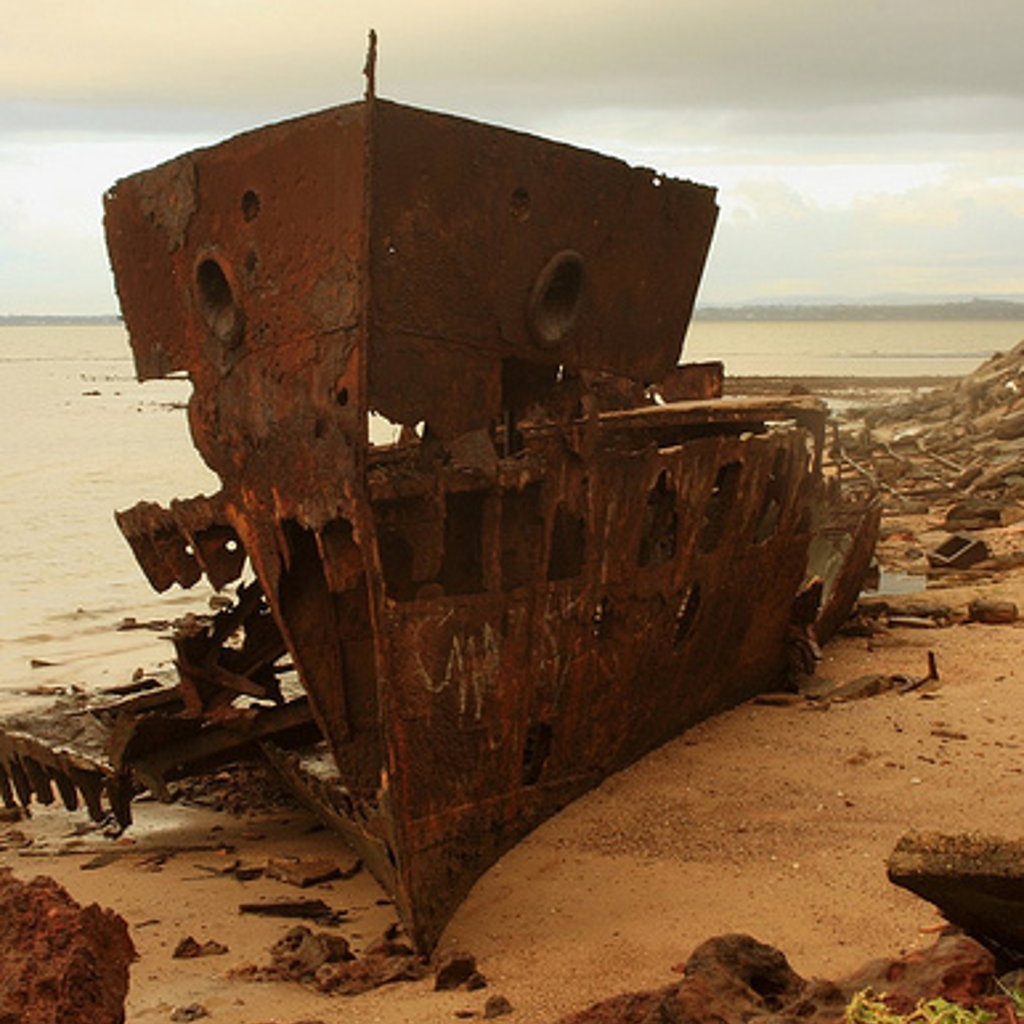} &
\includegraphics[width=\pcaw]{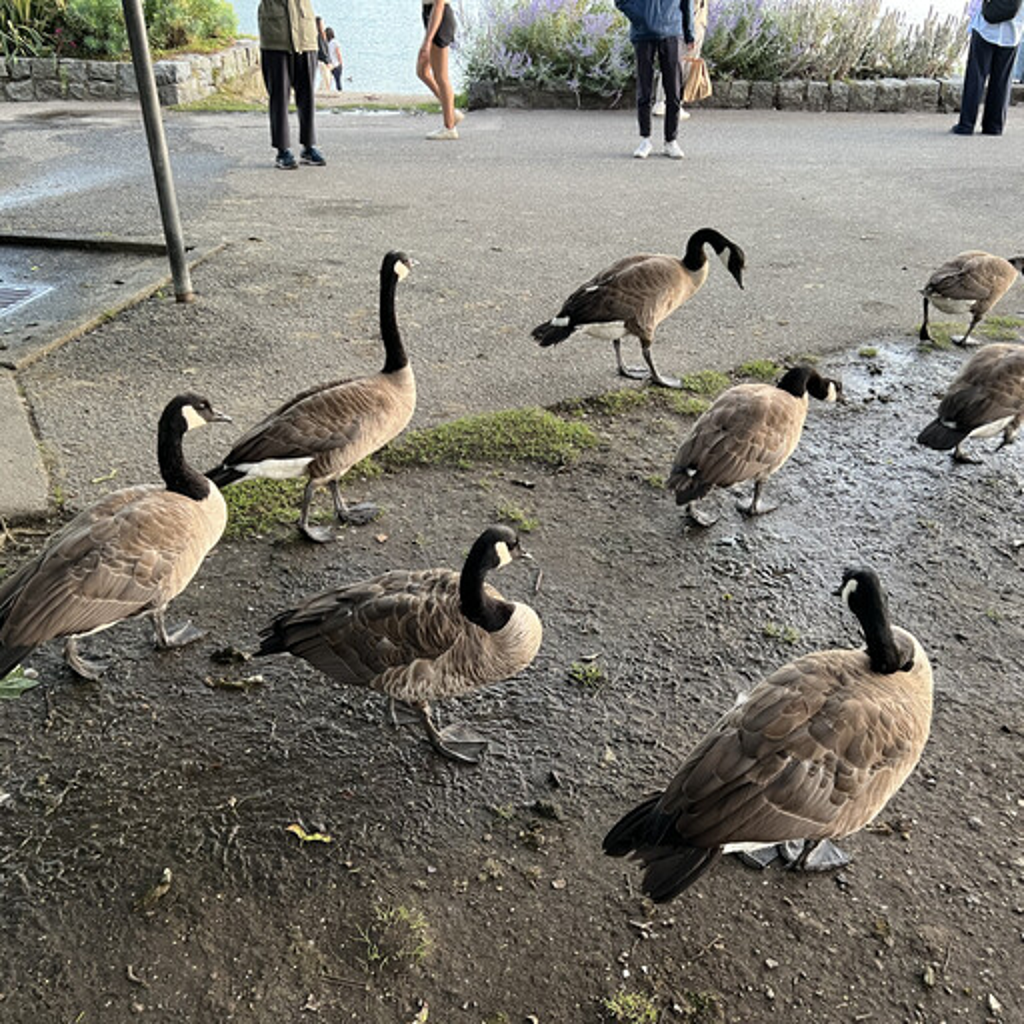}
\\
\noalign{\vspace{0.7pt}}
\patchquerypanel{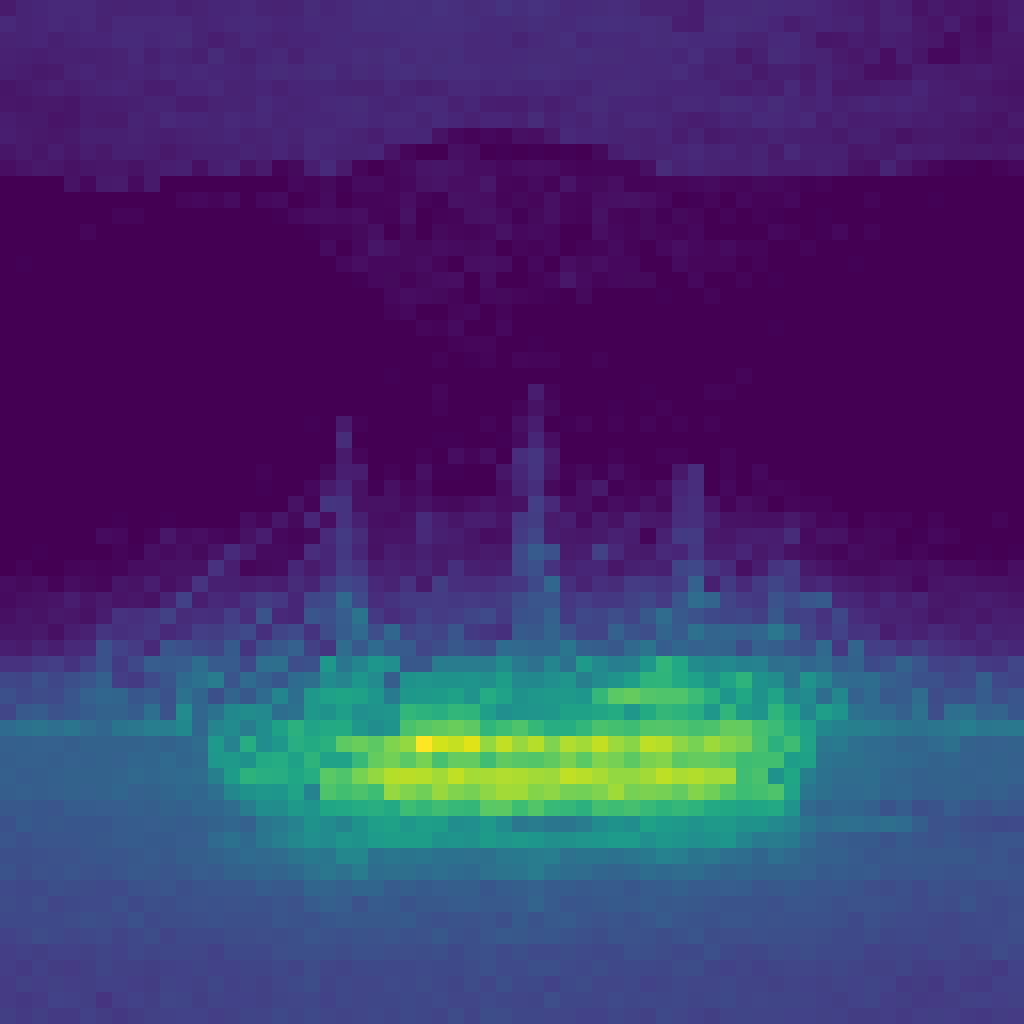}{0.4140625}{0.2734375} &
\patchquerypanel{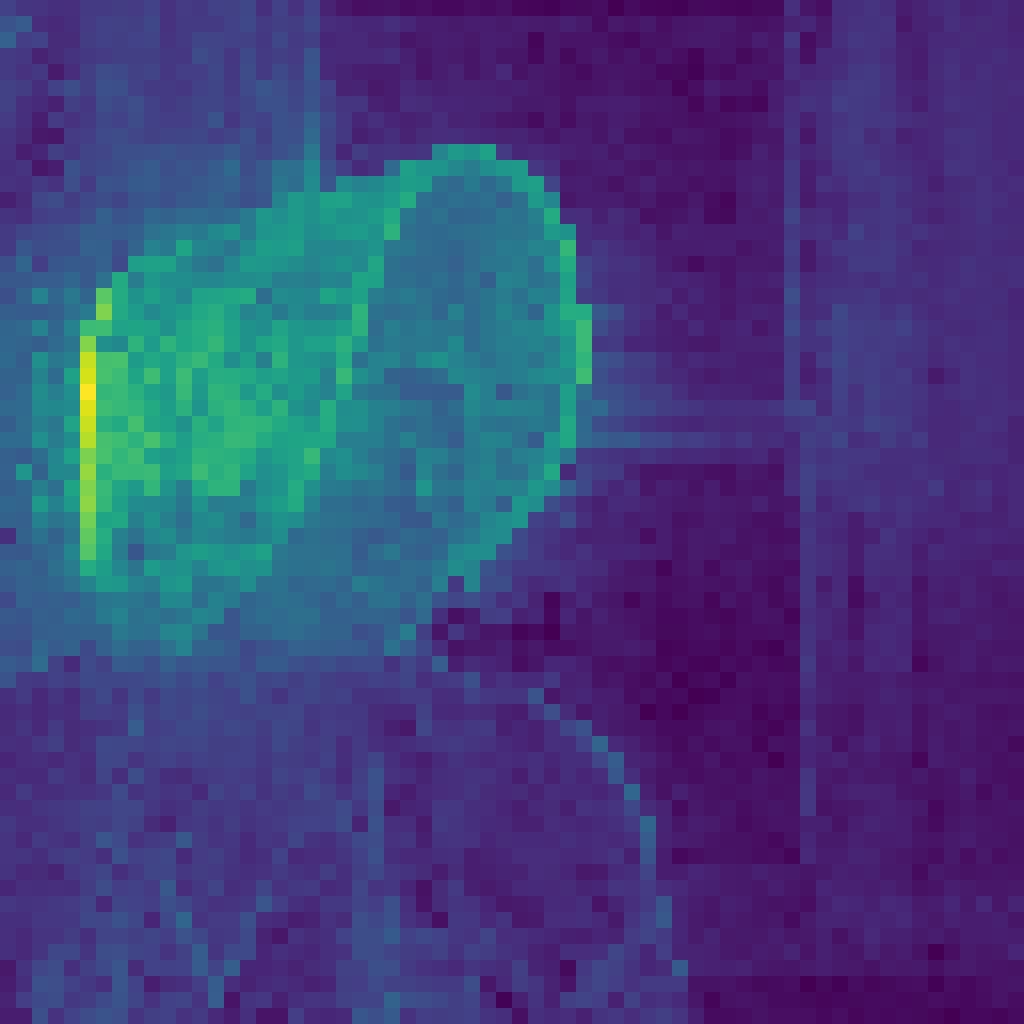}{0.0859375}{0.6171875} &
\patchquerypanel{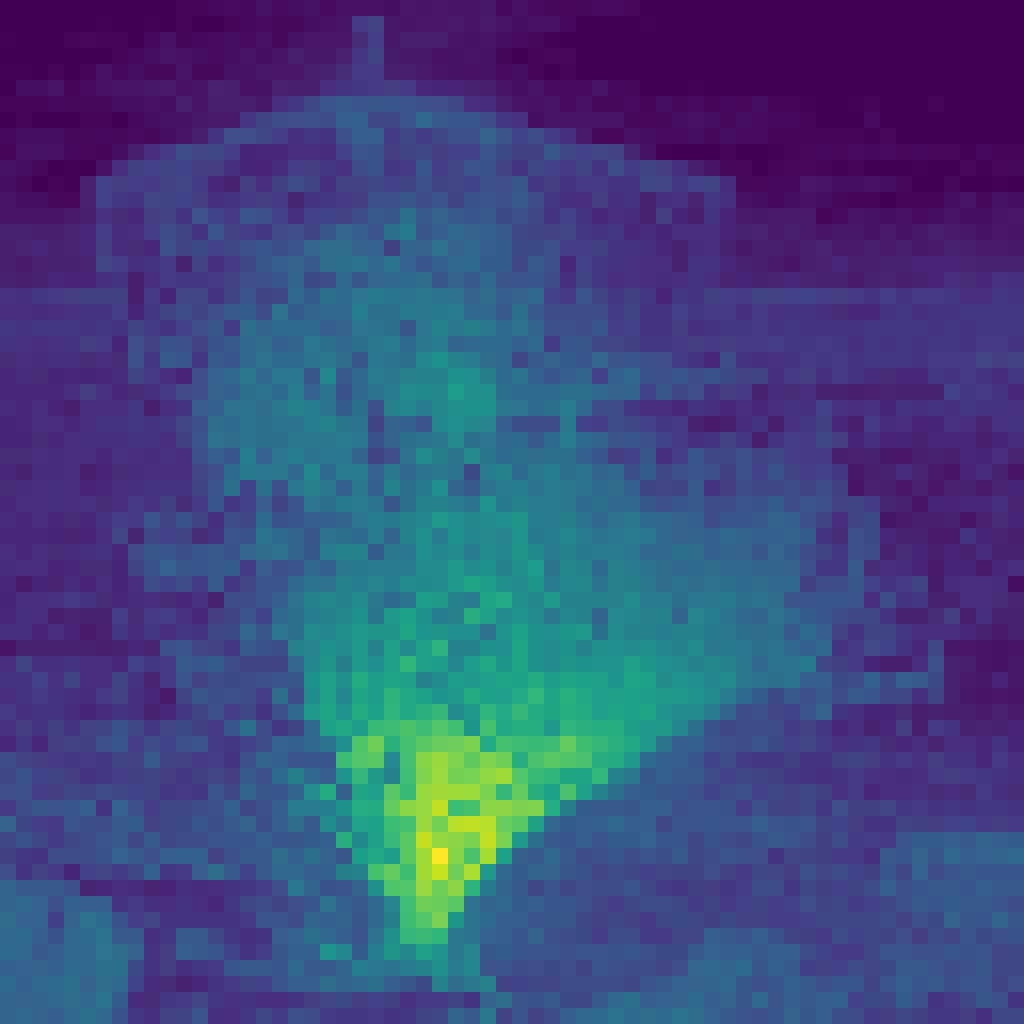}{0.4296875}{0.1640625} &
\patchquerypanel{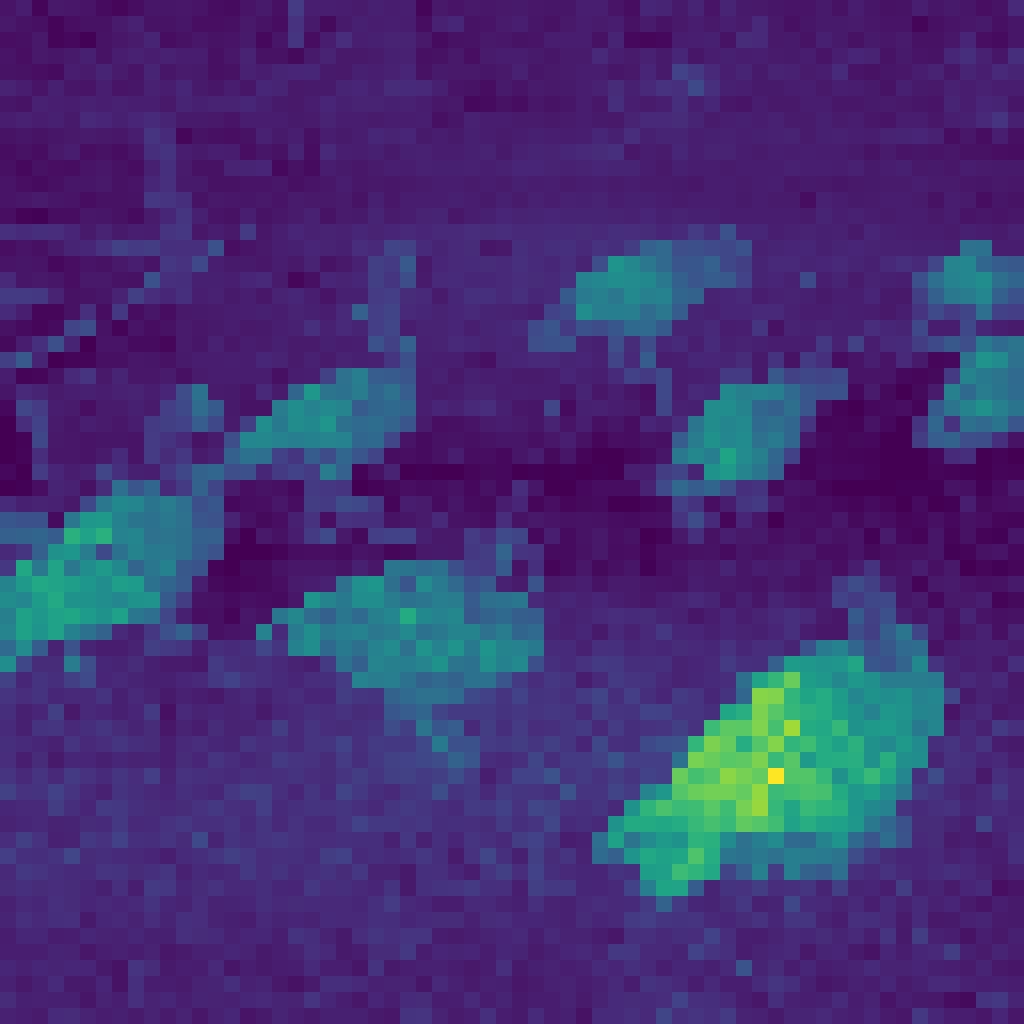}{0.7578125}{0.2421875}
\\
\noalign{\vspace{0.7pt}}
\patchquerypanel{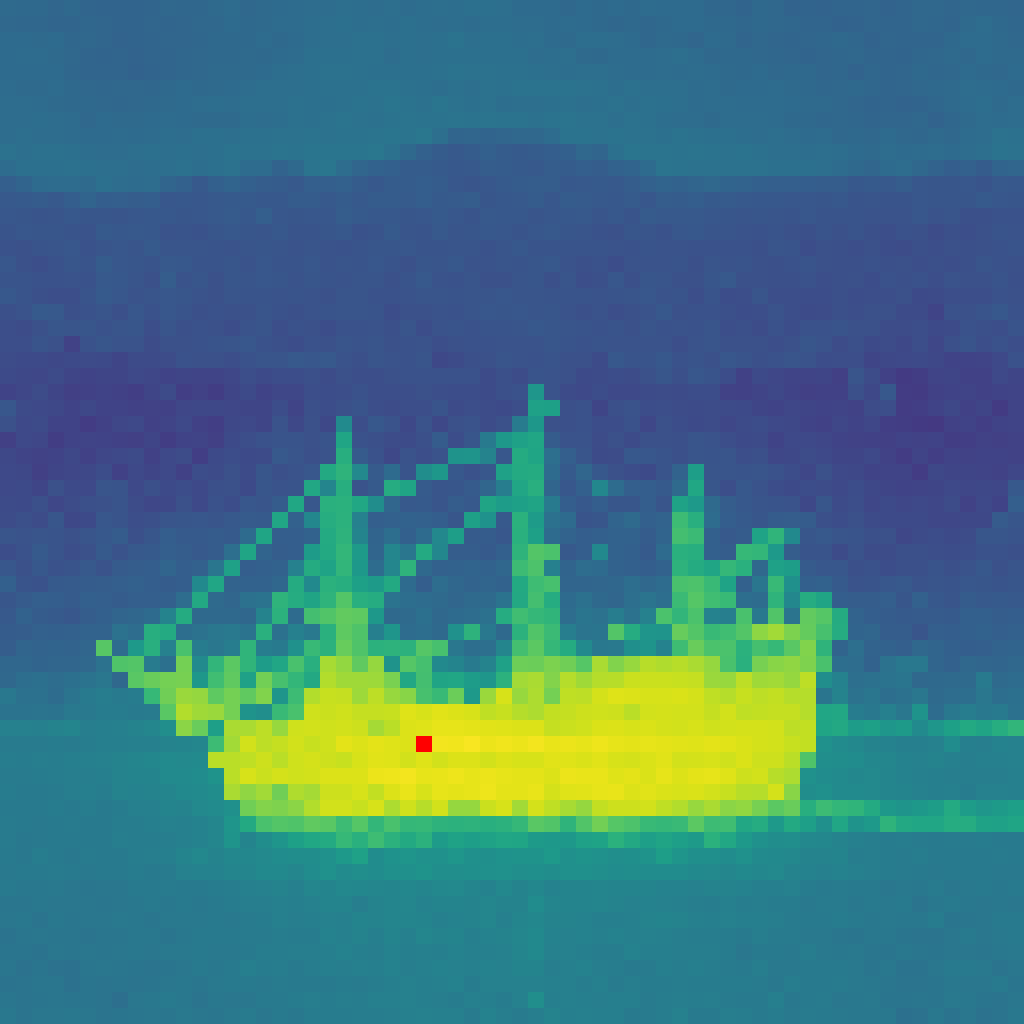}{0.4140625}{0.2734375} &
\patchquerypanel{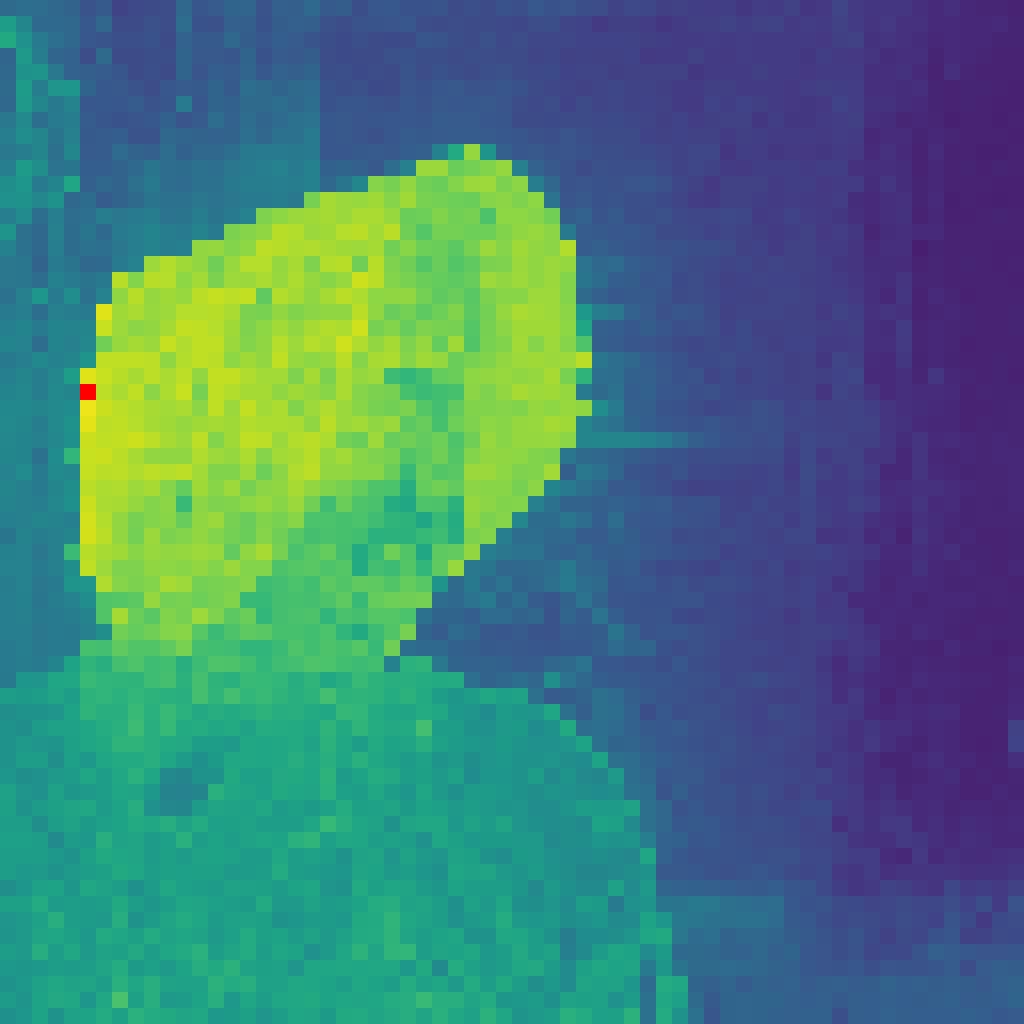}{0.0859375}{0.6171875} &
\patchquerypanel{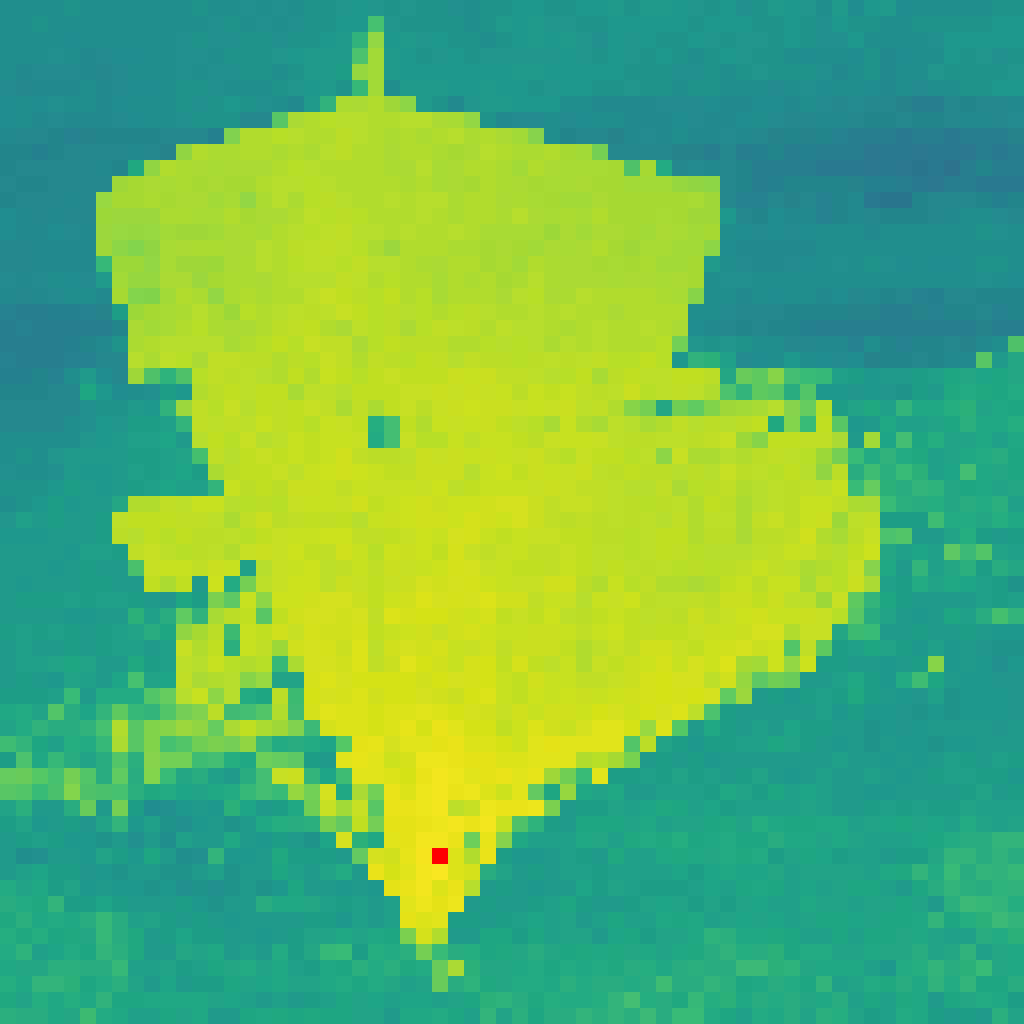}{0.4296875}{0.1640625} &
\patchquerypanel{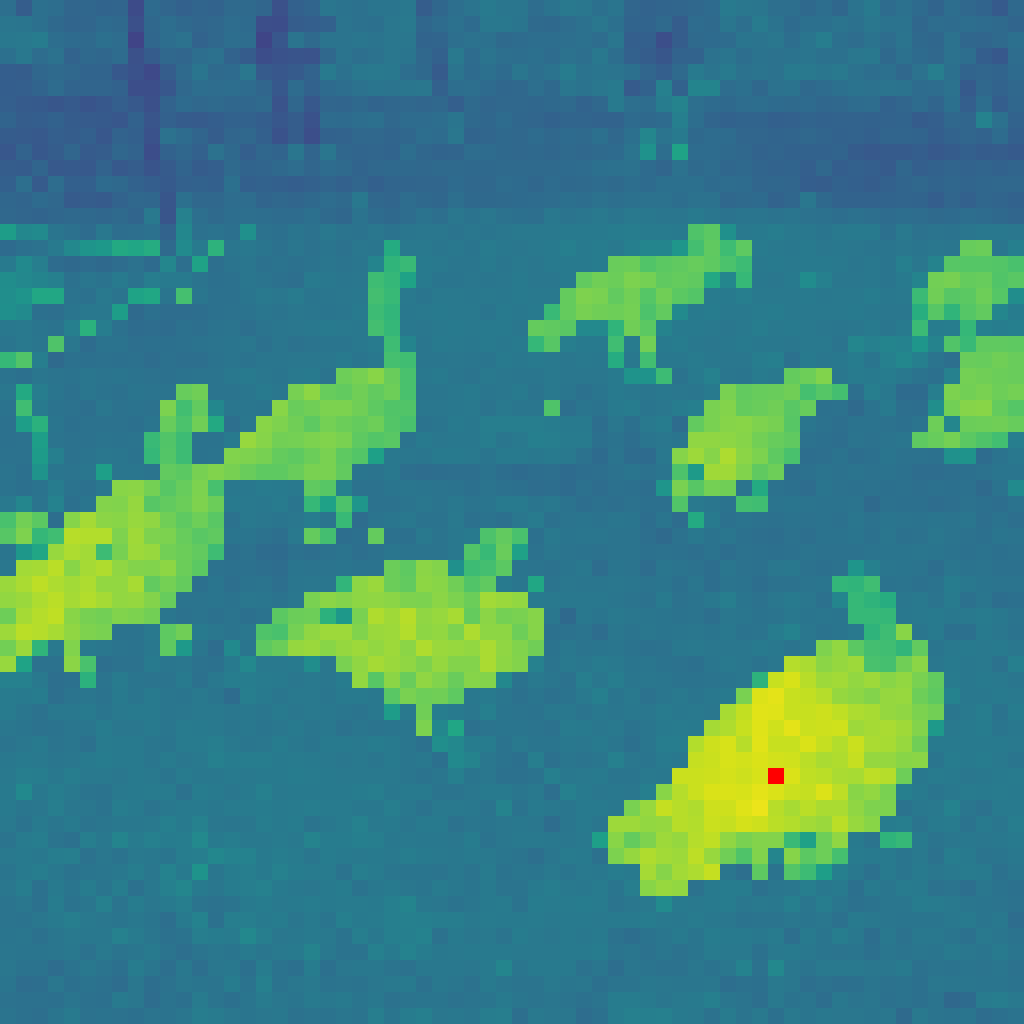}{0.7578125}{0.2421875}
\\
\end{tabular}%
}

\vspace{3pt}
\mbox{\footnotesize
\hspace{\dimexpr\pcalabw+\pcalabelsep\relax}%
\makebox[\pcagroupw]{(a) Prototype maps}%
\hspace{\pcagroupsep}%
\makebox[\pcagroupw]{(b) PCA of patch features}%
\hspace{\pcagroupsep}%
\makebox[\pcagroupw]{(c) Patch similarity}}

\caption{\textbf{Visualisations of iBOT and \methodname{}.} (a) Each patch coloured by its most probable prototype. (b) First three principal components of the patch features as RGB. (c) Cosine similarity to a query patch (red cross).}
\label{fig:visualisations}\label{fig:prototype_maps}\label{fig:pca}\label{fig:patch_locality}
\end{figure*}

 \noindent \textbf{Shared area and patch coverage.}
Figure~\ref{fig:region_selection}a and b examine how strictly regions and patches should be selected. We vary the minimum area that two views must share for their region to be used (a), and the fraction of a patch that must lie inside the region for it to enter the composition, alternatively keeping every patch weighted by this fraction (b). Requiring a larger shared area brings no benefit, as it only discards view pairs. Admitting partially covered patches slightly lowers both segmentation scores, even when they are weighted, because they encode content absent from the other view. The best option is therefore to keep overlaps of every tested size with all the patches fully inside them and not impose any further restrictions on the shared area.

 \noindent \textbf{Loss weight.}
Figure~\ref{fig:region_selection}c shows that performance is relatively insensitive to the regional loss weight $\lambda$. Dense performance plateaus around $\lambda=0.4$, while image-level accuracy decreases slightly at larger weights, as expected. We therefore use $\lambda=0.4$ throughout.

\subsection{Visualisations}
\label{sec:visualizations}
 \noindent \textbf{Prototype maps.}
The regional loss constrains only the composition of a shared region and never assigns a prototype to an individual patch. To show how the patch predictions change nonetheless, Figure~\ref{fig:prototype_maps}a colours each patch by its most probable prototype. iBOT's prototypes change from one patch to the next, even inside an object. \methodname{} instead covers each object with one or two prototypes, which are shared by objects of the same class across images and differ between classes.

\noindent
\textbf{PCA of patch features.}
The frozen-feature evaluations use the backbone features rather than the prototype predictions. Figure~\ref{fig:pca}b therefore maps the first three principal components of the patch features to RGB. \methodname{}'s components follow object boundaries more closely, separating objects from their background while keeping their parts distinct.

 \noindent \textbf{Patch similarity.}
Figure~\ref{fig:patch_locality}c shows the cosine similarity between a query patch and every other patch of the image. With iBOT, high similarity stays close to the query. With \methodname{}, it extends over the whole object and reaches other instances of its class. Patches of the same class thus lie close together in feature space, the property that $k$-NN segmentation, clustering and retrieval rely on.

%% file: img/variance_src/variance_bars.tex
\definecolor{varibot}{HTML}{B1B5BB}
\definecolor{varcont}{HTML}{6B7280}
\definecolor{varreco}{HTML}{2864F5}
\begin{tikzpicture}
\begin{axis}[
  ybar=0.6pt, bar width=5.6pt, scale only axis, width=2.95cm, height=2.25cm,
  ymin=0, ymax=0.82, ytick={0,0.2,0.4,0.6},
  xmin=-0.55, xmax=2.55, xtick={0,1,2},
  xticklabels={{within\\objects}, {between\\objects}, {between\\classes}},
  ymajorgrids, grid style={black!12},
  axis line style={line width=0.5pt}, xtick style={draw=none}, ytick pos=left, ytick style={line width=0.4pt},
  tick label style={font=\scriptsize},
  x tick label style={font=\scriptsize, align=center, yshift=1pt},
  legend style={at={(0.99,0.985)}, anchor=north east, legend columns=1, draw=none, fill=none, font=\scriptsize,
    row sep=-2.6pt, inner sep=0.5pt, legend cell align=left},
  legend image code/.code={\fill[#1, draw=none] (0cm,-0.08cm) rectangle (0.2cm,0.1cm);},
  error bars/y dir=both, error bars/y explicit,
  error bars/error bar style={black, line width=0.4pt}, error bars/error mark options={black, rotate=90, mark size=1.4pt, line width=0.4pt},
]
\addplot[fill=varibot, draw=none] coordinates {(0,0.475) +- (0,0.071) (1,0.261) +- (0,0.064) (2,0.108) +- (0,0.035)};
\addlegendentry{iBOT}
\addplot[fill=varcont, draw=none] coordinates {(0,0.400) +- (0,0.070) (1,0.235) +- (0,0.059) (2,0.120) +- (0,0.035)};
\addlegendentry{iBOT (+200 ep.)}
\addplot[fill=varreco, draw=none] coordinates {(0,0.145) +- (0,0.042) (1,0.269) +- (0,0.083) (2,0.341) +- (0,0.080)};
\addlegendentry{ReCoCo}
\end{axis}
\end{tikzpicture}

%% file: img/ablation_src/ablation_charts.tex
\definecolor{cknn}{HTML}{4B5BC8}
\definecolor{clin}{HTML}{F0468C}
\definecolor{cin}{HTML}{5BBF85}
\pgfplotsset{
  abl knn/.style={cknn, thick, mark=*, mark size=1.8pt, mark options={fill=white, solid}},
  abl lin/.style={clin, thick, mark=square*, mark size=1.8pt, mark options={fill=white, solid}},
  abl img/.style={cin, thick, mark=triangle*, mark size=1.8pt, mark options={fill=white, solid}},
  abl ref/.style={only marks, forget plot, mark options={fill=#1, solid}},
  abl axis/.style={
    scale only axis, width=\ablaxis, height=\ablaxish,
    ymin=59.3, ymax=70.6, ytick={60,62,64,66,68,70},
    ymajorgrids, xmajorgrids, grid style={black!10},
    axis x line*=bottom, axis y line*=left,
    every axis plot/.append style={line width=0.9pt},
    xlabel style={align=center, font=\small, yshift=3pt}, ylabel style={font=\small},
    tick label style={font=\small}, typeset ticklabels with strut,
  },
}
\newlength{\ablaxis}\setlength{\ablaxis}{.335\columnwidth}
\newlength{\ablaxish}\setlength{\ablaxish}{2.45cm}
\newcommand{\ablchartA}{%
\begin{tikzpicture}
\begin{axis}[abl axis,
  xtick={0.1,0.2,0.3,0.5}, xticklabels={10,20,30,50}, xmin=0.07, xmax=0.53,
  xlabel={minimum shared area (\%)\\[1pt](a) Shared area}, ylabel={Score (\%)}]
\addplot[abl knn] coordinates {(0.1,61.75) (0.2,61.61) (0.3,61.29) (0.5,60.24)};
\addplot[abl lin] coordinates {(0.1,66.67) (0.2,66.66) (0.3,66.59) (0.5,65.78)};
\addplot[abl img] coordinates {(0.1,69.11) (0.2,69.04) (0.3,68.96) (0.5,68.76)};
\addplot[abl knn, abl ref=cknn] coordinates {(0.1,61.75)};
\addplot[abl lin, abl ref=clin] coordinates {(0.1,66.67)};
\addplot[abl img, abl ref=cin] coordinates {(0.1,69.11)};
\node[font=\small, text=cin!75!black, anchor=north, inner sep=1pt] at (axis cs:0.25,68.65) {IN1k $k$-NN};
\node[font=\small, text=clin!85!black, anchor=north, inner sep=1pt] at (axis cs:0.25,66.2) {VOC linear};
\node[font=\small, text=cknn, anchor=south, inner sep=1pt] at (axis cs:0.25,61.9) {VOC $k$-NN};
\end{axis}
\end{tikzpicture}}
\newcommand{\ablchartB}{%
\begin{tikzpicture}
\begin{axis}[abl axis,
  xtick={-0.25,0.2,0.5,1}, xticklabels={\textit{weighted},20,50,100}, xmin=-0.45, xmax=1.06,
  xlabel={patch area in region (\%)\\[1pt](b) Patch coverage}]
\addplot[abl knn] coordinates {(0.2,61.36) (0.5,61.27) (1,61.75)};
\addplot[abl lin] coordinates {(0.2,66.27) (0.5,66.24) (1,66.67)};
\addplot[abl img] coordinates {(0.2,69.22) (0.5,69.12) (1,69.11)};
\draw[black!35, densely dashed, line width=0.5pt] (axis cs:0,59.3) -- (axis cs:0,70.6);
\addplot[abl knn, only marks, forget plot] coordinates {(-0.25,61.09)};
\addplot[abl lin, only marks, forget plot] coordinates {(-0.25,66.29)};
\addplot[abl img, only marks, forget plot] coordinates {(-0.25,69.18)};
\addplot[abl knn, abl ref=cknn] coordinates {(1,61.75)};
\addplot[abl lin, abl ref=clin] coordinates {(1,66.67)};
\addplot[abl img, abl ref=cin] coordinates {(1,69.11)};
\end{axis}
\end{tikzpicture}}
\newcommand{\ablchartC}{%
\begin{tikzpicture}
\begin{axis}[abl axis,
  xtick={0.2,0.4,0.6,0.8,1}, xticklabels={0.2,0.4,0.6,0.8,1.0}, xmin=0.15, xmax=1.05,
  xlabel={weight $\lambda$\\[1pt](c) Loss weight}, ylabel={Score (\%)}]
\addplot[abl knn] coordinates {(0.2,61.31) (0.4,61.75) (0.6,61.63) (0.8,61.47) (1,61.53)};
\addplot[abl lin] coordinates {(0.2,66.22) (0.4,66.67) (0.6,66.72) (0.8,66.41) (1,66.45)};
\addplot[abl img] coordinates {(0.2,69.35) (0.4,69.11) (0.6,68.87) (0.8,68.73) (1,68.59)};
\addplot[abl knn, abl ref=cknn] coordinates {(0.4,61.75)};
\addplot[abl lin, abl ref=clin] coordinates {(0.4,66.67)};
\addplot[abl img, abl ref=cin] coordinates {(0.4,69.11)};
\end{axis}
\end{tikzpicture}}

%% file: img/patch_query_marker.tex
\definecolor{patchQuerySelf}{RGB}{253,231,36}
\newcommand{\patchquerypanel}[3]{%
\begin{tikzpicture}[baseline=0pt]
\node[anchor=south west,inner sep=0,outer sep=0] (panel) at (0,0)
    {\includegraphics[width=\pcaw]{#1}};
\begin{scope}[x={(panel.south east)},y={(panel.north west)}]
\fill[patchQuerySelf] ({#2-1/128},{#3-1/128})
    rectangle ({#2+1/128},{#3+1/128});
\coordinate (query) at (#2,#3);
\end{scope}
\draw[white,line width=1.1pt] (query) ++(-1.8pt,0) -- ++(3.6pt,0);
\draw[white,line width=1.1pt] (query) ++(0,-1.8pt) -- ++(0,3.6pt);
\draw[red,line width=.65pt] (query) ++(-1.8pt,0) -- ++(3.6pt,0);
\draw[red,line width=.65pt] (query) ++(0,-1.8pt) -- ++(0,3.6pt);
\end{tikzpicture}%
}

%% file: sec/5_conclusions.tex
\section{Conclusions}
\label{section:conclusions}

We introduced \methodname{}, a regional self-supervised objective that extends cross-view consistency to the visual content shared between augmented views. By aligning compositions of patch predictions over learned prototypes, \methodname{} provides regional supervision without requiring patch correspondences or semantic grouping. Extensive experiments demonstrate substantial improvements in frozen-feature segmentation, clustering, and retrieval while maintaining strong image-level recognition. Our analysis reveals more coherent representations within objects and greater separation between classes, with these improvements reflected in the backbone features. These findings establish regional composition agreement as a simple and effective supervisory signal bridging image-level consistency and individual-patch prediction.

\noindent
\textbf{Limitations and Future Work.}
Regional composition encourages semantic consistency across views but does not explicitly preserve fine-grained spatial relationships. Thus, its benefits are less pronounced on geometry-sensitive tasks, with slightly degraded keypoint correspondence and largely unchanged depth estimation. Future work could explore how to improve geometric representations while retaining the benefits of correspondence-free regional supervision.

%% file: sec/suppl.tex
\clearpage
\setcounter{page}{1}
\maketitlesupplementary

\appendix

\input{img/analysis_src/analysis_charts.tex}

In these appendices, we first characterise the regional and masked-patch losses from an information-theoretic perspective (Appendix~\ref{sec:appx_information}). We then give the training and evaluation details (Appendix~\ref{sec:implementation}) and examine the initialisation from iBOT (Appendix~\ref{sec:fair_comparison}). Finally, we report additional ablations, in-context segmentation and end-to-end fine-tuning results, and an analysis of the features across backbone blocks (Appendix~\ref{sec:additional_experiments}), together with further visualisations (Appendix~\ref{sec:visual_examples}).

\section{Information-Theoretic Perspective}
\label{sec:appx_information}

We characterise the regional consistency loss and the complementary constraints of Sec.~\ref{sec:region_representation} in information-theoretic terms. The statements below hold for fixed teacher predictions, as under the stop-gradient. We consider a teacher view $v_i$ and a student view $v$ with shared region $o$ and omit these indices, denoting the predictions of the selected patches by $\mathbf q_p^t$ and $\mathbf q_p^s$ and their compositions (Eq.~\ref{eq:regional_composition}) by $\mathbf h^t$ and $\mathbf h^s$. The entropy is $H(\mathbf a)=H(\mathbf a,\mathbf a)$, and the Kullback--Leibler divergence $\mathrm{KL}(\mathbf a\,\|\,\mathbf b)=H(\mathbf a,\mathbf b)-H(\mathbf a)$ is non-negative and zero only if $\mathbf a=\mathbf b$~\cite{cover2006elements}. Predictions are strictly positive, with zero entries understood as limits.

\paragraph{Regional composition.}
As the cross-entropy $H(\mathbf a,\mathbf b)$ is linear in $\mathbf a$ and the composition $\mathbf h$ of either network is the mean of its $n$ selected predictions $\mathbf q_p$, its entropy decomposes as
\begin{equation}
\begin{aligned}
H(\mathbf h)&=\frac1n\sum_pH(\mathbf q_p,\mathbf h)\\
&={\underbrace{\frac1n\sum_pH(\mathbf q_p)}_{\text{uncertainty }\bar H}}
+{\underbrace{\frac1n\sum_p\mathrm{KL}(\mathbf q_p\,\|\,\mathbf h)}_{\text{diversity }D}}.
\end{aligned}
\label{eq:appx_entropy}
\end{equation}
The diversity $D$ is the generalised Jensen--Shannon divergence of the predictions with equal weights~\cite{lin1991divergence} and is zero only if they are all equal. In the limit of one-hot predictions, $\bar H\to0$, $h_k$ tends to the fraction of patches assigned to prototype~$k$, and $D$ tends to the entropy of these fractions.

\paragraph{Regional consistency loss.}
With Eq.~\ref{eq:appx_entropy} applied to the teacher, the regional cross-entropy decomposes as
\begin{equation}
\begin{aligned}
c_{v_i,v}&=H(\mathbf h^t,\mathbf h^s)=H(\mathbf h^t)+\mathrm{KL}(\mathbf h^t\,\|\,\mathbf h^s)\\
&=\bar H^t+D^t+\mathrm{KL}(\mathbf h^t\,\|\,\mathbf h^s).
\end{aligned}
\label{eq:appx_region}
\end{equation}
Only the last term depends on the student, so $c_{v_i,v}$ is minimal only if $\mathbf h^s=\mathbf h^t$, and it is unchanged by any permutation of the selected predictions within either view.


\paragraph{Regions at several scales.}
A teacher view $v_i$ provides the targets of $g-1+\ell$ regional terms (Eq.~\ref{eq:region_loss}), which average its predictions $\mathbf q_{v_i,p}^t$ over regions that can differ in size and position. We split the $n$ patches of a region into sub-regions with $n_j$ patches, compositions $\mathbf h_j$ and diversities $D_j$, so that $\mathbf h=\sum_j\frac{n_j}n\mathbf h_j$. As in Eq.~\ref{eq:appx_entropy}, the mean of $\mathrm{KL}(\mathbf q_p\,\|\,\mathbf h)$ over the patches of sub-region $j$ is $D_j+\mathrm{KL}(\mathbf h_j\,\|\,\mathbf h)$, hence
\begin{equation}
D={\underbrace{\sum_j\frac{n_j}nD_j}_{\text{within}}}+{\underbrace{\sum_j\frac{n_j}n\mathrm{KL}(\mathbf h_j\,\|\,\mathbf h)}_{\text{between}}}.
\label{eq:appx_scales}
\end{equation}
The compositions of the sub-regions determine the second term and leave open only the split of Eq.~\ref{eq:appx_entropy} within each sub-region; with single patches as sub-regions, nothing is left open. Where one region of $v_i$ contains another, Eq.~\ref{eq:appx_scales} applies with the patches of the smaller region and the remaining patches as sub-regions.

\paragraph{Complementary patch-level constraints.}
At the minimum of $c_{v_i,v}$, where $\mathbf h^s=\mathbf h^t$, Eq.~\ref{eq:appx_entropy} fixes the sum $\bar H^s+D^s=H(\mathbf h^t)$ but, with two or more selected patches in the student view, not its split: the constraint is collective. In the example of Sec.~\ref{sec:region_representation}, patches each assigned with certainty to one of two prototypes, half to each, have $(\bar H,D)=(0,\log2)$, whereas patches that all give probability $1/2$ to both have $(\log2,0)$, with the same composition. The masked-patch loss imposes individual constraints: each of its terms, $H(\mathbf q_{v_i,p}^t,\mathbf q_{v_i,p}^s)$ at a masked position $p\in\mathcal M_i$, is minimal only if the student's prediction equals its target. In the objective of Eq.~\ref{eq:full_objective}, a patch prediction of the student is compared with the teacher's prediction at its own position only at the masked positions of the global views; elsewhere, it enters only through compositions. The two losses are complementary in this sense: the minimum of the regional loss fixes the composition of the shared region, and thus $\bar H^s+D^s$, and that of the masked-patch loss fixes each masked prediction, and thus its entropy. If $v$ is a global view and the student's predictions equal the teacher's at all its selected patches, not only at the masked ones, then $\mathbf h_{v,o}^s=\mathbf h_{v,o}^t$, the target of $c_{v,v_i}$, and the last term of Eq.~\ref{eq:appx_region} becomes $\mathrm{KL}(\mathbf h_{v_i,o}^t\,\|\,\mathbf h_{v,o}^t)$. This term is zero only if the teacher's compositions of the shared region coincide in the two views: the regional loss is the only term of Eq.~\ref{eq:full_objective} that relates patch predictions of different views.

\section{Training and Evaluation Details}
\label{sec:implementation}

\subsection{Pretraining}
\label{sec:pretraining_details}

\paragraph{Initialisation.}
We initialise each backbone from the released iBOT~\cite{zhou2022ibot} checkpoint, restoring the student, the EMA teacher, the projection head and the teacher centres, and reset only the optimiser. Since training continues from a converged model, neither the learning-rate nor the teacher-temperature warm-up is repeated. A single projection head maps the $[\mathrm{CLS}]$ token, the masked patches and the patches of the regional compositions to the same $K$ prototypes.
In the released ViT-S/16 and ViT-B/16 checkpoints, the student and teacher weights are almost the same. In ViT-L/16, whose pretraining ends at a high learning rate, they differ substantially, and the EMA would overwrite the released teacher, the network that iBOT evaluates, with the student within the first epoch. For ViT-L/16, we therefore copy the released teacher into the student, so that training starts from the same weights in both networks, as for the smaller backbones.

\paragraph{Training.}
We train on ImageNet-1K~\cite{deng2009imagenet} with four GPUs for 200, 100 and 40 epochs for ViT-S/16, ViT-B/16 and ViT-L/16, respectively. Augmentations and block masking follow iBOT~\cite{zhou2022ibot}. The regional loss pairs each teacher global view with the other global view and with every local view, using only the patches that lie fully inside their shared region. We optimise with AdamW~\cite{loshchilov2019adamw} in FP16 precision, with cosine schedules for the learning rate, weight decay and teacher momentum. Table~\ref{tab:continuation_settings} lists the full configuration of the three backbones.

\begin{table}[t]
\centering
\caption{\textbf{Pretraining configuration} of \methodname{} for the three backbones, initialised from the released iBOT~\cite{zhou2022ibot} checkpoints.}
\label{tab:continuation_settings}
\footnotesize
\setlength{\tabcolsep}{3.5pt}
\begin{tabular}{@{}lccc@{}}
\toprule
Setting & ViT-S/16 & ViT-B/16 & ViT-L/16 \\
\midrule
\multicolumn{4}{@{}l}{\textit{\textbf{Views and masking}}} \\
Global views      & \multicolumn{3}{c}{$2\times224^2$, scale $[0.25,1.0]$} \\
Local views       & \multicolumn{3}{c}{$10\times96^2$, scale $[0.05,0.25]$} \\
Mask fraction     & \multicolumn{3}{c}{$0$ or $[0.1,0.5]$} \\
\midrule
\multicolumn{4}{@{}l}{\textit{\textbf{Objective}}} \\
Prototypes $K$    & \multicolumn{3}{c}{$8192$} \\
Teacher / student temperature & \multicolumn{3}{c}{$0.07$ / $0.1$} \\
Regional weight $\lambda$ & \multicolumn{3}{c}{$0.4$} \\
Regional view pairs & \multicolumn{3}{c}{global--global, global--local} \\
Patch selection   & \multicolumn{3}{c}{fully inside the shared region} \\
\midrule
\multicolumn{4}{@{}l}{\textit{\textbf{Optimisation}}} \\
Epochs            & $200$ & $100$ & $40$ \\
GPUs              & \multicolumn{3}{c}{$4$} \\
Batch per GPU     & $64$ & $32$ & $16$ \\
Accumulation steps & $1$ & $2$ & $2$ \\
Global batch      & $256$ & $256$ & $128$ \\
Initial LR        & $1.0\times10^{-5}$ & $1.0\times10^{-5}$ & $5.0\times10^{-6}$ \\
Final LR          & \multicolumn{3}{c}{$10^{-6}$} \\
Weight decay      & \multicolumn{3}{c}{$0.04\rightarrow0.5$} \\
Teacher momentum  & \multicolumn{3}{c}{$0.9995\rightarrow1$} \\
Teacher centre momentum & \multicolumn{3}{c}{$0.9$} \\
Drop path         & $0.1$ & $0.1$ & $0.2$ \\
Gradient clip     & \multicolumn{3}{c}{$3.0$} \\
Optimiser         & \multicolumn{3}{c}{AdamW, $\beta=(0.9,0.999)$, $\epsilon=10^{-8}$} \\
\bottomrule
\end{tabular}
\end{table}

\subsection{Initialisation from iBOT}
\label{sec:from_scratch}\label{sec:fair_comparison}

The \methodname{} models pretrained on ImageNet-1K~\cite{deng2009imagenet} continue the released iBOT~\cite{zhou2022ibot} checkpoints with the full objective of Eq.~\ref{eq:full_objective}. This objective differs from that of iBOT only by the regional loss: the data, the projection head and the prototypes are those of iBOT, its image-level and masked-patch losses remain active, and no parameters are added (training settings in Table~\ref{tab:continuation_settings}). We adopt this protocol for two reasons. First, it isolates the effect of the regional loss: the control introduced below and every ablation variant start from the same weights as \methodname{}, so any difference between them arises during the continuation. Second, a ViT-S/16 run takes 200 epochs, or 50 for an ablation variant, instead of the 800 epochs of iBOT's pretraining, which makes three backbones and the ablations of Sec.~\ref{sec:ablations} affordable. Starting from a pretrained self-supervised backbone is also common among dense methods~\cite{ziegler2022leopart,salehi2023timet,pariza2025neco}. The protocol nevertheless leaves two alternative explanations of the gains, the additional training epochs and the iBOT weights themselves, which we test in turn with ViT-S/16.

\begin{figure}[t]
\centering
\input{img/scratch_src/scratch_charts}
\caption{\textbf{Training from scratch and continued training} of ViT-S/16: frozen VOC $k$-NN and linear mIoU (top) and IN1k $k$-NN accuracy with a 10\% reference bank (bottom). Epochs count all training from random initialisation, so the continued models start from the released iBOT~\cite{zhou2022ibot} after its 800 epochs. Dotted lines give the final values of the released iBOT, and vertical lines the start of the masked-patch and regional losses of the model trained from scratch, which covers the first 400 epochs of its 800-epoch schedule.}
\label{fig:from_scratch}
\end{figure}
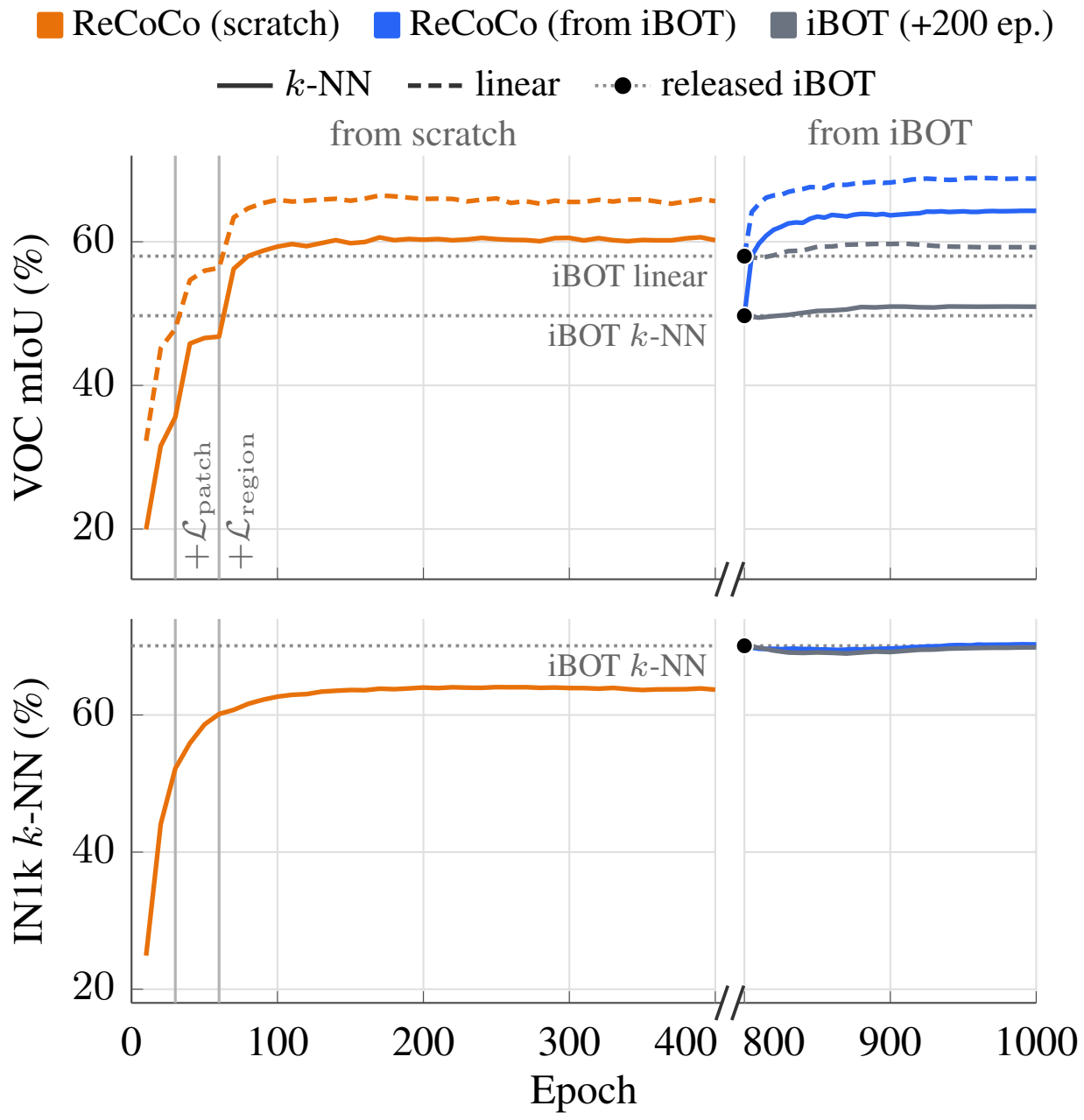

\begin{table*}[t]
\centering
\caption{\textbf{Fair comparison with iBOT} with frozen ViT-S/16 features on the benchmarks of the main paper (Tables~\ref{tab:semantic_segmentation}--\ref{tab:object_region_retrieval} and Figure~\ref{fig:feature_analysis}b), with the same evaluation protocols for the three models. iBOT (+200 ep.) continues the released iBOT~\cite{zhou2022ibot} for the same 200 epochs as \methodname{}, without the regional loss. All values are in \%, except depth, which is the RMSE in metres (lower is better). We highlight the \textbf{best}.}
\label{tab:fair_comparison}
\footnotesize
\setlength{\tabcolsep}{2pt}
\newcolumntype{Y}{>{\centering\arraybackslash}X}
\begin{tabularx}{\textwidth}{l*{16}{Y}}
\toprule
& \multicolumn{6}{c}{Semantic segmentation} & \multicolumn{6}{c}{Clustering} & \multicolumn{4}{c}{ImageNet-1K classification} \\
\cmidrule(lr){2-7}\cmidrule(lr){8-13}\cmidrule(l){14-17}
& \multicolumn{2}{c}{ADE20K} & \multicolumn{2}{c}{VOC} & \multicolumn{2}{c}{Cityscapes} & \multicolumn{3}{c}{VOC} & \multicolumn{3}{c}{COCO-Things} & \multicolumn{3}{c}{$k$-NN} & Linear \\
\cmidrule(lr){2-3}\cmidrule(lr){4-5}\cmidrule(lr){6-7}\cmidrule(lr){8-10}\cmidrule(lr){11-13}\cmidrule(lr){14-16}\cmidrule(l){17-17}
Model & $k$-NN & Lin. & $k$-NN & Lin. & $k$-NN & Lin. & GT & 300 & 500 & GT & 300 & 500 & 1\% & 10\% & 100\% & 100\% \\
\midrule
iBOT~\cite{zhou2022ibot} & 22.0 & 27.0 & 49.7 & 58.0 & 33.5 & 38.0 & 5.8 & 27.8 & 33.0 & 9.4 & 29.7 & 34.0 & 62.3 & 70.1 & 74.9 & 78.0 \\
iBOT (+200 ep.) & 22.6 & 27.7 & 51.0 & 59.3 & 32.9 & 37.9 & 5.9 & 27.9 & 33.0 & 9.4 & 31.0 & 34.0 & 61.8 & 69.9 & 74.8 & 78.0 \\
\rowcolor{blue}
\textbf{\methodname{}} & \textbf{29.2} & \textbf{31.2} & \textbf{64.3} & \textbf{68.8} & \textbf{35.0} & \textbf{39.8} & \textbf{18.7} & \textbf{61.7} & \textbf{64.8} & \textbf{24.6} & \textbf{59.3} & \textbf{61.6} & \textbf{63.3} & \textbf{70.3} & \textbf{75.1} & \textbf{78.6} \\
\bottomrule
\end{tabularx}

\vspace{5pt}
\begin{tabularx}{\textwidth}{l*{13}{Y}}
\toprule
& \multicolumn{6}{c}{Multi-label classification} & \multicolumn{4}{c}{Retrieval} & Corr. & \multicolumn{2}{c}{Depth $\downarrow$} \\
\cmidrule(lr){2-7}\cmidrule(lr){8-11}\cmidrule(lr){12-12}\cmidrule(l){13-14}
& \multicolumn{4}{c}{VOC} & COCO & VG & \multicolumn{2}{c}{COCO objects} & \multicolumn{2}{c}{VOC regions} & SPair & NYU & KITTI \\
\cmidrule(lr){2-5}\cmidrule(lr){6-6}\cmidrule(lr){7-7}\cmidrule(lr){8-9}\cmidrule(lr){10-11}\cmidrule(lr){12-12}\cmidrule(lr){13-13}\cmidrule(l){14-14}
Model & 1 Img & 2 Imgs & 5 Imgs & Full & Full & Full & mAP & P@50 & mAP & R@1 & PCK & RMSE & RMSE \\
\midrule
iBOT~\cite{zhou2022ibot} & 47.0 & 56.6 & 73.0 & 89.2 & 69.6 & 25.8 & 20.9 & 32.1 & 49.3 & 82.3 & \textbf{29.6} & 0.513 & 3.605 \\
iBOT (+200 ep.) & 46.7 & 57.1 & 72.7 & 89.2 & 70.2 & 26.2 & 22.9 & 34.2 & 50.7 & 82.2 & 29.2 & \textbf{0.500} & \textbf{3.574} \\
\rowcolor{blue}
\textbf{\methodname{}} & \textbf{51.3} & \textbf{60.6} & \textbf{75.1} & \textbf{89.8} & \textbf{71.1} & \textbf{26.8} & \textbf{38.4} & \textbf{48.4} & \textbf{63.0} & \textbf{86.0} & 26.7 & 0.502 & 3.622 \\
\bottomrule
\end{tabularx}
\end{table*}

\paragraph{Additional epochs.}
With ViT-S/16, \methodname{} totals 1,000 epochs of training, 800 from iBOT and 200 from the continuation. The control iBOT (+200 ep.) is its budget-matched baseline: it continues the same checkpoint for the same 200 epochs without the regional loss, with the settings of Table~\ref{tab:continuation_settings}. Table~\ref{tab:fair_comparison} compares the three models on the frozen benchmarks of the main paper, with identical protocols and both continuations, each trained once, taken at their final epoch. The additional epochs change iBOT little: segmentation mIoU moves by $-0.6$ to $+1.3$ points across the six settings, ImageNet-1K $k$-NN accuracy by $-0.5$ to $-0.1$ points and object retrieval mAP by $+2.0$ points. With the regional loss, the same 200 epochs change them by $+1.5$ to $+14.6$, $+0.2$ to $+1.0$ and $+17.5$ points. The control also remains below the best prior method pretrained on ImageNet-1K in all six ViT-S/16 segmentation settings of Table~\ref{tab:semantic_segmentation}, whereas \methodname{} exceeds it in all six, so additional iBOT epochs alone do not explain this lead; in a 50-epoch continuation, \methodname{} already exceeds the best prior VOC $k$-NN result ($62.4$ against $60.9$, Table~\ref{tab:ablations}c). The gains thus come from the regional loss rather than from the continuation, also during training (Figure~\ref{fig:from_scratch}, right): after five epochs, the dense probes of \methodname{} ($58.1$ and $64.1$) already exceed every value that those of the control reach in 200 epochs (at most $51.0$ and $59.7$).

\paragraph{Training from scratch.}
To test whether the gains require the iBOT weights, we train \methodname{} from random initialisation with the original recipe of iBOT ViT-S/16, including its 800-epoch schedule, and introduce the losses progressively: the masked-patch loss starts at epoch 30, an option that iBOT already provides, and the regional loss at epoch 60, once the patch predictions that it aggregates are being trained. Within our compute budget, we run the first 400 epochs of this schedule and follow it with the probes of Figure~\ref{fig:from_scratch} (left). While only the iBOT losses are active, the dense probes stay below those of the released iBOT and have almost stopped improving, with VOC $k$-NN and linear mIoU changing by $0.2$ and $0.4$ points between epochs 50 and 60. Ten epochs after the regional loss starts, they have risen from $46.8$ and $56.4$ to $56.2$ and $63.4$, above the $49.7$ and $58.0$ that iBOT reaches after its full 800 epochs. From epoch 100 onwards, they stay at about $60$ and $66$, around $10$ and $8$ points above the released iBOT. Global performance, measured by ImageNet-1K $k$-NN accuracy, does not drop when the regional loss starts and levels off at about $64\%$, whereas the $70.1\%$ of the released iBOT is measured at the end of its schedule. Most of the dense gain on these probes is thus obtained without the released iBOT weights.

\paragraph{Role of the regional loss.}
Together, the two experiments attribute the dense gains mainly to the regional loss: removing it from the continuation removes them, although the initialisation, the data and the number of epochs are unchanged, and adding it to a model trained from scratch lifts the dense probes above those of the fully trained iBOT within ten epochs. The form of the loss matters as well: in the 50-epoch continuations of Table~\ref{tab:ablations}a, a regional loss that aligns averaged backbone features directly leaves VOC $k$-NN mIoU at the level of iBOT ($49.8$ against $49.7$), whereas matching compositions raises it to $61.8$. We therefore keep the iBOT initialisation in the main paper, where it gives all backbones and ablations a common protocol.

\subsection{Evaluation Protocols}
\label{sec:evaluation_protocols}

Unless stated otherwise, evaluations use the frozen patch features of the EMA teacher after the final backbone LayerNorm and before the projection head. This section describes the protocols of the evaluations in the main paper.

\paragraph{Semantic segmentation.}
We use the released evaluation of CAPI~\cite{darcet2025capi}, whose $224\times224$ inputs give 256 patch tokens at patch size 14, and resize images to $256\times256$ so that our patch-16 backbones receive the same 256 tokens. The patch features are standardised. The $k$-NN probe assigns each patch the majority label of its $k$ nearest training patches, with $k\in\{1,3,10,30\}$ and cosine or $\ell_2$ distance, and the linear probe is a logistic regression. Probe hyperparameters are selected on a seeded 10\% holdout of the training set, and the probes are then refitted on all training images to report validation mIoU. VOC uses the original 1,464/1,449 segmentation training/validation images, without SBD~\cite{hariharan2011sbd} or \texttt{trainaug}; ADE20K uses the official splits and Cityscapes the fine annotations.

\paragraph{Frozen clustering evaluation.}
We use the released evaluation of NeCo~\cite{pariza2025neco}. Images are resized to $448\times448$ and scored with $100\times100$ masks. Features are standardised, reduced to 50 dimensions by PCA and clustered with FAISS~\cite{johnson2017faiss} $k$-means. Transformations and clusters are fitted on the validation features, so the evaluation is transductive, and labels are used only to match clusters to classes, one-to-one for $K$ equal to the number of classes and many-to-one for overclustering. Scores average five clustering seeds.

\paragraph{Multi-class classification.}
Weighted $k$-NN uses the $\ell_2$-normalised final $[\mathrm{CLS}]$ token, 20 neighbours and temperature $0.07$. The 1\% and 10\% banks are the fixed SimCLRv2~\cite{chen2020simclrv2} subsets, and every bank is evaluated on the full validation set. The linear probe adapts the linear evaluation of iBOT~\cite{zhou2022ibot} and concatenates the last four normalised $[\mathrm{CLS}]$ tokens. It is trained for 200 epochs on $224\times224$ random resized crops with horizontal flips, using SGD with momentum $0.9$, no weight decay, a global batch of 1,024, an initial learning rate of $0.004$ and cosine decay. Evaluation resizes the shorter side to 256 and centre-crops $224\times224$, and we report the final-epoch top-1 accuracy.

\paragraph{Multi-label classification.}
Multi-label probes use the same features and recipe with masked binary cross-entropy, keeping the image labels after cropping, for 500 epochs on VOC and 200 epochs on COCO~\cite{lin2014coco} and Visual Genome~\cite{krishna2017visualgenome}, and we report the final-epoch non-interpolated macro AP. All four models use the same held-out splits: VOC2007 trainval/test, COCO2017 train/val and the SSGRL VG500 train/test split. Low-shot VOC subsets use nested seed-$0$ per-class positive draws, deduplicate the sampled images and keep all their known labels; all VOC probes are scored on the full test split.

\paragraph{Object and region retrieval.}
For objects, one ground-truth instance in each of 50 COCO training images per category, selected with seed $0$ from the released query image IDs~\cite{shlapen2024region}, queries all 5,000 validation images. Features are upsampled bilinearly to the native image resolution and averaged within each mask, an image is scored by the maximum dot product between the query and its SAM gallery regions, and it is relevant if it contains the query category. mAP and Precision@50 are averaged over the queries of each category and then over categories. For regions, each object class present in a VOC2012 image forms one region, whose descriptor averages the patch features at $448\times448$, weighted by their overlap with the ground-truth mask, and is $\ell_2$-normalised. The regions of the validation images query those of the training images by cosine similarity, a gallery region is relevant if it has the query class, and we report class-macro mAP and Recall@1.

\paragraph{Post-training foundation models.}
\label{sec:post_training_details}
DINOv2R ViT-S/14 and DINOv3 ViT-S/16 are post-trained on the COCO train2017 images with the regional loss as the only objective, on global-to-global and global-to-local view pairs and through a randomly initialised projection head with 8,192 prototypes. Each image supplies two global and four local views, of $518$ and $98$ pixels for ViT-S/14 and of $512$ and $96$ pixels for ViT-S/16. Training uses a global batch of 64 and AdamW with learning rates of $10^{-6}$ for the backbone and $10^{-4}$ for the head, with a 25-epoch cosine schedule stopped after 7 epochs. Clustering follows the protocol above.

\paragraph{Feature analysis.}
The variance of Figure~\ref{fig:feature_analysis}a uses the $\ell_2$-normalised patch features of the VOC2012 validation images at $448\times448$ and the 2,184 objects with at least eight patches of at least 90\% instance purity. Within objects, it is the mean squared distance of the patches to their object mean; between objects, that of the object means to their class mean; and between classes, that of the class means to their overall mean. Values are averaged over the 20 classes, and the error bars give their standard deviation across classes.

\paragraph{Correspondence and depth.}
\label{sec:depth_estimation}\label{sec:correspondence_protocol}
Both follow Probe3D~\cite{elbanani2024probe3d} with raw block outputs taken before the final LayerNorm. For keypoint correspondence on SPair-71k~\cite{min2019spair}, images are resized to $800\times800$ without bounding-box cropping, source keypoints take bilinearly interpolated features from the $\ell_2$-normalised final-block patch maps and are matched to the target location of maximum cosine similarity, and PCK@0.1 is relative to the bounding box, over at most 200 sampled test pairs per category. For depth, only a DPT~\cite{ranftl2021dpt} decoder on blocks $3,6,9,12$ is trained, for 10 epochs with AdamW, an effective batch of 32, learning rate $5\times10^{-4}$ and 1.5 warm-up epochs, and we report the RMSE in metres without test-time scale or shift alignment. NYUv2~\cite{silberman2012nyuv2} uses $480\times480$ centre crops, the GeoNet~\cite{qi2018geonet} training split and the 654 labelled test images. KITTI~\cite{geiger2012kitti} uses the Eigen split lists released by BTS~\cite{lee2019big}, with the 652 test images that have ground truth, images resized to $480\times480$ without rotation or flipping, and an 80\,m depth limit, and predictions are scored at the original depth-map resolution on the valid pixels of the Garg crop.

\section{Additional Experiments}
\label{sec:additional_experiments}

\subsection{Ablations}
\label{sec:appx_ablations}

\begin{table}[t]
\centering
\caption{\textbf{Additional ablations.} VOC $k$-NN and linear (lin.) mIoU and IN1k $k$-NN accuracy (\%, 10\% reference bank) with the protocol of Table~\ref{tab:ablations}, whose reference is shown in blue.}
\label{tab:appx_ablations}\label{tab:prediction_space}
\footnotesize
\setlength{\tabcolsep}{6pt}
\begin{tabular}{lccc}
\toprule
& \multicolumn{2}{c}{VOC} & IN1k \\ \cmidrule(lr){2-3}\cmidrule(l){4-4}
Configuration & $k$-NN & lin. & $k$-NN \\
\midrule
\rowcolor{blue}
\textit{Reference} & \textbf{61.8} & \textbf{66.7} & 69.1 \\
Region token & 55.7 & 63.5 & \textbf{69.6} \\
Unmasked views & 60.3 & 66.3 & 68.5 \\
\bottomrule
\end{tabular}
\end{table}

\paragraph{Region token.}
The composition is a fixed average of patch predictions, with no parameters of its own, which raises the question of whether the aggregation of a region should instead be learned. We test this with a region token (Table~\ref{tab:appx_ablations}). A learnable token and one Transformer block of the same size as those of the backbone are added on top of the backbone, in the student and, through the EMA, in the teacher. In each view of a pair, the token is the only query of this block and attends to the output patch tokens that lie fully inside the shared region, the same patches that form the composition, as in attention pooling~\cite{lee2019settransformer}; it cannot attend to the $[\mathrm{CLS}]$ token or to patches outside the region. After the attention and MLP residuals of the block and a LayerNorm, the shared projection head maps the token to a single prediction over the $K$ prototypes. The regional loss then becomes the cross-entropy between the teacher's region prediction in one view and the student's prediction in the other view, averaged over both directions. The new block is randomly initialised, adds about $1.8$M parameters, or $8\%$ of the backbone, and is discarded after pretraining.

The token learns to predict consistently across views, as its loss decreases from $8.6$ nats in the first epoch to $1.1$ in the last. Dense transfer nevertheless falls $6.1$ and $3.2$ points below the composition in VOC $k$-NN and linear mIoU, although it remains above iBOT ($49.7$ and $58.0$), and ImageNet accuracy is $0.5$ points higher. The region token is also below the $\mathbf q(\mathrm{mean}(\mathbf f_p))$ variant of Table~\ref{tab:ablations}a ($59.9$ and $64.6$), which passes a plain average of the same patch tokens through the head, so learning the aggregation does not help even among the variants that summarise the region by one vector. A plausible reason is that the added block can absorb part of the cross-view agreement, for instance through the patches that the token chooses to read, whereas the regional loss reaches the patch tokens only through this block and no longer constrains their predictions. Consistent with this, the teacher's patch predictions remain less confident than with the composition, with a mean maximum probability of $0.42$ against $0.72$ in the last epoch. The composition leaves no such freedom: every selected patch contributes its own prediction with equal weight, so the agreement must come from the patch predictions themselves, and no parameters are added.

\paragraph{Unmasked views.}
Table~\ref{tab:appx_ablations} also examines whether the regional loss benefits from unmasked student inputs. Instead of the two masked global views, it is applied to four additional global views that the student receives unmasked, over all six pairs of these views. Although this variant processes more views at every step, about doubling the patches of the student and tripling those of the teacher, it lowers all three metrics, suggesting that the regional loss does not require unmasked inputs.

\subsection{Additional Evaluations and Analysis}
\label{sec:appx_benchmarks}

\begin{figure}[t]
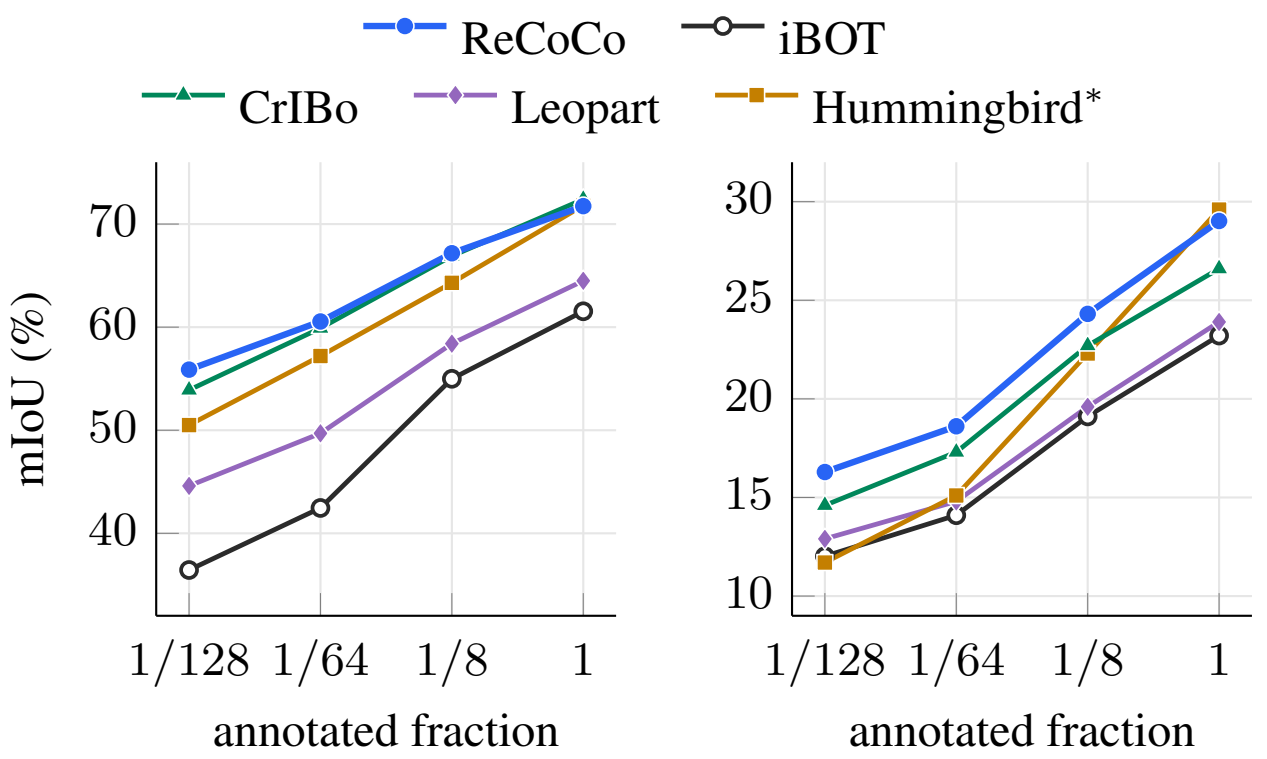

\centering
\iclegend\par\vspace{3pt}
\icchartA\hfill\icchartB
\caption{\textbf{In-context segmentation.} mIoU (\%) with a fixed-size support memory against the fraction of annotated training images, for iBOT~\cite{zhou2022ibot}, Leopart~\cite{ziegler2022leopart}, CrIBo~\cite{lebailly2024cribo}, Hummingbird~\cite{balazevic2023hummingbird} and \methodname{}. $^{*}$ViT-B/16; all other methods use ViT-S/16.}
\label{fig:incontext}\label{tab:neco_dense_retrieval}
\end{figure}

\paragraph{In-context segmentation.}
\label{sec:appx_incontext}
In-context segmentation tests whether labels transfer from a few annotated images through patch retrieval alone, without training any parameters. We follow the Open Hummingbird protocol~\cite{balazevic2023hummingbird}, which labels each validation patch from its 30 nearest neighbours in a fixed-size memory of annotated training patches, and vary the fraction of annotated training images on VOC and ADE20K (Figure~\ref{fig:incontext}). Inputs are $512\times512$, and the scores at partial fractions are averaged over the five released subsets. \methodname{} achieves the best mIoU in all partial-support settings: at $1/128$, it reaches $55.9$ on VOC and $16.3$ on ADE20K, against $53.9$ and $14.6$ for CrIBo and $36.4$ and $12.0$ for iBOT. With full support, it remains close to CrIBo on VOC ($71.7$ against $72.4$) and leads the ViT-S/16 methods on ADE20K ($29.0$ against $26.6$). On VOC, its margin over CrIBo is largest when annotations are scarce, whereas on ADE20K it does not shrink with more annotations.

\input{img/incontext_src/incontext_charts}

\paragraph{End-to-end fine-tuning.}
\label{sec:appx_finetuning}
Pretrained backbones are often used to initialise supervised models. We evaluate this with the end-to-end fine-tuning protocol of CrIBo~\cite{lebailly2024cribo}, which trains the backbone jointly with a Segmenter~\cite{strudel2021segmenter} decoder on PASCAL Context~\cite{mottaghi2014context}, PASCAL VOC, COCO-Stuff~\cite{caesar2018cocostuff} and ADE20K (mIoU, Table~\ref{tab:full_finetuning}). Unlike frozen-feature evaluation, end-to-end performance reflects both the quality of the pretrained representations and the suitability of the fine-tuning recipe. Due to computational constraints, we adopt the existing recipe without model-specific hyperparameter tuning. Even under this single-run setting, \methodname{} achieves the best mIoU on three of the four datasets with ViT-B/16 and remains competitive with iBOT~\cite{zhou2022ibot} with ViT-S/16. These results are encouraging given the absence of fine-tuning hyperparameter optimisation.

\begin{table}[t]
\centering
\caption{\textbf{End-to-end fine-tuning.}
Validation mIoU (\%) after jointly fine-tuning the backbone and Segmenter decoder. We highlight the \textbf{best} and \underline{second-best} number of each column within each backbone. }
\label{tab:full_finetuning}
\footnotesize
\setlength{\tabcolsep}{2.5pt}
\renewcommand{\arraystretch}{1.0}
\begin{adjustbox}{max width=\columnwidth}
\begin{tabular}{lccccc}
\toprule
Method & Backbone & Context & VOC & COCO-Stuff & ADE20K \\
\midrule
DINO~\cite{caron2021dino} & ViT-S/16 & 46.0 & 80.3 & 43.2 & 43.3 \\
CrOC~\cite{stegmuller2023croc} & ViT-S/16 & 46.0 & 80.9 & 42.9 & 42.8 \\
TimeT~\cite{salehi2023timet} & ViT-S/16 & 47.4 & 80.4 & 43.1 & 43.5 \\
CrIBo~\cite{lebailly2024cribo} & ViT-S/16 & \underline{49.3} & \textbf{82.3} & \textbf{43.9} & \underline{45.2} \\
iBOT~\cite{zhou2022ibot} & ViT-S/16 & \textbf{50.18} & \underline{81.87} & \underline{43.86} & \textbf{45.48} \\
\rowcolor{blue}
\textbf{\methodname{}} & ViT-S/16 & 49.06 & 80.12 & 42.96 & 44.41 \\
\textcolor{gray}{DINOv2R$^{*}$~\cite{darcet2024vision}} & \textcolor{gray}{ViT-S/14} & \textcolor{gray}{59.3} & \textcolor{gray}{86.2} & \textcolor{gray}{46.9} & \textcolor{gray}{48.6} \\
\textcolor{gray}{NeCo$^{*}$~\cite{pariza2025neco}} & \textcolor{gray}{ViT-S/14} & \textcolor{gray}{59.7} & \textcolor{gray}{87.0} & \textcolor{gray}{47.3} & \textcolor{gray}{48.9} \\
\midrule
DINO~\cite{caron2021dino} & ViT-B/16 & 45.8 & 82.2 & 44.4 & 45.0 \\
MAE~\cite{he2022mae} & ViT-B/16 & 47.9 & 82.7 & 45.5 & 46.4 \\
CrIBo~\cite{lebailly2024cribo} & ViT-B/16 & 49.2 & 83.4 & 44.6 & 46.0 \\
iBOT~\cite{zhou2022ibot} & ViT-B/16 & \underline{53.19} & \underline{83.49} & \textbf{46.29} & \underline{48.16} \\
\rowcolor{blue}
\textbf{\methodname{}} & ViT-B/16 & \textbf{53.25} & \textbf{83.66} & \underline{46.03} & \textbf{48.67} \\
\textcolor{gray}{DINOv2R$^{*}$~\cite{darcet2024vision}} & \textcolor{gray}{ViT-B/14} & \textcolor{gray}{62.4} & \textcolor{gray}{86.0} & \textcolor{gray}{48.8} & \textcolor{gray}{52.3} \\
\textcolor{gray}{NeCo$^{*}$~\cite{pariza2025neco}} & \textcolor{gray}{ViT-B/14} & \textcolor{gray}{63.0} & \textcolor{gray}{87.7} & \textcolor{gray}{49.3} & \textcolor{gray}{52.5} \\
\bottomrule
\end{tabular}
\end{adjustbox}
\vspace{1mm}
\raggedright
\scriptsize
$^{*}$Pretrained on LVD-142M~\cite{oquab2024dinov2}.
\end{table}

\begin{figure}[t]
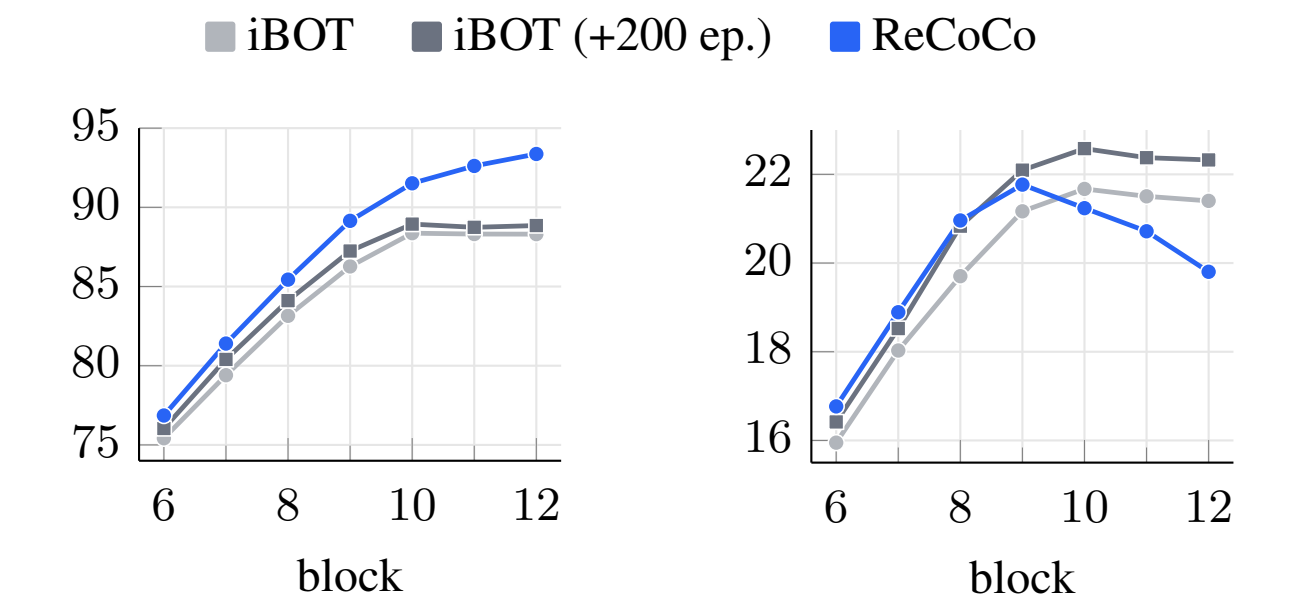

\centering
\depthlegend\par\vspace{3pt}
\depthchartB\depthchartC
\caption{\textbf{Where the organisation forms.} Patch tokens at the output of blocks 6 to 12 of iBOT, iBOT (+200 ep.) and \methodname{} with ViT-S/16, each passed through the final LayerNorm. (a) Share (\%) of the ten most similar patches from other images that have the class of the query patch (20,000 VOC queries of at least 90\% class purity, $448\times448$). (b) Keypoint correspondence on SPair-71k~\cite{min2019spair}, PCK@0.1 in \%.}
\label{fig:depth}
\end{figure}

\paragraph{Where the organisation forms.}
\label{sec:appx_depth}
To locate where the regional loss acts in the backbone, Figure~\ref{fig:depth} evaluates the patch tokens at the output of blocks 6 to 12 with two measures that require no training. The share of the nearest patches in other images that have the class of the query reflects the semantic agreement on which $k$-NN segmentation relies, and keypoint correspondence reflects how well the positions inside an object remain distinguishable. Up to block 8, \methodname{} is on par with iBOT (+200 ep.) on both, so the regional loss preserves the earlier blocks and acts on the last ones. There, it continues the semantic organisation that iBOT stops building: same-class neighbours saturate at 88--89\% from block 10 in both iBOT models, whereas in \methodname{} they increase with every block, to $93.4\%$ at the output. Correspondence peaks at block 9, where \methodname{} matches the best block of the released iBOT ($21.8$ against $21.7$), and decreases by $2.0$ points towards the output. The lower correspondence of Figure~\ref{fig:feature_analysis}b is thus confined to the last blocks, and geometry-sensitive tasks can read the intermediate ones, as the depth decoder does. The regional loss therefore adds semantic organisation in the output blocks while the representation that precedes them is preserved.

\section{Additional Visualisations}
\label{sec:visual_examples}

\begin{figure*}[t]
\centering
\includegraphics[width=\textwidth]{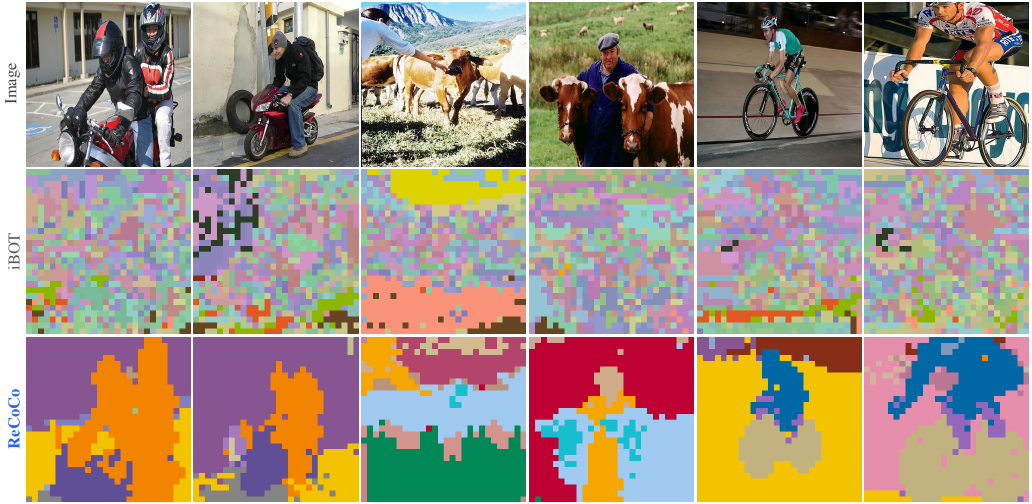}
\caption{\textbf{Prototype maps.} Each patch is coloured by its most probable teacher prototype, with one colour per prototype across the figure, for $448\times448$ inputs. Each pair of columns shows two scenes whose classes share the main prototypes of \methodname{}.}
\label{fig:gallery_prototypes}
\end{figure*}

\begin{figure*}[t]
\centering
\includegraphics[width=\textwidth]{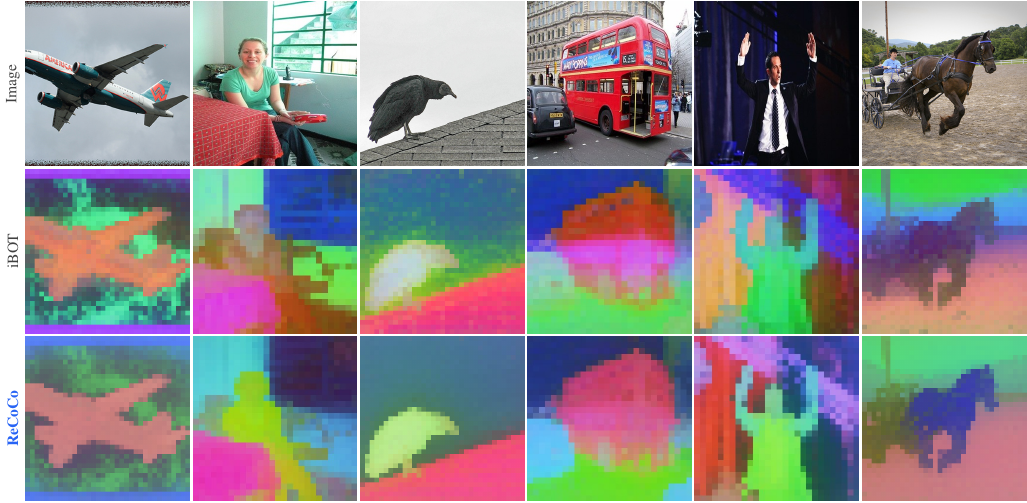}
\caption{\textbf{PCA of patch features.} For each image, we show the RGB projection of the final-block patch tokens onto their first three principal components, computed individually on each image and model, with the components of \methodname{} ordered and signed to match those of iBOT. Columns 1, 3 and 6 are ImageNet validation images, and the VOC images use a whitened PCA.}
\label{fig:gallery_pca}
\end{figure*}

\begin{figure*}[t]
\centering
\includegraphics[width=\textwidth]{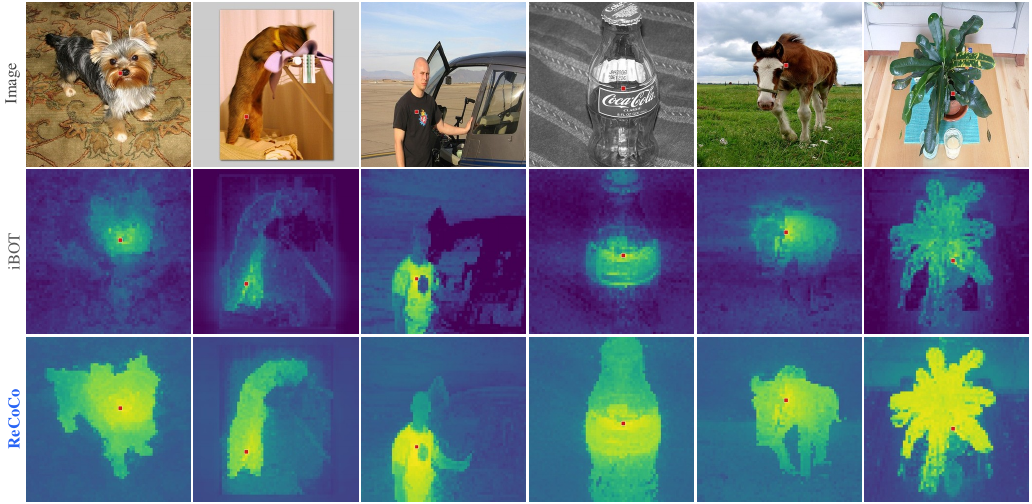}
\caption{\textbf{Similarity to a query patch.} Cosine similarity between the final-block patch tokens and the token of the query patch (red) for $1024\times1024$ inputs. The query is the most interior patch of the largest object of a class, and the colour scale runs from 0 (dark blue) to 1 (yellow).}
\label{fig:gallery_similarity}
\end{figure*}

\begin{figure*}[t]
\centering
\includegraphics[width=\textwidth]{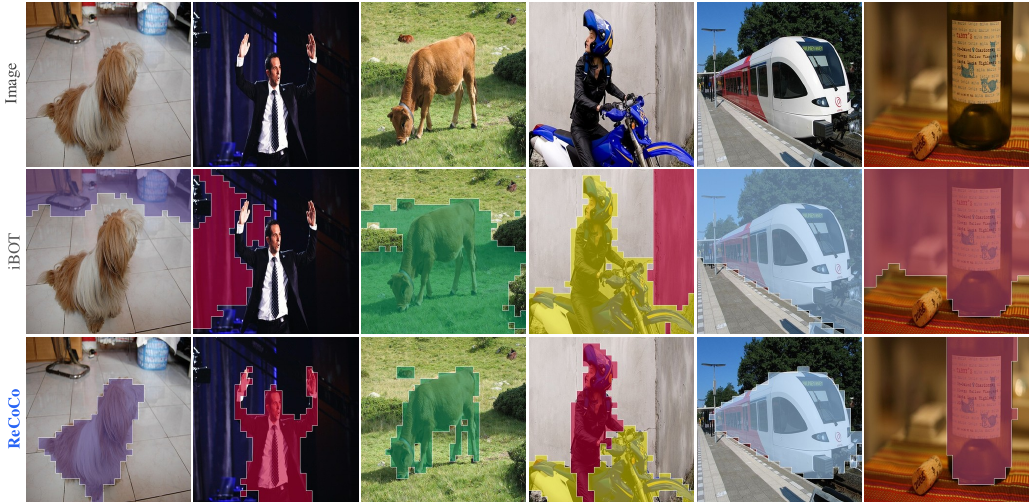}
\caption{\textbf{$k$-means clustering.} The final-block patch tokens of each image are clustered into as many groups as the image has labels, background included. Clusters are matched to the labels with the Hungarian algorithm and tinted with their class colour, except those matched to background.}
\label{fig:gallery_kmeans}
\end{figure*}

\begin{figure*}[t]
\centering
\includegraphics[width=\textwidth]{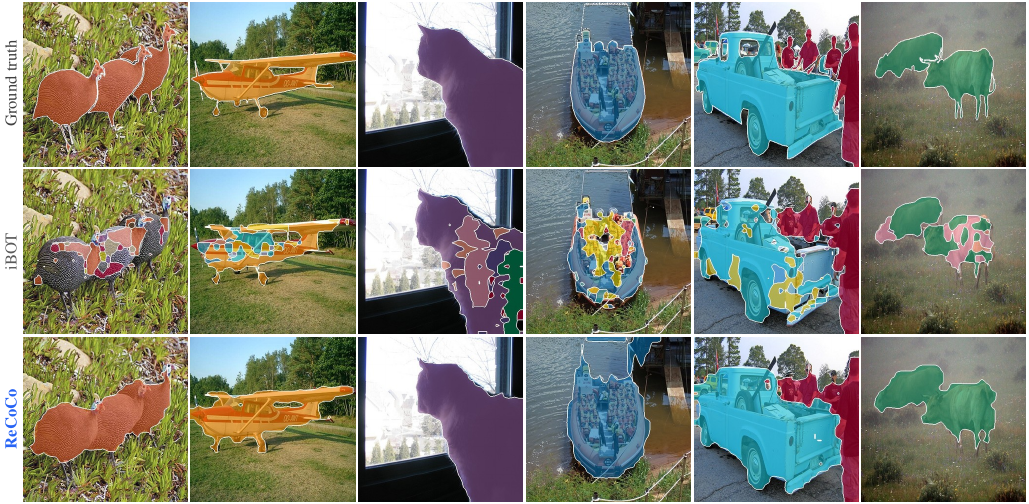}
\caption{\textbf{$k$-NN label transfer.} Each patch takes the label favoured by its 20 most similar patches of the VOC training images, weighted by $\exp(\cos/0.07)$, for $448\times448$ inputs, a simplified version of the $k$-NN probe. The top row shows the ground truth.}
\label{fig:gallery_knn}
\end{figure*}

We show visualisations of the released iBOT and \methodname{} ViT-S/16 teachers in Figures~\ref{fig:gallery_prototypes} to~\ref{fig:gallery_knn}. Each column shows an input image above the visualisations of both models. All visualisations are computed on the patch grid, from the final-block patch tokens or, for the prototype maps, from their predictions. Unless a caption states otherwise, the images are VOC validation images on which \methodname{} improves most over iBOT for the visualised readout.

%% file: img/analysis_src/analysis_charts.tex
\definecolor{anreco}{HTML}{2864F5}
\definecolor{ancon}{HTML}{6B7280}
\definecolor{anibot}{HTML}{B1B5BB}
\providecommand{\anbox}[1]{\tikz[baseline=-0.6ex]\fill[#1, rounded corners=0.6pt] (0,-0.1) rectangle (0.2,0.1);}
\pgfplotsset{
  an axis/.style={
    scale only axis, width=\anaxis, height=\anaxish,
    /pgf/number format/1000 sep={},
    ymajorgrids, xmajorgrids, grid style={black!10},
    axis x line*=bottom, axis y line*=left,
    every axis plot/.append style={line width=0.9pt, line join=round},
    xlabel style={align=center, font=\small, yshift=3pt}, ylabel style={font=\small},
    tick label style={font=\small}, typeset ticklabels with strut,
  },
  an dyn/.style={an axis, xmin=0, xmax=200, xtick={0,100,200}, every axis plot/.append style={mark=none}},
  an lin/.style={dash pattern=on 2.8pt off 1.6pt},
  an start/.style={only marks, forget plot, mark=*, mark size=1.6pt, mark options={fill=black, draw=white, line width=0.3pt}},
  an blk/.style={an axis, xmin=5.6, xmax=12.4, xtick={6,8,10,12}, minor xtick={7,9,11}, minor grid style={black!10}, xminorgrids,
    yticklabel style={text width=2.1em, align=right}},
  an mk/.style={mark=*, mark size=1.5pt, mark options={solid, fill=#1, draw=white, line width=0.3pt}},
  an mksq/.style={mark=square*, mark size=1.4pt, mark options={solid, fill=#1, draw=white, line width=0.3pt}},
}
\ifdefined\anaxis\else\newlength{\anaxis}\newlength{\anaxish}\fi
\setlength{\anaxis}{.335\columnwidth}
\setlength{\anaxish}{2.2cm}
\newcommand{\dynlegend}{{\small\anbox{anreco}~\methodname{}\hspace{1.2em}\anbox{ancon}~iBOT (+200 ep.)\hspace{1.2em}%
\tikz[baseline=-0.6ex]\fill[black] (0,0) circle (1.6pt);~released iBOT}}
\newcommand{\depthlegend}{{\small\anbox{anibot}~iBOT\hspace{1.2em}\anbox{ancon}~iBOT (+200 ep.)\hspace{1.2em}\anbox{anreco}~\methodname{}}}
\newcommand{\dynchartA}{%
\begin{tikzpicture}
\begin{axis}[an dyn, ymin=45, ymax=75, ytick={50,60,70},
  xlabel={epoch\\[1pt](a) VOC segmentation}, ylabel={mIoU (\%)}]
\addplot[ancon] coordinates {(0,49.7) (10,49.437) (20,49.648) (30,49.823) (40,50.095) (50,50.367) (60,50.426) (70,50.572) (80,50.914) (90,50.859) (100,50.98) (110,50.966) (120,50.879) (130,50.849) (140,50.985) (160,50.964) (170,50.974) (180,50.981) (190,50.966) (200,50.961)};
\addplot[ancon, an lin] coordinates {(0,58.0) (10,57.703) (20,58.113) (30,58.702) (40,58.777) (50,59.31) (60,59.347) (70,59.591) (80,59.636) (90,59.702) (100,59.686) (110,59.725) (120,59.614) (130,59.426) (140,59.345) (160,59.254) (170,59.252) (180,59.254) (190,59.251) (200,59.241)};
\addplot[anreco] coordinates {(0,49.7) (5,58.056) (10,59.764) (15,60.855) (20,61.659) (25,62.069) (30,62.553) (35,62.69) (40,62.637) (45,63.215) (50,63.508) (55,63.367) (60,63.742) (65,63.642) (70,63.519) (75,63.77) (80,63.883) (85,63.859) (90,63.761) (95,63.874) (100,63.679) (105,63.767) (110,63.838) (115,63.921) (120,63.967) (125,64.2) (130,64.181) (135,64.244) (140,64.148) (145,64.187) (150,64.239) (155,64.163) (160,64.163) (165,64.246) (170,64.277) (175,64.249) (180,64.257) (185,64.281) (190,64.313) (195,64.309) (200,64.306)};
\addplot[anreco, an lin] coordinates {(0,58.0) (5,64.128) (10,65.291) (15,66.159) (20,66.448) (25,66.516) (30,66.938) (35,67.169) (40,67.339) (45,67.622) (50,67.589) (55,67.476) (60,67.907) (65,67.926) (70,67.93) (75,68.09) (80,68.186) (85,68.257) (90,68.358) (95,68.205) (100,68.251) (105,68.394) (110,68.511) (115,68.68) (120,68.753) (125,68.802) (130,68.75) (135,68.673) (140,68.618) (145,68.643) (150,68.842) (155,68.898) (160,68.87) (165,68.866) (170,68.861) (175,68.811) (180,68.774) (185,68.795) (190,68.812) (195,68.805) (200,68.8)};
\addplot[an start] coordinates {(0,49.7) (0,58.0)};
\node[font=\small, anchor=south, inner sep=1pt] at (axis cs:110,70.5) {linear};
\node[font=\small, anchor=north, inner sep=1pt] at (axis cs:110,50.2) {$k$-NN};
\end{axis}
\end{tikzpicture}}
\newcommand{\dynchartB}{%
\begin{tikzpicture}
\begin{axis}[an dyn, ymin=68.5, ymax=70.5, ytick={69,70},
  xlabel={epoch\\[1pt](b) IN1k $k$-NN (\%)}]
\addplot[ancon] coordinates {(0,70.09) (10,69.81) (20,69.404) (30,69.114) (40,69.052) (50,69.116) (60,69.062) (70,68.96) (80,69.116) (90,69.254) (100,69.184) (110,69.352) (120,69.506) (130,69.542) (140,69.684) (160,69.768) (170,69.796) (180,69.804) (190,69.848) (200,69.864)};
\addplot[anreco] coordinates {(0,70.09) (5,69.826) (10,69.654) (15,69.634) (20,69.652) (25,69.604) (30,69.658) (35,69.674) (40,69.606) (45,69.642) (50,69.55) (55,69.56) (60,69.518) (65,69.49) (70,69.542) (75,69.612) (80,69.536) (85,69.618) (90,69.618) (95,69.666) (100,69.692) (105,69.744) (110,69.786) (115,69.914) (120,69.912) (125,70.004) (130,70.014) (135,70.036) (140,70.148) (145,70.186) (150,70.19) (155,70.148) (160,70.266) (165,70.238) (170,70.248) (175,70.256) (180,70.264) (185,70.29) (190,70.308) (195,70.3) (200,70.286)};
\addplot[an start] coordinates {(0,70.09)};
\end{axis}
\end{tikzpicture}}
\newcommand{\dynchartC}{%
\begin{tikzpicture}
\begin{axis}[an dyn, ymin=0, ymax=4, ytick={0,1,2,3,4},
  xlabel={epoch\\[1pt](c) Masked-patch loss}, ylabel={cross-entropy}]
\addplot[ancon] coordinates {(1,3.6233) (3,3.6824) (5,3.7008) (7,3.6655) (9,3.5971) (11,3.4898) (13,3.3717) (15,3.2561) (17,3.1567) (19,3.0388) (21,2.9525) (23,2.8667) (25,2.7937) (27,2.7493) (29,2.7023) (31,2.6522) (33,2.6252) (35,2.5916) (37,2.5635) (39,2.5288) (41,2.5109) (43,2.4973) (45,2.4727) (47,2.464) (49,2.4523) (51,2.4601) (53,2.4441) (55,2.439) (57,2.4279) (59,2.4322) (61,2.4182) (63,2.4132) (65,2.42) (67,2.4158) (69,2.4126) (71,2.4172) (73,2.4192) (75,2.4386) (77,2.4405) (79,2.4492) (81,2.4541) (83,2.4783) (85,2.4821) (87,2.4936) (89,2.5066) (91,2.5257) (93,2.5317) (95,2.5596) (97,2.5712) (99,2.5861) (101,2.6132) (103,2.6312) (105,2.6554) (107,2.6709) (109,2.6922) (111,2.7114) (113,2.723) (115,2.7439) (117,2.7717) (119,2.802) (121,2.8233) (123,2.853) (125,2.8647) (127,2.8945) (129,2.9164) (131,2.9283) (133,2.9511) (135,2.9657) (137,2.9892) (139,3.0122) (141,3.0246) (143,3.0493) (145,3.065) (147,3.0773) (149,3.0877) (151,3.1014) (153,3.1161) (155,3.1344) (157,3.1394) (159,3.1542) (161,3.1633) (163,3.1688) (165,3.1803) (167,3.1925) (169,3.1911) (171,3.2031) (173,3.1995) (175,3.2018) (177,3.2128) (179,3.2152) (181,3.2111) (183,3.2216) (185,3.2253) (187,3.2285) (189,3.2275) (191,3.2262) (193,3.2351) (195,3.2264) (197,3.2325) (199,3.2302)};
\addplot[anreco] coordinates {(1,3.4075) (3,2.0329) (5,1.5819) (7,1.3652) (9,1.2268) (11,1.133) (13,1.0609) (15,1.0098) (17,0.9698) (19,0.9386) (21,0.9121) (23,0.8914) (25,0.8709) (27,0.8554) (29,0.842) (31,0.8254) (33,0.816) (35,0.804) (37,0.7955) (39,0.7861) (41,0.7792) (43,0.7745) (45,0.7737) (47,0.7733) (49,0.7735) (51,0.7675) (53,0.7643) (55,0.7615) (57,0.7573) (59,0.75) (61,0.7454) (63,0.7418) (65,0.7393) (67,0.7382) (69,0.7389) (71,0.7393) (73,0.7401) (75,0.7402) (77,0.738) (79,0.7371) (81,0.7336) (83,0.7306) (85,0.7287) (87,0.7295) (89,0.7331) (91,0.7362) (93,0.7399) (95,0.7412) (97,0.7424) (99,0.7453) (101,0.7497) (103,0.7491) (105,0.751) (107,0.7516) (109,0.7515) (111,0.7512) (113,0.7515) (115,0.7524) (117,0.7518) (119,0.7542) (121,0.7567) (123,0.7611) (125,0.764) (127,0.767) (129,0.7711) (131,0.7772) (133,0.7793) (135,0.7824) (137,0.7869) (139,0.7869) (141,0.7887) (143,0.7893) (145,0.7906) (147,0.7914) (149,0.7914) (151,0.7916) (153,0.7934) (155,0.7949) (157,0.7971) (159,0.7981) (161,0.8003) (163,0.8027) (165,0.8046) (167,0.806) (169,0.8095) (171,0.8118) (173,0.813) (175,0.8148) (177,0.8175) (179,0.8192) (181,0.8213) (183,0.8225) (185,0.8224) (187,0.8272) (189,0.8298) (191,0.8314) (193,0.8298) (195,0.8335) (197,0.8313) (199,0.8357)};
\end{axis}
\end{tikzpicture}}
\newcommand{\dynlegendtwo}{{\small\anbox{anreco}~\methodname{}\hspace{1.2em}\anbox{ancon}~iBOT (+200 ep.)}}
\newcommand{\dynlossPatch}{%
\begin{tikzpicture}
\begin{axis}[an dyn, ymin=0, ymax=4, ytick={0,1,2,3,4},
  xlabel={epoch\\[1pt](a) Masked-patch loss}, ylabel={cross-entropy}]
\addplot[ancon] coordinates {(1,3.6233) (3,3.6824) (5,3.7008) (7,3.6655) (9,3.5971) (11,3.4898) (13,3.3717) (15,3.2561) (17,3.1567) (19,3.0388) (21,2.9525) (23,2.8667) (25,2.7937) (27,2.7493) (29,2.7023) (31,2.6522) (33,2.6252) (35,2.5916) (37,2.5635) (39,2.5288) (41,2.5109) (43,2.4973) (45,2.4727) (47,2.4640) (49,2.4523) (51,2.4601) (53,2.4441) (55,2.4390) (57,2.4279) (59,2.4322) (61,2.4182) (63,2.4132) (65,2.4200) (67,2.4158) (69,2.4126) (71,2.4172) (73,2.4192) (75,2.4386) (77,2.4405) (79,2.4492) (81,2.4541) (83,2.4783) (85,2.4821) (87,2.4936) (89,2.5066) (91,2.5257) (93,2.5317) (95,2.5596) (97,2.5712) (99,2.5861) (101,2.6132) (103,2.6312) (105,2.6554) (107,2.6709) (109,2.6922) (111,2.7114) (113,2.7230) (115,2.7439) (117,2.7717) (119,2.8020) (121,2.8233) (123,2.8530) (125,2.8647) (127,2.8945) (129,2.9164) (131,2.9283) (133,2.9511) (135,2.9657) (137,2.9892) (139,3.0122) (141,3.0246) (143,3.0493) (145,3.0650) (147,3.0773) (149,3.0877) (151,3.1014) (153,3.1161) (155,3.1344) (157,3.1394) (159,3.1542) (161,3.1633) (163,3.1688) (165,3.1803) (167,3.1925) (169,3.1911) (171,3.2031) (173,3.1995) (175,3.2018) (177,3.2128) (179,3.2152) (181,3.2111) (183,3.2216) (185,3.2253) (187,3.2285) (189,3.2275) (191,3.2262) (193,3.2351) (195,3.2264) (197,3.2325) (199,3.2302) (200,3.2248)};
\addplot[anreco] coordinates {(1,3.4075) (3,2.0329) (5,1.5819) (7,1.3652) (9,1.2268) (11,1.1330) (13,1.0609) (15,1.0098) (17,0.9698) (19,0.9386) (21,0.9121) (23,0.8914) (25,0.8709) (27,0.8554) (29,0.8420) (31,0.8254) (33,0.8160) (35,0.8040) (37,0.7955) (39,0.7861) (41,0.7792) (43,0.7745) (45,0.7737) (47,0.7733) (49,0.7735) (51,0.7675) (53,0.7643) (55,0.7615) (57,0.7573) (59,0.7500) (61,0.7454) (63,0.7418) (65,0.7393) (67,0.7382) (69,0.7389) (71,0.7393) (73,0.7401) (75,0.7402) (77,0.7380) (79,0.7371) (81,0.7336) (83,0.7306) (85,0.7287) (87,0.7295) (89,0.7331) (91,0.7362) (93,0.7399) (95,0.7412) (97,0.7424) (99,0.7453) (101,0.7497) (103,0.7491) (105,0.7510) (107,0.7516) (109,0.7515) (111,0.7512) (113,0.7515) (115,0.7524) (117,0.7518) (119,0.7542) (121,0.7567) (123,0.7611) (125,0.7640) (127,0.7670) (129,0.7711) (131,0.7772) (133,0.7793) (135,0.7824) (137,0.7869) (139,0.7869) (141,0.7887) (143,0.7893) (145,0.7906) (147,0.7914) (149,0.7914) (151,0.7916) (153,0.7934) (155,0.7949) (157,0.7971) (159,0.7981) (161,0.8003) (163,0.8027) (165,0.8046) (167,0.8060) (169,0.8095) (171,0.8118) (173,0.8130) (175,0.8148) (177,0.8175) (179,0.8192) (181,0.8213) (183,0.8225) (185,0.8224) (187,0.8272) (189,0.8298) (191,0.8314) (193,0.8298) (195,0.8335) (197,0.8313) (199,0.8357) (200,0.8336)};
\end{axis}
\end{tikzpicture}}
\newcommand{\dynlossCls}{%
\begin{tikzpicture}
\begin{axis}[an dyn, ymin=0, ymax=4, ytick={0,1,2,3,4},
  xlabel={epoch\\[1pt](b) Image-level loss}]
\addplot[ancon] coordinates {(1,3.0170) (3,2.9814) (5,2.9498) (7,2.9244) (9,2.9026) (11,2.8835) (13,2.8705) (15,2.8599) (17,2.8488) (19,2.8422) (21,2.8351) (23,2.8242) (25,2.8171) (27,2.8124) (29,2.8060) (31,2.7991) (33,2.7923) (35,2.7868) (37,2.7827) (39,2.7778) (41,2.7741) (43,2.7693) (45,2.7654) (47,2.7613) (49,2.7563) (51,2.7500) (53,2.7466) (55,2.7426) (57,2.7393) (59,2.7387) (61,2.7367) (63,2.7336) (65,2.7303) (67,2.7279) (69,2.7251) (71,2.7237) (73,2.7243) (75,2.7238) (77,2.7230) (79,2.7221) (81,2.7202) (83,2.7213) (85,2.7206) (87,2.7217) (89,2.7222) (91,2.7239) (93,2.7239) (95,2.7237) (97,2.7249) (99,2.7266) (101,2.7288) (103,2.7295) (105,2.7326) (107,2.7346) (109,2.7359) (111,2.7390) (113,2.7410) (115,2.7432) (117,2.7477) (119,2.7502) (121,2.7518) (123,2.7547) (125,2.7579) (127,2.7600) (129,2.7641) (131,2.7655) (133,2.7690) (135,2.7713) (137,2.7725) (139,2.7762) (141,2.7777) (143,2.7790) (145,2.7812) (147,2.7838) (149,2.7846) (151,2.7880) (153,2.7888) (155,2.7903) (157,2.7909) (159,2.7932) (161,2.7942) (163,2.7960) (165,2.7973) (167,2.7977) (169,2.7985) (171,2.7987) (173,2.7998) (175,2.8014) (177,2.8012) (179,2.8022) (181,2.8029) (183,2.8025) (185,2.8037) (187,2.8042) (189,2.8053) (191,2.8054) (193,2.8059) (195,2.8071) (197,2.8072) (199,2.8072) (200,2.8092)};
\addplot[anreco] coordinates {(1,2.9819) (3,2.9846) (5,2.9679) (7,2.9347) (9,2.8974) (11,2.8541) (13,2.8137) (15,2.7898) (17,2.7734) (19,2.7604) (21,2.7518) (23,2.7386) (25,2.7288) (27,2.7248) (29,2.7171) (31,2.7084) (33,2.7004) (35,2.6793) (37,2.6596) (39,2.6426) (41,2.6256) (43,2.6133) (45,2.5996) (47,2.5786) (49,2.5522) (51,2.5252) (53,2.5008) (55,2.4743) (57,2.4505) (59,2.4229) (61,2.3999) (63,2.3791) (65,2.3595) (67,2.3395) (69,2.3170) (71,2.3000) (73,2.2802) (75,2.2619) (77,2.2469) (79,2.2332) (81,2.2266) (83,2.2207) (85,2.2174) (87,2.2107) (89,2.2012) (91,2.1949) (93,2.1887) (95,2.1864) (97,2.1846) (99,2.1876) (101,2.1937) (103,2.1958) (105,2.1988) (107,2.2038) (109,2.2111) (111,2.2210) (113,2.2264) (115,2.2307) (117,2.2358) (119,2.2399) (121,2.2420) (123,2.2446) (125,2.2479) (127,2.2518) (129,2.2568) (131,2.2618) (133,2.2672) (135,2.2701) (137,2.2745) (139,2.2801) (141,2.2886) (143,2.2951) (145,2.3002) (147,2.3044) (149,2.3081) (151,2.3135) (153,2.3154) (155,2.3171) (157,2.3184) (159,2.3185) (161,2.3211) (163,2.3215) (165,2.3237) (167,2.3271) (169,2.3278) (171,2.3310) (173,2.3327) (175,2.3360) (177,2.3386) (179,2.3407) (181,2.3449) (183,2.3459) (185,2.3490) (187,2.3513) (189,2.3553) (191,2.3573) (193,2.3602) (195,2.3643) (197,2.3666) (199,2.3697) (200,2.3709)};
\end{axis}
\end{tikzpicture}}
\newcommand{\dynlossReg}{%
\begin{tikzpicture}
\begin{axis}[an dyn, ymin=0, ymax=8, ytick={0,2,4,6,8},
  xlabel={epoch\\[1pt](c) Regional loss}, ylabel={cross-entropy}]
\addplot[anreco] coordinates {(1,7.7285) (3,4.9207) (5,4.0065) (7,3.5682) (9,3.2953) (11,3.1084) (13,2.9679) (15,2.8657) (17,2.7884) (19,2.7264) (21,2.6757) (23,2.6317) (25,2.5938) (27,2.5631) (29,2.5336) (31,2.5051) (33,2.4844) (35,2.4614) (37,2.4412) (39,2.4230) (41,2.4095) (43,2.3990) (45,2.3963) (47,2.3945) (49,2.3896) (51,2.3827) (53,2.3772) (55,2.3692) (57,2.3561) (59,2.3427) (61,2.3316) (63,2.3226) (65,2.3184) (67,2.3155) (69,2.3135) (71,2.3134) (73,2.3124) (75,2.3099) (77,2.3056) (79,2.2995) (81,2.2916) (83,2.2836) (85,2.2752) (87,2.2752) (89,2.2775) (91,2.2820) (93,2.2844) (95,2.2855) (97,2.2862) (99,2.2897) (101,2.2919) (103,2.2920) (105,2.2918) (107,2.2907) (109,2.2863) (111,2.2863) (113,2.2840) (115,2.2818) (117,2.2795) (119,2.2806) (121,2.2832) (123,2.2862) (125,2.2904) (127,2.2959) (129,2.3010) (131,2.3070) (133,2.3119) (135,2.3164) (137,2.3204) (139,2.3224) (141,2.3235) (143,2.3231) (145,2.3231) (147,2.3236) (149,2.3228) (151,2.3236) (153,2.3238) (155,2.3250) (157,2.3273) (159,2.3307) (161,2.3340) (163,2.3375) (165,2.3408) (167,2.3445) (169,2.3481) (171,2.3514) (173,2.3545) (175,2.3576) (177,2.3612) (179,2.3630) (181,2.3669) (183,2.3687) (185,2.3723) (187,2.3743) (189,2.3771) (191,2.3794) (193,2.3820) (195,2.3843) (197,2.3860) (199,2.3880) (200,2.3888)};
\end{axis}
\end{tikzpicture}}
\newcommand{\depthchartA}{%
\begin{tikzpicture}
\begin{axis}[an blk, ymin=270, ymax=350, ytick={280,300,320,340},
  xlabel={block\\[1pt](a) Effective rank}]
\addplot[anibot, an mk=anibot] coordinates {(6,317.98) (7,328.33) (8,335.64) (9,341.05) (10,344.09) (11,340.22) (12,331.97)};
\addplot[ancon, an mksq=ancon] coordinates {(6,318.51) (7,328.5) (8,335.02) (9,340.16) (10,342.57) (11,336.07) (12,323.81)};
\addplot[anreco, an mk=anreco] coordinates {(6,319.4) (7,328.17) (8,333.76) (9,336.2) (10,334.82) (11,324.61) (12,274.67)};
\end{axis}
\end{tikzpicture}}
\newcommand{\depthchartB}{%
\begin{tikzpicture}
\begin{axis}[an blk, ymin=74, ymax=95, ytick={75,80,85,90,95},
  xlabel={block\\[1pt](b) Same-class neighbours (\%)}]
\addplot[anibot, an mk=anibot] coordinates {(6,75.421) (7,79.4015) (8,83.151) (9,86.2695) (10,88.3685) (11,88.3135) (12,88.3075)};
\addplot[ancon, an mksq=ancon] coordinates {(6,76.0365) (7,80.3985) (8,84.11) (9,87.2425) (10,88.941) (11,88.7385) (12,88.85)};
\addplot[anreco, an mk=anreco] coordinates {(6,76.857) (7,81.3995) (8,85.436) (9,89.1485) (10,91.5175) (11,92.6135) (12,93.369)};
\end{axis}
\end{tikzpicture}}
\newcommand{\depthchartC}{%
\begin{tikzpicture}
\begin{axis}[an blk, ymin=15.5, ymax=23, ytick={16,18,20,22},
  xlabel={block\\[1pt](c) SPair-71k PCK@0.1 (\%)}]
\addplot[anibot, an mk=anibot] coordinates {(6,15.9501) (7,18.0282) (8,19.7038) (9,21.1667) (10,21.6714) (11,21.5044) (12,21.4017)};
\addplot[ancon, an mksq=ancon] coordinates {(6,16.4174) (7,18.5228) (8,20.8314) (9,22.0947) (10,22.5804) (11,22.3758) (12,22.3252)};
\addplot[anreco, an mk=anreco] coordinates {(6,16.7675) (7,18.8909) (8,20.9602) (9,21.7683) (10,21.2374) (11,20.7164) (12,19.8029)};
\end{axis}
\end{tikzpicture}}
\newcommand{\depthchartD}{%
\begin{tikzpicture}
\begin{axis}[an blk, ymin=0, ymax=0.2, ytick={0,0.05,0.1,0.15,0.2}, yticklabel style={/pgf/number format/fixed, /pgf/number format/precision=2},
  xlabel={block\\[1pt](d) Patch--$[\mathrm{CLS}]$ cosine}]
\addplot[anibot, an mk=anibot] coordinates {(6,0.0285) (7,0.0402) (8,0.0515) (9,0.0544) (10,0.0623) (11,0.0531) (12,0.0708)};
\addplot[ancon, an mksq=ancon] coordinates {(6,0.0253) (7,0.0431) (8,0.0509) (9,0.0391) (10,0.0554) (11,0.0215) (12,0.0444)};
\addplot[anreco, an mk=anreco] coordinates {(6,0.017) (7,0.0552) (8,0.0731) (9,0.0725) (10,0.1373) (11,0.182) (12,0.1885)};
\end{axis}
\end{tikzpicture}}

%% file: img/scratch_src/scratch_charts.tex
\definecolor{cscr}{HTML}{E8710A}
\definecolor{ccon}{HTML}{2864F5}
\definecolor{cibot}{HTML}{6B7280}
\ifdefined\scrWl\else\newlength{\scrWl}\newlength{\scrWr}\fi
\setlength{\scrWl}{0.6667\dimexpr\columnwidth-1.72cm\relax}
\setlength{\scrWr}{0.3333\dimexpr\columnwidth-1.72cm\relax}
\pgfplotsset{
  scr axis/.style={
    scale only axis,
    /pgf/number format/1000 sep={},
    xmajorgrids, ymajorgrids, grid style={black!12},
    axis x line*=bottom, axis y line*=left, axis line style={black!70}, tick style={black!70},
    major tick length=2.5pt,
    every axis plot/.append style={line width=1pt, mark=none, line join=round},
    xlabel style={align=center, font=\small, yshift=3pt}, ylabel style={font=\small}, ylabel shift=-2pt,
    tick label style={font=\small}, typeset ticklabels with strut,
  },
  scr right/.style={
    axis y line*=right, y axis line style={draw=none}, major y tick style={draw=none}, yticklabels={},
    clip=false,
  },
}
\tikzset{
  scr stage/.style={black!30, line width=0.6pt},
  scr ref/.style={black!45, densely dotted, line width=0.7pt},
  scr mark/.style={only marks, mark=*, mark size=1.8pt, black, mark options={draw=white, line width=0.4pt}},
  scr break/.style={black!80, line width=0.7pt},
  scr lin/.style={dash pattern=on 2.8pt off 1.6pt},
  scr head/.style={font=\footnotesize, text=black!60, anchor=south, inner sep=2pt},
  scr note/.style={font=\scriptsize, text=black!60, inner sep=1.5pt},
  scr curve/.style={font=\scriptsize, inner sep=1pt},
}
\providecommand{\scrkey}[1]{\tikz[baseline=-0.6ex]\draw[#1] (0,0)--(0.42,0);}
\providecommand{\scrbox}[1]{\tikz[baseline=-0.6ex]\fill[#1, rounded corners=0.6pt] (0,-0.1) rectangle (0.2,0.1);}
{\footnotesize
\scrbox{cscr}~\methodname{} (scratch)\hspace{0.9em}%
\scrbox{ccon}~\methodname{} (from iBOT)\hspace{0.9em}%
\scrbox{cibot}~iBOT (+200 ep.)\\[1pt]
\scrkey{black!80, line width=1pt}~$k$-NN\hspace{0.9em}%
\scrkey{black!80, line width=1pt, scr lin}~linear\hspace{0.9em}%
\tikz[baseline=-0.6ex]{\draw[scr ref] (0,0)--(0.42,0); \fill[black] (0.21,0) circle (1.8pt);}~released iBOT}\\[2pt]
\begin{tikzpicture}
\begin{axis}[scr axis, name=vl, width=\scrWl, height=3.2cm, xmin=0, xmax=400,
  xtick={0,100,200,300,400}, xticklabels={}, ymin=13, ymax=72, ytick={20,40,60},
  ylabel={VOC mIoU (\%)}]
\draw[scr stage] (axis cs:30,13) -- (axis cs:30,72);
\draw[scr stage] (axis cs:60,13) -- (axis cs:60,72);
\node[font=\scriptsize, text=black!60, rotate=90, anchor=north west, inner sep=1.5pt] at (axis cs:30,13) {$+\mathcal L_{\mathrm{patch}}$};
\node[font=\scriptsize, text=black!60, rotate=90, anchor=north west, inner sep=1.5pt] at (axis cs:60,13) {$+\mathcal L_{\mathrm{region}}$};
\draw[scr ref] (axis cs:0,49.7) -- (axis cs:400,49.7);
\draw[scr ref] (axis cs:0,58.0) -- (axis cs:400,58.0);
\node[scr note, anchor=north east] at (axis cs:400,58.0) {iBOT linear};
\node[scr note, anchor=north east] at (axis cs:400,49.7) {iBOT $k$-NN};
\addplot[cscr] coordinates {(10,19.95) (20,31.57) (30,35.57) (40,45.83) (50,46.60) (60,46.79) (70,56.24) (80,58.02) (90,58.75) (100,59.33) (110,59.68) (120,59.39) (130,59.81) (140,60.23) (150,59.80) (160,59.97) (170,60.61) (180,60.22) (190,60.39) (200,60.29) (210,60.38) (220,60.21) (230,60.32) (240,60.54) (250,60.37) (260,60.27) (270,60.24) (280,60.08) (290,60.51) (300,60.54) (310,60.18) (320,60.51) (330,60.22) (340,60.08) (350,60.25) (360,60.20) (370,60.20) (380,60.48) (390,60.63) (400,60.21)};
\addplot[cscr, scr lin] coordinates {(10,32.28) (20,45.18) (30,47.89) (40,54.66) (50,55.97) (60,56.37) (70,63.40) (80,64.67) (90,65.39) (100,65.85) (110,65.60) (120,65.73) (130,65.88) (140,65.99) (150,65.71) (160,66.02) (170,66.45) (180,66.33) (190,66.14) (200,65.97) (210,66.00) (220,65.97) (230,65.62) (240,65.86) (250,66.04) (260,65.41) (270,65.65) (280,65.29) (290,65.73) (300,65.55) (310,65.55) (320,65.85) (330,65.60) (340,65.88) (350,65.91) (360,65.60) (370,65.30) (380,65.62) (390,65.92) (400,65.67)};
\end{axis}
\begin{axis}[scr axis, scr right, name=vr, at={(vl.south east)}, anchor=south west, xshift=0.22cm,
  width=\scrWr, height=3.2cm, xmin=800, xmax=1000, xtick={800,900,1000}, xticklabels={},
  ymin=13, ymax=72, ytick={20,40,60}]
\draw[scr ref] (axis cs:800,49.7) -- (axis cs:1000,49.7);
\draw[scr ref] (axis cs:800,58.0) -- (axis cs:1000,58.0);
\addplot[ccon] coordinates {(800,49.70) (805,58.06) (810,59.76) (815,60.86) (820,61.66) (825,62.07) (830,62.55) (835,62.69) (840,62.64) (845,63.21) (850,63.51) (855,63.37) (860,63.74) (865,63.64) (870,63.52) (875,63.77) (880,63.88) (885,63.86) (890,63.76) (895,63.87) (900,63.68) (905,63.77) (910,63.84) (915,63.92) (920,63.97) (925,64.20) (930,64.18) (935,64.24) (940,64.15) (945,64.19) (950,64.24) (955,64.16) (960,64.16) (965,64.25) (970,64.28) (975,64.25) (980,64.26) (985,64.28) (990,64.31) (995,64.31) (1000,64.31)};
\addplot[ccon, scr lin] coordinates {(800,58.00) (805,64.13) (810,65.29) (815,66.16) (820,66.45) (825,66.52) (830,66.94) (835,67.17) (840,67.34) (845,67.62) (850,67.59) (855,67.48) (860,67.91) (865,67.93) (870,67.93) (875,68.09) (880,68.19) (885,68.26) (890,68.36) (895,68.20) (900,68.25) (905,68.39) (910,68.51) (915,68.68) (920,68.75) (925,68.80) (930,68.75) (935,68.67) (940,68.62) (945,68.64) (950,68.84) (955,68.90) (960,68.87) (965,68.87) (970,68.86) (975,68.81) (980,68.77) (985,68.79) (990,68.81) (995,68.80) (1000,68.80)};
\addplot[cibot] coordinates {(800,49.70) (810,49.44) (820,49.65) (830,49.82) (840,50.10) (850,50.37) (860,50.43) (870,50.57) (880,50.91) (890,50.86) (900,50.98) (910,50.97) (920,50.88) (930,50.85) (940,50.99) (960,50.96) (970,50.97) (980,50.98) (990,50.97) (1000,50.96)};
\addplot[cibot, scr lin] coordinates {(800,58.00) (810,57.70) (820,58.11) (830,58.70) (840,58.78) (850,59.31) (860,59.35) (870,59.59) (880,59.64) (890,59.70) (900,59.69) (910,59.73) (920,59.61) (930,59.43) (940,59.34) (960,59.25) (970,59.25) (980,59.25) (990,59.25) (1000,59.24)};
\addplot[scr mark] coordinates {(800,49.7) (800,58.0)};
\end{axis}
\draw[black!70] (vl.south east) -- (vr.south west) coordinate[midway] (vc);
\fill[white] ([shift={(-0.105,-0.13)}]vc) -- ([shift={(-0.015,0.13)}]vc) -- ([shift={(0.105,0.13)}]vc) -- ([shift={(0.015,-0.13)}]vc) -- cycle;
\draw[scr break] ([shift={(-0.105,-0.13)}]vc) -- ++(0.09,0.26);
\draw[scr break] ([shift={(0.015,-0.13)}]vc) -- ++(0.09,0.26);
\node[scr head] at (vl.north) {from scratch};
\node[scr head] at (vr.north) {from iBOT};
\begin{axis}[scr axis, name=il, at={(vl.south west)}, anchor=north west, yshift=-0.3cm, width=\scrWl, height=2.9cm, xmin=0, xmax=400,
  xtick={0,100,200,300}, xticklabels={0,100,200,300},
  extra x ticks={400}, extra x tick labels={400}, extra x tick style={tick label style={anchor=north east, xshift=4pt}}, ymin=18, ymax=74, ytick={20,40,60},
  ylabel={IN1k $k$-NN (\%)}]
\draw[scr stage] (axis cs:30,18) -- (axis cs:30,74);
\draw[scr stage] (axis cs:60,18) -- (axis cs:60,74);
\draw[scr ref] (axis cs:0,70.09) -- (axis cs:400,70.09);
\node[scr note, anchor=north east] at (axis cs:400,70.09) {iBOT $k$-NN};
\addplot[cscr] coordinates {(10,24.89) (20,44.07) (30,52.19) (40,55.89) (50,58.61) (60,60.14) (70,60.74) (80,61.63) (90,62.22) (100,62.67) (110,62.94) (120,63.04) (130,63.40) (140,63.53) (150,63.64) (160,63.61) (170,63.83) (180,63.76) (190,63.87) (200,64.00) (210,63.92) (220,64.04) (230,63.98) (240,63.96) (250,64.06) (260,64.04) (270,64.05) (280,63.96) (290,64.00) (300,63.93) (310,63.92) (320,63.83) (330,63.95) (340,63.77) (350,63.64) (360,63.73) (370,63.74) (380,63.76) (390,63.87) (400,63.69)};
\end{axis}
\begin{axis}[scr axis, scr right, name=ir, at={(il.south east)}, anchor=south west, xshift=0.22cm,
  width=\scrWr, height=2.9cm, xmin=800, xmax=1000, xtick={900,1000}, xticklabels={900,1000},
  extra x ticks={800}, extra x tick labels={800}, extra x tick style={tick label style={anchor=north west, xshift=-4pt}},
  ymin=18, ymax=74, ytick={20,40,60}]
\draw[scr ref] (axis cs:800,70.09) -- (axis cs:1000,70.09);
\addplot[ccon] coordinates {(800,70.09) (805,69.83) (810,69.65) (815,69.63) (820,69.65) (825,69.60) (830,69.66) (835,69.67) (840,69.61) (845,69.64) (850,69.55) (855,69.56) (860,69.52) (865,69.49) (870,69.54) (875,69.61) (880,69.54) (885,69.62) (890,69.62) (895,69.67) (900,69.69) (905,69.74) (910,69.79) (915,69.91) (920,69.91) (925,70.00) (930,70.01) (935,70.04) (940,70.15) (945,70.19) (950,70.19) (955,70.15) (960,70.27) (965,70.24) (970,70.25) (975,70.26) (980,70.26) (985,70.29) (990,70.31) (995,70.30) (1000,70.29)};
\addplot[cibot] coordinates {(800,70.09) (810,69.81) (820,69.40) (830,69.11) (840,69.05) (850,69.12) (860,69.06) (870,68.96) (880,69.12) (890,69.25) (900,69.18) (910,69.35) (920,69.51) (930,69.54) (940,69.68) (960,69.77) (970,69.80) (980,69.80) (990,69.85) (1000,69.86)};
\addplot[scr mark] coordinates {(800,70.09)};
\end{axis}
\draw[black!70] (il.south east) -- (ir.south west) coordinate[midway] (ic);
\fill[white] ([shift={(-0.105,-0.13)}]ic) -- ([shift={(-0.015,0.13)}]ic) -- ([shift={(0.105,0.13)}]ic) -- ([shift={(0.015,-0.13)}]ic) -- cycle;
\draw[scr break] ([shift={(-0.105,-0.13)}]ic) -- ++(0.09,0.26);
\draw[scr break] ([shift={(0.015,-0.13)}]ic) -- ++(0.09,0.26);
\path (il.south west) -- (ir.south east) node[midway, below=0.42cm, font=\small, align=center] {Epoch};
\end{tikzpicture}

%% file: img/incontext_src/incontext_charts.tex
\definecolor{icreco}{HTML}{2864F5}
\definecolor{icibot}{HTML}{2B2B2B}
\definecolor{iccribo}{HTML}{00875A}
\definecolor{icleo}{HTML}{9467BD}
\definecolor{ichum}{HTML}{C47F00}
\ifdefined\icaxis\else\newlength{\icaxis}\newlength{\icaxish}\fi
\setlength{\icaxis}{.365\columnwidth}
\setlength{\icaxish}{3.0cm}
\pgfplotsset{
  ic axis/.style={
    scale only axis, width=\icaxis, height=\icaxish,
    ymajorgrids, xmajorgrids, grid style={black!10},
    axis x line*=bottom, axis y line*=left,
    xmin=-0.25, xmax=3.25, xtick={0,1,2,3}, xticklabels={$1/128$,$1/64$,$1/8$,$1$},
    every axis plot/.append style={line join=round},
    xlabel style={align=center, font=\small, yshift=3pt}, ylabel style={font=\small},
    tick label style={font=\small}, typeset ticklabels with strut,
  },
  ic reco/.style={icreco, line width=1.2pt, mark=*, mark size=1.7pt,
    mark options={solid, fill=icreco, draw=white, line width=0.3pt}},
  ic ibot/.style={icibot, line width=0.9pt, mark=*, mark size=1.5pt,
    mark options={solid, fill=white, draw=icibot, line width=0.7pt}},
  ic cribo/.style={iccribo, line width=0.9pt, mark=triangle*, mark size=2.0pt,
    mark options={solid, fill=iccribo, draw=white, line width=0.3pt}},
  ic leo/.style={icleo, line width=0.9pt, mark=diamond*, mark size=2.0pt,
    mark options={solid, fill=icleo, draw=white, line width=0.3pt}},
  ic hum/.style={ichum, line width=0.9pt, mark=square*, mark size=1.5pt,
    mark options={solid, fill=ichum, draw=white, line width=0.3pt}},
}
\providecommand{\ickey}[1]{\tikz[baseline=-0.6ex]{\begin{axis}[hide axis, scale only axis, width=0.55cm, height=0.2cm,
  xmin=0, xmax=1, ymin=-1, ymax=1, clip=false]
  \addplot[#1, mark indices={2}] coordinates {(0,0) (0.5,0) (1,0)};\end{axis}}}
\newcommand{\iclegend}{{\small
\ickey{ic reco}~\methodname{}\hspace{1.1em}\ickey{ic ibot}~iBOT\\[1pt]
\ickey{ic cribo}~CrIBo\hspace{1.1em}\ickey{ic leo}~Leopart\hspace{1.1em}\ickey{ic hum}~Hummingbird$^{*}$}}
\newcommand{\icchartA}{%
\begin{tikzpicture}
\begin{axis}[ic axis, ymin=32, ymax=76, ytick={40,50,60,70},
  xlabel={annotated fraction\\[1pt](a) PASCAL VOC}, ylabel={mIoU (\%)}]
\addplot[ic ibot] coordinates {(0,36.44) (1,42.46) (2,54.99) (3,61.54)};
\addplot[ic leo] coordinates {(0,44.6) (1,49.7) (2,58.4) (3,64.5)};
\addplot[ic hum] coordinates {(0,50.5) (1,57.2) (2,64.3) (3,71.8)};
\addplot[ic cribo] coordinates {(0,53.9) (1,59.9) (2,66.9) (3,72.4)};
\addplot[ic reco] coordinates {(0,55.88) (1,60.55) (2,67.17) (3,71.74)};
\end{axis}
\end{tikzpicture}}
\newcommand{\icchartB}{%
\begin{tikzpicture}
\begin{axis}[ic axis, ymin=9, ymax=32, ytick={10,15,20,25,30},
  xlabel={annotated fraction\\[1pt](b) ADE20K}]
\addplot[ic ibot] coordinates {(0,12.02) (1,14.10) (2,19.11) (3,23.21)};
\addplot[ic leo] coordinates {(0,12.9) (1,14.8) (2,19.6) (3,23.9)};
\addplot[ic hum] coordinates {(0,11.7) (1,15.1) (2,22.3) (3,29.6)};
\addplot[ic cribo] coordinates {(0,14.6) (1,17.3) (2,22.7) (3,26.6)};
\addplot[ic reco] coordinates {(0,16.29) (1,18.61) (2,24.31) (3,29.02)};
\end{axis}
\end{tikzpicture}}